\documentclass[a4paper,11pt]{article}
\usepackage{graphicx}

\usepackage{caption}
\usepackage{subcaption}
\usepackage{wrapfig}
\usepackage{amssymb}
\usepackage{hyperref}
\usepackage{amsmath}
\usepackage{amsfonts}
\usepackage{amsfonts}
\usepackage{mathtools}
\usepackage{bbm}
\usepackage{comment}
\usepackage{url}
\usepackage[symbol]{footmisc}
\usepackage{algorithm}
\usepackage{algcompatible} 
\usepackage{algpseudocode}
\usepackage[percent]{overpic}
\usepackage{mathtools}
\usepackage{amsthm}
\usepackage{float}
\usepackage{tikz}
\usepackage{physics}
\usepackage[left=1.5cm,right=1.5cm,top=2.5cm,bottom=3.5cm]{geometry}

\mathtoolsset{showonlyrefs}

\usetikzlibrary{spy}
\usetikzlibrary{positioning}
\usetikzlibrary{3d}
\usetikzlibrary{calc}
\usepackage{amsmath}
\DeclareMathOperator*{\argmax}{arg\,max}
\DeclareMathOperator*{\argmin}{arg\,min}

\newcommand{\R}{\mathbb{R}}
\newcommand{\tT}{\mathrm{T}}

\newtheoremstyle{boldremark}
    {\dimexpr\topsep/2\relax} 
    {\dimexpr\topsep/2\relax} 
    {}          
    {}          
    {\bfseries} 
    {.}         
    {.5em}      
    {}          

\theoremstyle{boldremark}
\newtheorem{rem}{Remark}
\newtheorem*{ex}{Example}

\begin{document}
\title{
{L}earning from small data sets: Patch-based regularizers in inverse problems for image reconstruction}

\author{Moritz Piening$^1$, 
Fabian Altekr\"uger$^2$, Johannes Hertrich$^2$,\\Paul Hagemann$^1$, Andrea Walther$^2$, Gabriele Steidl$^1$}

\maketitle
\renewcommand*{\thefootnote}{\arabic{footnote}}
\footnotetext[1]{{Institute of Mathematics}, {Technische Universit\"at Berlin}, {{Stra{\ss}e des 17. Juni 136, D-10623 Berlin}, {Germany}}}

\footnotetext[2]{{Department of Mathematics}, {Humboldt-Universit\"at zu Berlin}, {{Unter den Linden 6, D-10099 Berlin}, {Germany}}}

\footnotetext[3]{{Department of Computer Science}, {University College London}, {{90 High Holborn, London, WC1V 6LJ}, {United Kingdom}}}

\begin{abstract}
The solution of inverse problems
is of fundamental interest in 
medical and astronomical imaging, geophysics as well as engineering and life sciences.
Recent advances were made by using methods from  machine learning, in particular
deep neural networks.
Most of these methods require a huge amount of (paired) data and computer capacity to train the networks, which often may not be available.
Our paper addresses the issue of learning from small data sets by taking patches of very few images into account. 
We focus on the combination of model-based and data-driven methods by approximating just
the image prior, also known as regularizer in the variational model.
We review two methodically different approaches, namely 
optimizing the maximum log-likelihood of the patch distribution, 
and penalizing Wasserstein-like discrepancies of 
whole empirical patch distributions.
From the point of view of Bayesian inverse problems, we show how we can achieve uncertainty quantification by approximating the posterior using Langevin Monte Carlo methods.
We demonstrate the power of the methods 
in computed tomography, image super-resolution, and inpainting.
Indeed, the approach provides also high-quality results in  
zero-shot super-resolution, where only a low-resolution image is available.

The paper is accompanied by a GitHub repository containing implementations of all methods 
as well as data examples so that the reader can get their own insight into 
the performance. 
\end{abstract}

\section{Introduction}

 In medical and astronomical imaging, engineering, and life sciences,
 data in the form of transformed images is acquired. This transformation is the result of
a forward process that underlies a physical model.
In general, the data is corrupted by ``noise'' and the inversion of the
forward operator to get the original image back
is not possible, since there may exist many solutions and/or
the noise would be heavily amplified.
This is the typical setting in ill-posed inverse problems in imaging. 
It
is critical in applications like image-guided medical diagnostics,
where decision-making is based on the recovered
image.
One strategy for treating such problems
is to include prior knowledge of the desired images in the model.
This leads to a variational formulation of the problem
that typically contains two kinds of terms. 
The first one is a ``distance'' term between the received data
and the acquisition model,  
which includes the forward operator. The
chosen distance reflects the noise model.
The second one is an image prior, also called a regularizer, 
since it should force the variational problem to become well-posed, 
see \cite{benning2018modern,bertero2021introduction,engl1996regularization,tikhonov1963solution}.
The choice of the image prior is more difficult. A prominent example is the total variation regularizer \cite{rudin1992nonlinear}
and its vast amount of adaptations. 

The past decades have witnessed a paradigm shift 
in data processing due to the emergence of the artificial intelligence revolution. 
Sophisticated optimization strategies 
based on the reverse mode of automatic differentiation methods \cite{GW2008}, also known as backpropagation,
were developed.
The great success of deep learning methods has entered the field of inverse problems in imaging
in quite different ways.
For an overview of certain techniques, we refer to \cite{AMOS2019}.

However, for many applications, there is only limited data available such that most deep learning based methods cannot be applied. In particular, for very high dimensional problems in image processing, the necessary amount of training data pairs is often out of reach and the computational costs for model training are high. 

On the other hand, the most powerful denoising methods before deep learning  entered
the field were patched-based as BM3D \cite{dabov2007} 
or  MMSE-based techniques \cite{LNPS17,LBM2013}.

This review paper aims to advertise 
a combination of model-based and data-driven methods for
learning from small data sets.
The idea consists of retaining the distance term in the variational model 
and to establish a new regularizer
that takes the internal image statistics, 
in particular the patch distribution of very few images into account.  
To this end, we follow the path outlined below.

\paragraph{Outline of the paper.}
We start by recalling Bayesian inverse problems in Section \ref{sec:inv_prob}. In particular, we
highlight the difference between 
\begin{itemize}
    \item
the maximum a posteriori approach which leads to a variational model whose minimization provides one solution to the inverse problem,
and 
\item the approximation of the whole posterior distribution from which we intend to sample 
in order to get, e.g., uncertainty estimations. 
\end{itemize}
We demonstrate by example the 
notations of well-posedness due to Hadamard and Stuart's Bayesian viewpoint.

Section \ref{sec:istat} shows the relevance of internal image statistics.
Although we will exclusively deal with image patches, we briefly sketch feature extraction by neural networks.
Then we explain two different strategies to incorporate feature information into an image prior (regularizer).
The first one is based on maximum likelihood estimations of the patch distribution which can also be formulated in terms of minimizing the forward Kullback-Leibler divergence.
The second one penalizes Wasserstein-like divergences between
the empirical measure obtained from the patches.
Section \ref{sec:EPLL} addresses three methods for parameterizing the function 
in the maximum likelihood approach, namely via Gaussian mixture models,
the push-forward of a Gaussian by a normalizing flow, and a local adversarial approach.
Section \ref{sec:OT} shows three methods for choosing, based on the Wasserstein-2 distance, appropriate divergences
for comparing the empirical patch measures.
Having determined various patch-based regularizers,
we use them to approximate the posterior measure
and describe how to sample from this measure using a Langevin Monte Carlo approach in Section \ref{sec:uncertainty}.

Section \ref{sec:exp} illustrates the performance of the different approaches by
numerical examples in computed tomography (CT), image super-resolution, and inpainting. Moreover, we consider zero-shot reconstructions in super-resolution.
Further,  we give an example for sampling from the posterior in inpainting and for uncertainty quantification in computed tomography.
Since  quality measures in image processing reflecting the human visual impressions
are still a topic of research, 
we decided to give an  
impression of the different quality measures used in this section
at the beginning.

The code base for the experiments is made publicly available on GitHub\footnote{\url{https://github.com/MoePien/PatchbasedRegularizer}} to allow for benchmarking for future research. It includes ready-to-use regularizers within a common framework and multiple examples. Implementation on top of the popular programming language Python and the library PyTorch \cite{PyTorch2019} that enables algorithmic differentiation enhances its accessibility. 

Finally, note that alternative patch-based regularization strategies exist in addition to the presented ones, e.g., based on patch-based denoisers \cite{gilton2019learned} or an estimation of the latent dimension of the patch manifold \cite{osher2017low}.

\section{Inverse Problems: a Bayesian Viewpoint} \label{sec:inv_prob}
Throughout this paper, we consider digital gray-valued images of size $d_1 \times d_2$ as arrays $x \in \R^{d_1,d_2}$ or alternatively, by reordering their columns, as vectors $x \in \R^d$, $d = d_1 d_2$.
For simplicity, we ignore that in practice gray values are encoded as finite discrete sets. 
The methods can directly be transferred to RGB color images by considering three arrays of the above form for the red, green, and blue color channels.

In inverse problems in image processing, we are interested in the reconstruction of an image $x \in \mathbb R^d$ from its noisy measurement
\begin{equation} \label{inv_1}
y = \mathrm{noisy} (F(x)) ,
\end{equation}
where $F \colon \mathbb R^d \rightarrow \mathbb R^{\tilde d}$  is a forward operator 
and ``noisy'' describes the underlying noise model.
In all applications of this paper, $F$ is a linear operator 
which is either not invertible as in image super-resolution and inpainting or
ill-conditioned as in computed tomography, so that the direct inversion of $F$ would amplify the noise.
A typical noise model is additive Gaussian noise, resulting in
\begin{equation}
y = F(x) + \xi, 
\end{equation}
where $\xi$ is a realization of a Gaussian random variable $\Xi \sim \mathcal N(0, \sigma^2 I_{\tilde d})$. 
Recall that the density function of the normal distribution $\mathcal N(m, \Sigma)$
with mean $m \in \mathbb R^{\tilde d}$ and covariance matrix $\Sigma \in \mathbb R^{\tilde d, \tilde d}$
is given by
\begin{equation}\label{normal_density_eq}
    \varphi(x|\,m, \Sigma) \coloneqq (2\pi)^{-\frac{\tilde d}{2}}|\Sigma|^{-\frac{1}{2}}\exp\left(-\frac{1}{2}(x-m)^\intercal \Sigma^{-1}(x-m)\right).
\end{equation}
More generally, we may assume that $x$ itself is a realization of a continuous random variable $X \in \mathbb R^d$ 
with law $P_X$ determined by the density function $p_X\colon\R^d\to[0,\infty)$ with $\int_{\R^d}p_X(x)\,\text{d} x=1$, i.e., $x$ is a sample from $P_X$.
Then we can consider the random variable
\begin{equation}\label{inv:general}
Y = F(X) + \Xi, \quad \Xi \sim \mathcal N(0, \sigma^2 I_{\tilde d})
\end{equation}
and the posterior distribution $P_{X|Y=y}$ for given $y \in \R^{\tilde d}$.
The crucial law to handle this is Bayes' rule
\begin{equation}\label{bayes}
\underbrace{p_{X|Y=y}}_{\text{posterior}} (x) 
= 
\frac{\overbrace{p_{Y|X=x}}^{\text{likelihood}} (y) \overbrace{p_X(x)}^{\text{prior}}}{\underbrace{ p_Y(y)}_{\text{evidence}}}.
\end{equation}
Now we can ask at least for three different quantities.
\paragraph{1. MAP estimator} The maximum a posteriori (MAP) estimator 
provides the value with the highest probability of the posterior
\begin{align} 
x_{\mathrm{MAP}}(y) 
&\in \argmax_{x \in \R^d} \left\{p_{X|Y=y}(x) \right\}
= 
\argmax_{x \in \R^d} \left\{ \log p_{X|Y=y}(x) \right\}.
\end{align}
By Bayes' rule \eqref{bayes} and since the evidence is constant, this can be rewritten as
\begin{align} 
x_{\mathrm{MAP}}(y) 
&\in
\argmax_{x \in \R^d} \left\{ \log p_{Y|X=x}(y) + \log p_{X}(x) \right\}.
\end{align}
The first term depends on the noise model, while the second one on the distribution within the image class.
Assuming that 
$p_{Y|X=x}(y) = C \exp(-\mathcal D(Fx,y))$ and 
a \emph{Gibbs prior} distribution 
\begin{align} \label{eq:prior_reg}
p_X (x) = C_\beta \exp(-\beta \mathcal R(x)),    
\end{align}
we arrive at the variational model for solving inverse problems
\begin{align}  \label{inv_reg}                               
x_{\mathrm{MAP}}(y) 
&\in
\argmin_{x \in \R^d} 
\Big\{ \underbrace{\mathcal D(F(x),y)}_{\text{data term} }
\, + \, \beta \underbrace{\mathcal R(x)}_{\text{prior}} 
\Big\}, \quad \beta > 0.
\end{align}
Instead of a ``prior'' term, $\mathcal R$ is also known as a ``regularizer'' in inverse problems since it often transfers
the original ill-posed or ill-conditioned problem into a well-posed one. By Hadamard's definition, this means that for any $y$ there exists a unique solution that continuously depends on the input data.
For example, for Gaussian noise as in \eqref{inv:general} we have 
that $F(x) + \Xi \sim \mathcal N(F(x), \sigma^2 I_{\tilde d})$ so that by \eqref{normal_density_eq} we get
\begin{align} \label{eq:Gaussian_likelihood}
\log p_{Y|X=x}(y) =      
\log (2\pi \sigma^2)^{-\frac{\tilde d}{2}} - \frac{1}{2 \sigma^2} \|F(x) -y\|^2,
\end{align}
which results with $\alpha \coloneqq \sigma^2 \beta$ in 
\begin{equation} \label{Gaussian_noise}
x_{\mathrm{MAP}}(y) \in \argmin _{x \in \R^d} \Big\{ \frac12 \|F(x) -y\|^2 + \alpha \mathcal R(x)\Big\}.
\end{equation}
\paragraph{2. Posterior distribution} 
Here we are searching for a measure $P_{X|Y=y} \in \mathcal P(\R^d)$ 
and not as in MAP for a single sample that is most likely for a given $y$.
We will see that approximating the posterior, 
which mainly means to find a way to sample from it,
provides a tool for uncertainty quantification.
It was shown in \cite{Latz2020,Sprungk2020}
that the posterior $P_{X|Y=y}$ is often locally Lipschitz continuous with respect to $y$,
i.e.,
$$
\text{d}(P_{X|Y=y_1},P_{X|Y=y_2}) \le L \|y_1 - y_2\|
$$
with some $L >0$ and a discrepancy $\text{d}$ between measures as the  Kullback-Leibler divergence or Wasserstein distances explained in Section \ref{sec:istat}.
Indeed, this Lipschitz continuity is the key feature of 
Stuart's formulation of a well-posed \emph{Bayesian} inverse problem
\cite{Stuart2010} as a counterpart of Hadamard's definition.

There are only few settings in \eqref{inv:general} where the posterior can be computed analytically, see \cite{GFO2017,HHS2023}, 
namely
if $X$ is distributed by a \emph{Gaussian mixture model} (GMM)   
$X \sim \sum_{k=1}^K \alpha_k \mathcal N(m_k,\Sigma_k) \in \mathbb R^d$, i.e.,
\begin{equation} \label{GMM}
p_X = \sum_{k=1}^K \alpha_k \varphi(\cdot|m_k, \Sigma_k), \quad \sum_{k=1}^K \alpha_k = 1, \, \alpha_k > 0,
\end{equation}
the forward operator $F \in \R^{\tilde d,d}$ is linear and $\Xi \sim N(0,\sigma^2 I_{\tilde d})$.
Then it holds
\begin{equation} \label{posterior_GMM}
p_{X|Y=y} = \sum_{k=1}^K \tilde \alpha_k \varphi(\cdot|\tilde m_k,\tilde \Sigma_k)
\end{equation}
with
\begin{align}
\tilde \Sigma_k &\coloneqq (\tfrac{1}{\sigma^2}F^\tT F+\Sigma_k^{-1})^{-1},
\quad
\tilde m_k \coloneqq \tilde\Sigma_k (\tfrac1{\sigma^2}F^\tT y+\Sigma_k^{-1}\mu_k),
\\
\tilde \alpha_k &\coloneqq \alpha_k \exp\left(\frac12 (\tilde m_k^\tT \tilde \Sigma_k^{-1} \tilde m_k - m_k^\tT \Sigma_k^{-1} m_k)\right).
\end{align}
\paragraph{3. MMSE estimator} The maximum mean square error (MMSE) estimator is just the expectation value of the posterior, i.e.,
\begin{align}
x_{\mathrm{MMSE}}(y) 
&=   \mathbb E[X|Y=y] 
= \int_{\mathbb R^d} x p_{X|Y = y} (x) \, \text{d} x.
\end{align}
If $X \sim \mathcal N(m,\Sigma)$, $F$ is linear and $\Xi \sim N(0,\sigma^2 I_{\tilde d})$, then the MMSE can be computed analytically by
\begin{equation} \label{blue}
x_{\mathrm{MMSE}}(y) 
= 
m + \Sigma F^\tT(F \Sigma F^\tT + \sigma^2 I_{\tilde d})^{-1} (y - F m).
\end{equation}
We would like to note that for more general distributions
the estimator \eqref{blue} is known as the 
\emph{best linear unbiased estimator} (BLUE). 
MMSE techniques in conjunction with patch-based techniques were among the most powerful techniques for image denoising before ML-based methods entered the field, see \cite{LNPS17,LCBM2012}.
\medskip

The following simple one-dimensional example illustrates the behavior of the posterior in contrast to the
MAP estimator and MMSE.

\begin{figure}[!t]
\captionsetup[subfigure]{font=normal,justification=centering}
\centering
\begin{subfigure}{.245\textwidth}
  \includegraphics[width=\linewidth]{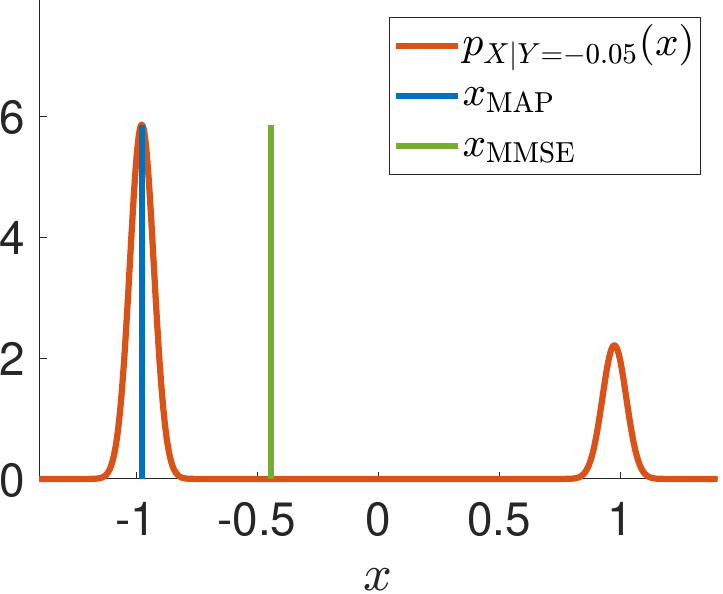}
  \caption*{$y=-0.05$}
\end{subfigure}%
\hfill
\begin{subfigure}{.245\textwidth}
  \includegraphics[width=\linewidth]{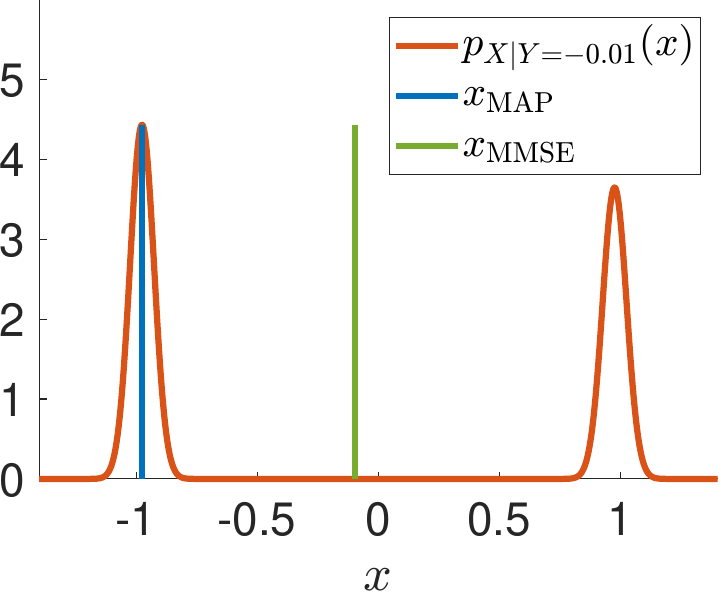}
    \caption*{$y=-0.01$}
\end{subfigure}%
\hfill
\begin{subfigure}{.245\textwidth}
  \includegraphics[width=\linewidth]{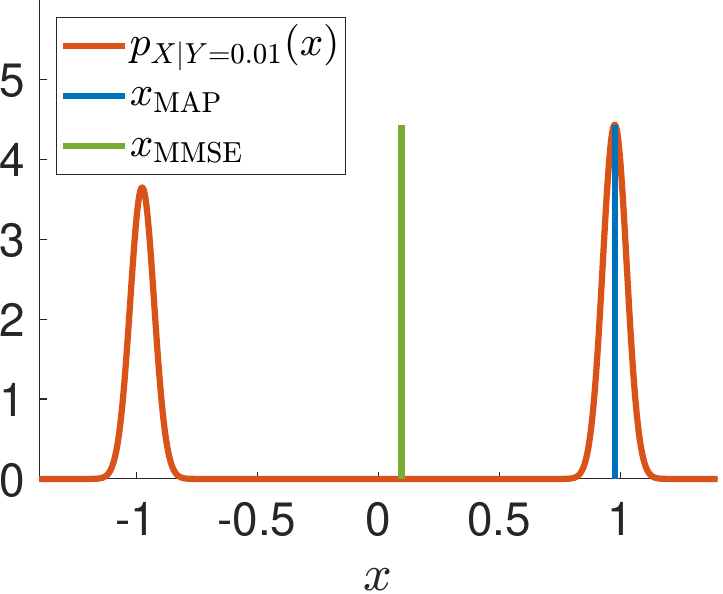}
    \caption*{$y=0.01$}
\end{subfigure}%
\hfill
\begin{subfigure}{.245\textwidth}
  \includegraphics[width=\linewidth]{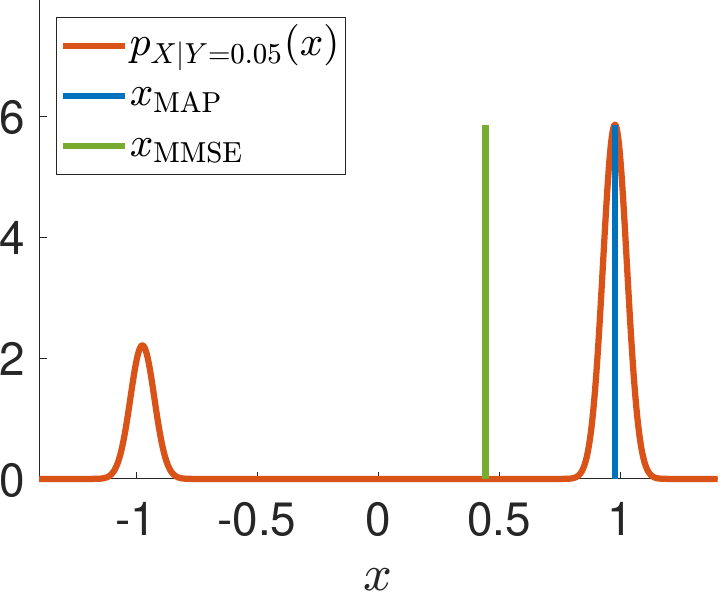}
    \caption*{$y=0.05$}
\end{subfigure}%
\caption{Posterior density (red), MAP estimator (blue), and MMSE (green) for different observations $y=-0.05,-0.01,0.01,0.05$ (from left to right). 
While the MAP estimator is discontinuous with respect to the observation $y$ at zero,  the posterior 
is continuous with respect to y. Its expectation value, the MMSE,
is far away from the value with the highest probability. Image is taken from \cite{AHS2023a}.
} \label{fig:posterior_density}
\end{figure}


\begin{ex}[\cite{AHS2023a}]
For $\varepsilon^2 = 0.05^2$, let 
$$X \sim \tfrac{1}{2} \mathcal{N}(-1,\varepsilon^2) + \tfrac{1}{2} \mathcal{N}(1,\varepsilon^2),$$
$F = I$ and $\Xi \sim \mathcal{N}(0,\sigma^2)$
with $\sigma^2= 0.1$. 
The  MAP estimator is given by
\begin{align}
x_{\mathrm{MAP}}(y) 
&=
\argmin_{x\in\R}\Big\{ \tfrac{1}{2 \sigma^2} ( y - x )^2 - \log  \big( \tfrac{1}{2} (  e^{-\frac{1}{2 \varepsilon^2} ( x - 1 )^2} 
+  e^{-\frac{1}{2 \varepsilon^2} ( x + 1 )^2} )\big) \Big\}\\
&= 
\argmin_{x\in\R}\Big\{ \tfrac{1}{2 \sigma^2} ( y - x )^2  + \tfrac{1}{2 \varepsilon^2} (x^2 + 1) - \log \left( \cosh \left(\frac{x}{\varepsilon^2} \right) \right)\Big\}.
\end{align}
This minimization problem has a unique solution for $y \not = 0$ which we computed numerically.
By \eqref{posterior_GMM}, we can compute the posterior 
\begin{align}
P_{X|Y=y} = \frac{1}{\tilde \alpha_1 + \tilde \alpha_2} (\tilde \alpha_1 \mathcal{N}(\cdot | \tilde m_1 , \tilde \sigma^2) + \tilde \alpha_2 \mathcal{N}(\cdot | \tilde m_2 \tilde \sigma^2) )
\end{align}
with 
\begin{align}
\tilde \sigma^2 &= \frac{\sigma^2 \varepsilon^2}{\sigma^2 + \varepsilon^2}, \quad 
\tilde m_1 = \frac{\varepsilon^2 y + \sigma^2}{\varepsilon^2 + \sigma^2}, \quad \tilde m_2 = \frac{\varepsilon^2 y - \sigma^2}{\varepsilon^2 + \sigma^2},
\\
\tilde \alpha_1 &= \frac{1}{2 \varepsilon} \exp \Big(\frac{1}{2\varepsilon^2} \Big( \frac{(\varepsilon^2 y + \sigma^2)^2}{\sigma^2  (\varepsilon^2 + \sigma^2)} - 1 \Big) \Big), \quad 
\tilde \alpha_2 = \frac{1}{2 \varepsilon} 
\exp \Big(\frac{1}{2\varepsilon^2} \Big( \frac{(\varepsilon^2 y - \sigma^2)^2}{\sigma^2  (\varepsilon^2 + \sigma^2)} - 1 \Big) \Big) .
\end{align}
Finally, the MMSE is given by 
\begin{align}
x_{\mathrm{MMSE}}(y) 
&=
\frac{1}{\tilde \alpha_1 + \tilde \alpha_2}\frac{1}{\varepsilon ( \varepsilon^2 + \sigma^2)} e^{\frac{\varepsilon^2 y^2 - \sigma^2}{2 \sigma^2 (\varepsilon^2 + \sigma^2)} } \Big(\varepsilon^2 y 
\cosh \big(\frac{y}{\varepsilon^2 + \sigma^2} \big) 
+ 
\sigma^2 
\sinh \big( \frac{y}{\varepsilon^2 + \sigma^2} \big) \Big).
\end{align}
Figure \ref{fig:posterior_density} illustrates the behavior of the three quantities. The main observation is that in contrast to the posterior,
the MAP estimator is discontinuous at $y=0$.
\end{ex}

In the numerical experiments we will consider the MAP estimator in Sections~\ref{sec:CT}, \ref{sec:superres}, and \ref{sec:img_inpainting}, and the posterior sampling in Section~\ref{sec:posterior_inpainting}.

\section{Internal Image Statistics} \label{sec:istat}
In this paper, when dealing with the MAP estimator, 
i.e., with problems of the form \eqref{inv_reg},
we follow a physics-informed approach, where both the forward operator and the noise model are known. 
Then the data
term $\mathcal D(F(x),y)$ is completely determined.
The challenging part is the modeling of the prior distribution $P_X$, where we only know 
samples from.
In contrast to deep learning methods which rely on a huge amount of (paired) ground truth data,
we are in a situation, where only one or a few images are available.
Then, instead of working with the distribution $P_X$ of whole images in the prior, we consider typical features of the images and ask for the feature distribution. These features live in a much lower dimensional space than the images. Indeed, one key finding in image processing was the expressiveness of this internal image statistics \cite{simoncelli2001natural,torralba2003statistics}.
Clearly, there are many ways to extract meaningful features and we refer only to the``field of experts'' framework \cite{roth2005fields} here.
In the following, we explain two typical choices of meaningful features, namely image patches and features obtained from a nonlinear filtering process of a neural network. 

\begin{figure}[!t]
    \centering
    \begin{overpic}[width=\textwidth]{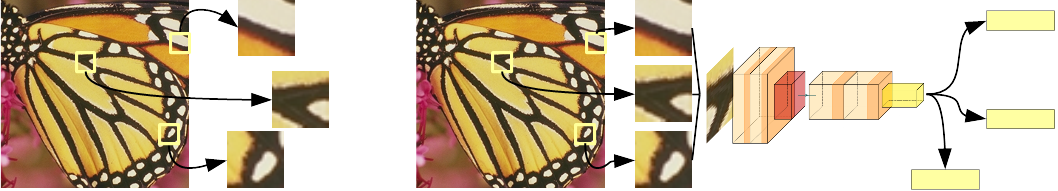} 
     \put (18.3,11.5) {$P_{245, 40}$} 
     \put (18.7,6.5) {$P_{105, 55}$} 
     \put (18,-1.3) {$P_{235, 200}$} 
     \put (83,13) {$F_{245, 40}$} 
     \put (90.7,9.1) {$F_{105, 55}$}
     \put (81,4.7) {$F_{235, 200}$}     
    \end{overpic}
    \caption{Visualization of the process of patch extraction (left) and hidden feature extraction (right).} \label{fig:extraction}
\end{figure}
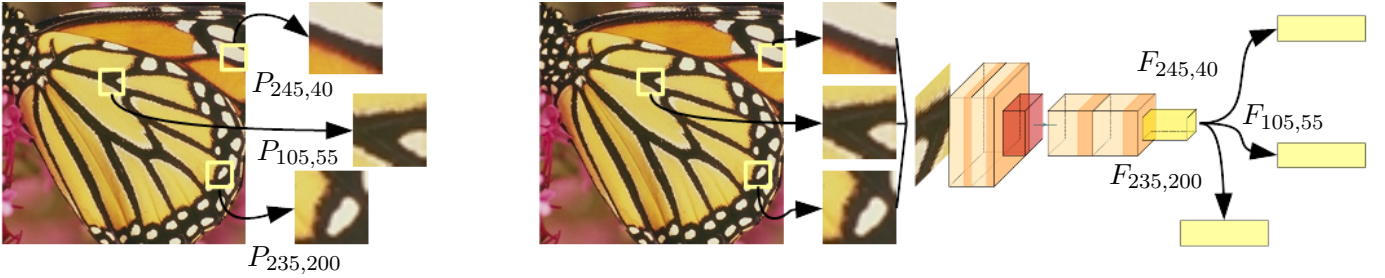

\begin{figure}[!tb]
    \centering
    \begin{overpic}[width=0.9\textwidth]{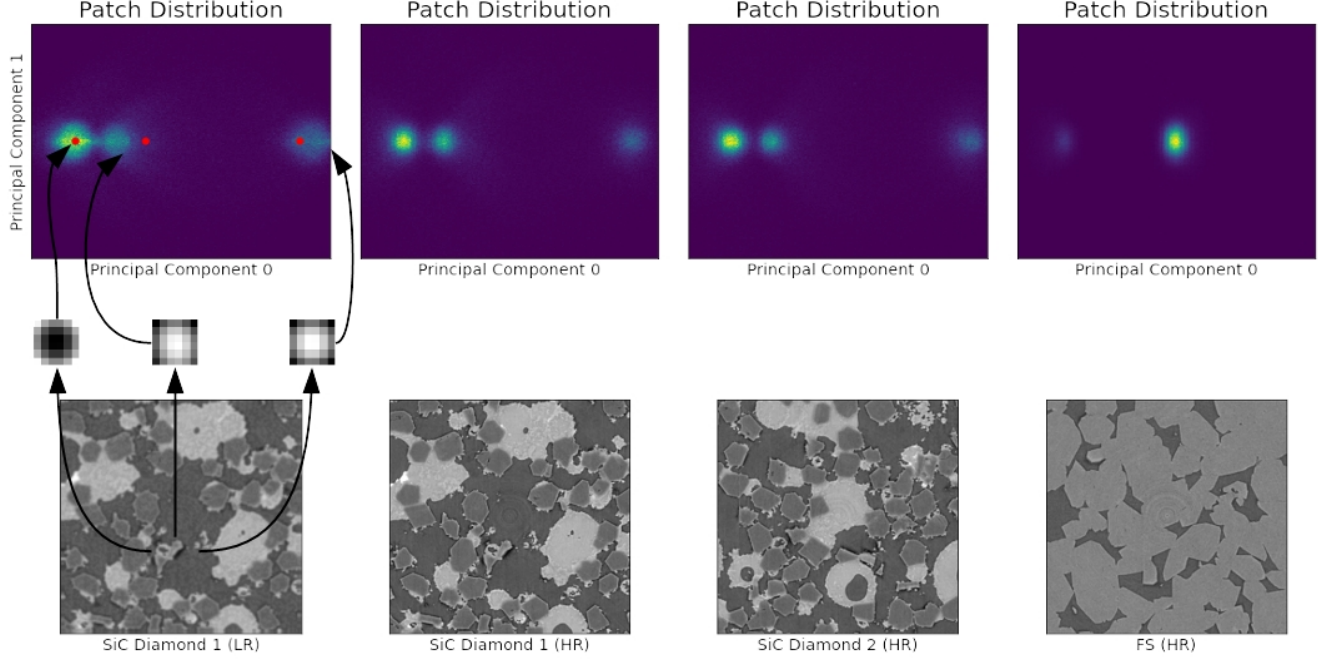} 
    \end{overpic}
     \caption[]{
    Left to right:
    Illustration of the patch distribution of a 
    low resolution (LR) image (downsampled) from a composite of silicon and diamonds (``SiC Diamond''), 
    two different high-resolution (HR) images from the same material
    and one from the material Fontainebleau sandstone (``FS'').
    Patches of size $6 \times 6$ from the respective images are extracted
    and a principal component analysis is applied to the corresponding vectors in $\mathbb R^{36}$
    in order to project onto the plane spanned by the two principal directions of the largest patch variance.  
    The empirical distribution of the first two principal components in the form of 2D histograms is depicted in the top row.
    All patches of HR images of the first material have a similar distribution, 
    which is easy to distinguish from those of the second material. 
    This is also true for the LR image but with a slightly larger spread of the clusters. 
    Example patches (red dots) from the histogram are displayed in the upper row, left. }
    \label{patchpca_fig1}
\end{figure}

\paragraph{Image patches}
Image patches are square-shaped (or rectangular) regions of size $p \times p$ within an image $x \in \R^{d_1,d_2}$
which can be extracted by operators $P_i: \R^{d_1,d_2} \to \R^{p,p}$, $i=(i_1,i_2) \in \{1,\ldots,d_1\}\times \{1,\ldots,d_2\}$
via $P_i(x) = (x_{l_1,l_2})_{l_1=i_1,l_2 = i_2}^{i_1+p,i_2+p}$.
The patch extraction is visualized in Figure \ref{fig:extraction} (left).
The use of such patches for image reconstruction has a long history \cite{dong2011sparsity,peyre2008non} 
and statistical analyses of empirical patch distributions reveal their importance to image characterization \cite{zontak2011internalpatchstatistics}. 
Furthermore, the patch distributions are similar at different scales for many image classes. Therefore, the approach is not sensitive to scale shifts.
Figure \ref{patchpca_fig1} illustrates this behavior,  see also Figure \ref{patchpca_scales_fig}.
Indeed, replication of patch distributions by means of patch sampling \cite{liang2001realtexture} or statistical distance minimization \cite{houdard2021wasserstein} enables the synthesis of high-quality texture images. Neural network image generators are able to generate diverse outputs on the basis of patch discriminators \cite{shaham2019singan, shocher2019ingan} or patch distribution matching \cite{elnekave2022generating, granot2022drop}. Furthermore, patch-matching methods have successfully been employed for style transfer \cite{chen2016patchstyle}.

\paragraph{Neural network filtered features}
Several feature methods for image reconstruction, as, e.g., in the ``field of experts'' framework \cite{roth2005fields} are based on features obtained from various linear filter responses 
possibly finally followed by an application of a nonlinear function.
 More recently, such techniques were further extended 
 by using a pre-trained classification convolutional neural network, e.g., a VGG architecture trained on the ImageNet dataset \cite{vgg19}.
 In each layer, multiple nonlinear filters (convolutions and component-wise nonlinear activation function)
 are applied to the downsampled result of the previous layer.  
   Typically, the outputs of the first convolutional layer after a downsampling step are utilized as \emph{hidden features}.
  This is illustrated in Figure \ref{fig:extraction} right.
 Since every nonlinear filter of a convolutional filter is applied locally, these extracted features represent nonlinear transformations of patches of different sizes.  
 The use of such features has been pioneered by Gatys et al. \cite{gatys2016image},
 who used them to construct a loss function for the style transfer between two images based on a statistical distance between their hidden feature distributions \cite{demyst_neural_style}. Furthermore, such features have been utilized for texture synthesis in \cite{gatys2015texture, houdard2023generative} and for image similarity comparison in \cite{lpips}.

\paragraph{Internal image statistics in image priors}
In the rest of the paper, we will concentrate on patches as features. We assume that we are
given a small number $n$ of images from an image class, say, $n=1$ high-resolution material image or $n=6$ computed tomography scans. 
For simplicity, let us enumerate the patch operators by $P_i$, $i=1,\ldots,N$.
Further, let us denote by $P_X$ the \emph{patch distribution}.
Then we will follow two different strategies
to incorporate them into the prior $\mathcal R$
of model \eqref{inv_reg}, which we describe next. 
\begin{itemize}
    \item[1.] 
    \textbf{Patch maximum log-likelihood }:
    We approximate the patch distribution $P_X$ by a distribution $P_{X_\theta}$
    with density  $p_\theta$ depending on some parameter $\theta$. Then, we learn its parameter via a \emph{maximum log-likelihood} (ML) estimator:
    \begin{align} 
    \hat \theta &= 
    \argmax_\theta \Big\{\prod\limits_{j=1}^n  \prod_{i=1}^N p_\theta \left(P_i(x_j) \right)\Big\}
= \argmax_\theta \Big\{\sum_{j=1}^n  \sum_{i=1}^N \log \left( p_\theta\left(P_i(x_j)\right) \right)\Big\}\\
&= 
\argmin_\theta \Big\{ \underbrace{-\sum_{j=1}^n  \sum_{i=1}^N \log \left( p_\theta \left(P_i(x_j) \right) \right) }_{=:\mathcal L(\theta)} \Big\} . \label{ML}
    \end{align}
    Once the optimal parameter $\hat \theta$ is determined
    by minimizing the \emph{loss function} 
    $\mathcal L(\theta)$,
    we can use 
    \begin{equation} \label{MLR}
    \mathcal R(x) \coloneqq - \frac{1}{N} \sum_{i=1}^N \log \left( p_{\hat \theta} \left(\text{P}_i(x)\right) \right)
    \end{equation}
    as a prior in our minimization problem \eqref{inv_reg}. Indeed this value should become small,
    if  the patches of the wanted image $x$ are distributed according to $p_{\hat \theta}$.

    The above model can be derived from another point of view using the \emph{Kullback-Leibler} (KL) \emph{divergence} between $P_X$ and $P_{X_\theta}$.
    As a measure divergence, the KL is non-negative
    and becomes zero if and only if both measures coincide.
    For $P_X$ and
    $P_{X_\theta}$ on $\R^d$, $d = p^2$ with densities $p_X$ and $p_{\theta}$, respectively, the KL divergence is given (if it exists) by
    \begin{align}\label{eq:kl_div}
    \text{KL}(P_X, P_{ X_\theta})
    &= 
\int_{\R^d} \log 
\left(
\frac{ p_X(x) }{ p_{ \theta}(x) } 
\right) 
p_X (x) \, \text{d} x
\\
&=  \int_{\R^d} \underbrace{\log \left( p_X(x) \right) p_X(x)  }_{\mathrm{const}}\, \text{d} x
-  
\int_{\R^d} 
\log \left( 
p_{\theta}(x) 
\right) 
p_X(x) \, \text{d} x.
\end{align}
Skipping the constant part with respect to $\theta$, this becomes
$$
\text{KL}(P_X, P_{X_\theta})
    \propto  
-  
\int_{\R^d} 
\log \left( 
p_{\theta}(x) 
\right) 
p_X(x) \, \text{d} x 
= - \mathbb E_{x \sim X} \big[ p_\theta(x) \big].    
$$
Here $\propto$ denotes equality up to an additive constant.
Replacing the expectation value with the empirical one and neglecting the factor $\frac{1}{nN}$, we arrive exactly at the loss function $\mathcal L(\theta)$ in \eqref{ML}.
    \item[2.]
    \textbf{Divergences between empirical patch measures}:
    We can associate empirical measures to the image patches by
    \begin{equation} \label{e-measures}
    \nu  \coloneqq \frac{1}{nN} \sum_{j=1}^n \sum_{i=1}^N \delta_{P_i(x_j)} 
    \quad \text{and} \quad
    \mu_x  \coloneqq \frac1N\sum_{i=1}^N \delta_{P_i(x)} 
    \end{equation}
    as illustrated in Figure \ref{fig:pm}.
    Then we use a prior
        \begin{equation} \label{w_prior}
    \mathcal R(x) \coloneqq \text{dist}(\mu_x,\nu),
    \end{equation}  
with some distance, respectively divergence, between measures.

\begin{figure}[!b]
\captionsetup[subfigure]{font=normal,justification=centering}
\begin{center}
\begin{subfigure}[c]{.42\textwidth}
\includegraphics[width=\textwidth]{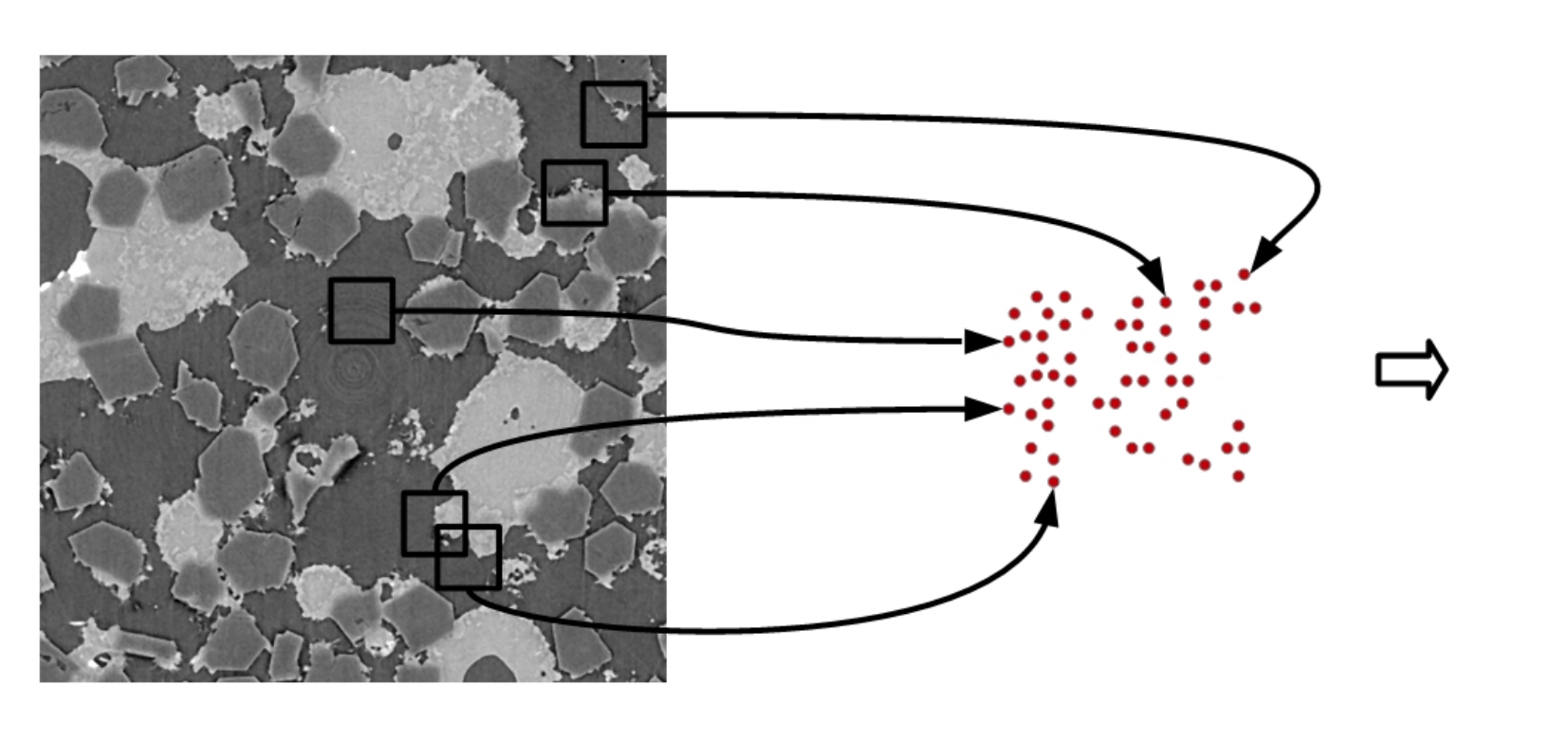}
\vspace{-1cm}
\caption*{}
\end{subfigure}
\hspace{-.35cm}
\begin{subfigure}[c]{.1\textwidth}
$$
\mu_x=\frac1N\sum_{i=1}^N\delta_{P_ix}
$$
\caption*{} 
\vspace{-.5cm}
\end{subfigure}
\caption{From patches to measures.}\label{fig:pm}
\end{center}
\begin{picture}(0, 0)
    \put (145, 40) {\scalebox{1}{$x \in \mathbb{R}^{600, 600}$}}
    \put (270, 40) {\scalebox{1}{$\mathbb{R}^{36}$}}
    \put (370, 40) {\scalebox{1}{$\mathcal P(\mathbb{R}^{36})$}}
    \put (280, 122) {\scalebox{.7}{$\operatorname{P}_{10} x$}}
    \put (245, 110) {\scalebox{.7}{$\operatorname{P}_{20} x$}}
    \put (255, 57.5) {\scalebox{.7}{$\operatorname{P}_{1000} x$}}
\end{picture}
\end{figure}
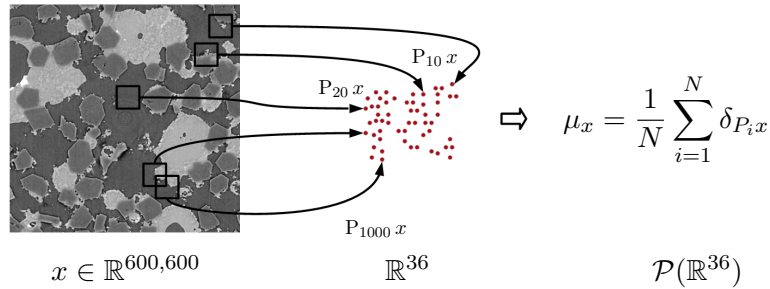

Our distances of choice in \eqref{w_prior}
will be Wasserstein-like distances.
Let $\mathcal P_p(\mathbb R^d)$, $p \in [1,\infty)$,
denote the set of probability measures with finite $p$-th moments.
The  Wasserstein-$p$ distance 
$\operatorname{W}_p \colon \mathcal P_p(\mathbb R^d) \times \mathcal P_p(\mathbb R^d) \to \R$
is defined by
    \begin{equation} \label{wdist}
\operatorname{W}_p^p(\mu, \nu) \coloneqq  
\inf_{\gamma \in \Pi(\mu, \nu)} \int \limits_{\R^d \times \R^d} \| x-y\|^p \,\text{d} \pi(x, y),
\end{equation}
where 
$\Pi(\mu, \nu) \coloneqq 
\{\pi \in \mathcal{P}(\R^d \times \R^d):
(\text{proj}_1)_\# \pi = \mu, (\text{proj}_2)_\# \pi = \nu\}$
is the set of all couplings with marginals $\mu$ and $\nu$.
Here $\text{proj}_i$, $i=1,2$ denote the projection onto the $i$-th marginals.
Further, we used the notation of a push-forward measure.
In general, for a measurable function $T:\R^d \to \R^{\tilde d}$ 
and a measure $\mu$ on $\R^d$, the
\emph{push-forward measure} of $\mu$ by $T$ on $\R^{\tilde d}$ is defined as
$$
T_\# \mu (A) = \mu \left( T^{-1} (A) \right), 
\quad \text{i.e.}, \quad
\int_{A}  g(y) \, \text{d} (T_\# \mu) (y)
= 
\int_{T^{-1} (A)}  g\left(T(x) \right) \, \text{d}  \mu (x)
$$
for all $g \in C_0(\R^{\tilde d})$ and for all Borel measurable sets $A \subseteq \R^{\tilde d}$.
The push-forward measure
and the corresponding densities $p_\mu$ 
and $p_{T_{\#}\mu}$ are related by the 
\emph{transformation formula}
\begin{equation} \label{trans_formula}
p_{T_{\#}\mu} (x) = p_\mu
\left( T^{-1} (x)\right) 
|
\det \left( \nabla T^{-1} (x) \right)
|.
\end{equation}
An example of the Wasserstein-2 distance is given in Figure \ref{optimal_transport_plan}.
\end{itemize}
Note that both strategies can easily be generalized to multi-scale regularization by adding the composition $\mathcal R  \circ D$ for a downsampling operator $D$.

\section{Patch Maximum Log-Likelihood Priors} \label{sec:EPLL}
For the prior in \eqref{MLR}, it remains to find an appropriate 
parameterized function $p(\cdot|\theta)$.
In the following, we present three different regularizers, namely obtained via
Gaussian mixture models (GMM-EPLL),
normalizing flows (patchNR),
and adversarial neural networks (ALR). 

We will compare their performance in inverse problems later in the experimental section.

\subsection{Gaussian Mixture Model (GMM-EPLL)}\label{subsec:GMM}
A classical approach assumes that
the patch distribution can be approximated by a GMM \eqref{GMM}, i.e.,
\begin{equation}
p_{\theta}(x) = \sum_{k=1}^{K} \alpha_k \varphi(x\,|\, m_k, \Sigma_k), \quad 
\theta = \left(m_k,\Sigma_k\right)_{k=1}^K.
\end{equation}
This is justified by the fact that any probability distribution can be approximated arbitrarily well in the Wasserstein distance
by a GMM. However, the number of modes $K$ has to be fixed in advance.
Then the maximization problem becomes
$$
\hat \theta = \argmax_\theta \bigg\{\sum_{j=1}^n \sum_{i=1}^N  \log \left( \sum\limits_{k=1}^{K} \alpha_k \varphi\left(\operatorname{P}_{i}(x_j) \, | \, m_k, \Sigma_k\right) \right)\bigg\}.
$$
This is typically solved by the Expectation-Maximization (EM) Algorithm \ref{alg:em_gmm}, 
with the guarantee of convergence to a local maximizer, see \cite{dempster1977maximum}.
The corresponding regularizer \eqref{MLR} becomes
\begin{align*}
\mathcal{R} (x) = \text{EPLL}(x) \coloneqq \frac{1}{N}
\sum_{i=1}^N - \log \left( \sum\limits_{k=1}^{K} \alpha_k \varphi\left(\operatorname{P}_{i}(x) \, | \, m_k, \Sigma_k\right) \right).
\end{align*}
It was suggested for solving inverse problems under the name expected patch log-likelihood (EPLL)
by Zoran and Weiss \cite{epll}.

\begin{algorithm}[!t]
\caption{Expectation-Maximization for Gaussian Mixture Model}
    \label{alg:em_gmm}    
\begin{algorithmic}
\State \textbf{Input:} Patches $\{x_1,  \ldots, x_{M}\}$, $M = Nn$, number of GMM components $K$, stopping criterion 
\State \textbf{Output:} GMM parameters $\{m_k, \Sigma_k, \alpha_k\}_{k=1}^{K}$
\State \textbf{Initialization:} $\{m_k^{(0)}, \Sigma_k^{(0)}, \alpha_k^{(0)}\}_{k=1}^{K}$
\FOR{$r=0, 1, \ldots$ until stopping criterion}
    \begin{enumerate}
        \item \textbf{E}xpectation step:
            For $k=1, \ldots,K$ and $i=1,\ldots,M$ compute 
            \[
                \beta^{(r)}_{i,k} = \frac{\alpha_k^{(r)} \varphi(x_i|m_k^{(r)}, \Sigma_k^{(r)})}{\sum_{j=1}^{K} \alpha_j^{(t)} \varphi(x_i|m_j^{(r)},\Sigma_k^{(r)})}
            \]        
        \item \textbf{M}aximization step:
        For $k=1, \ldots,K$ and $i=1,\ldots,M$
            update parameters
            \begin{align*}
                m_k^{(r+1)} &= \frac{\sum_{i=1}^{M} \beta^{(r)}_{i,k} x_i}{\sum_{i=1}^{M} \beta^{(r)}_{i,k}},  \\
                \Sigma_k^{(r+1)} 
                &= \frac{\sum_{i=1}^{M} \beta^{(t)}_{i,k} (x_i - m_k^{(r+1)})(x_i - m_k^{(r+1)})^\intercal}{\sum_{i=1}^{M} \beta^{(r)}_{i,k}}, \\
                \alpha_k^{(r+1)} 
                &= \frac{1}{M} \sum_{i=1}^{M} \beta^{(r)}_{i,k}
            \end{align*}        
    \end{enumerate}
\ENDFOR
\end{algorithmic}
\end{algorithm}

\begin{rem} \label{rem:prior_EPLL}
By the relation \eqref{eq:prior_reg} the EPLL defines a prior distribution $p_X (x) = C_\beta \exp(-\beta \text{EPLL}(x))$. The integrability of the function $p_X$ can be shown by similar arguments as in the proof of \cite[Prop. 6]{altekruger2023patchnr}.
\end{rem}

While originally the variational formulation \eqref{inv_reg} with the EPLL was solved using \emph{half quadratic splitting}, in our implementation we use a stochastic gradient descent for minimizing \eqref{inv_reg}.
Meanwhile there exist many extensions and improvements:
GMMs may be replaced with other families of  distributions \cite{deledalle2018image,HHLS2021,he2018image,NHAB2023}
or multiple image scales can be included \cite{papyan2015multi}. 
The intrinsic dimension of the Gaussian components 
can be restricted as in \cite{houdard2018high, wang2013sure} or in the PCA reduced GMM model \cite{HL2022}.
Finally, image restoration can be accelerated by introducing flat-tail Gaussian components, balanced search trees, and restricting the sum of the EPLL to a stochastically chosen subset of patch indices \cite{parameswaran2018accelerating}.
For the inclusion of learned local features into the model, we refer to \cite{yu2023epll, zach2023explicit}.

In the next subsections, we will see that machine learning based models can further improve the performance.

\subsection{Patch Normalizing Flow Regularizer (patchNR)}\label{subsec:PatchNR}

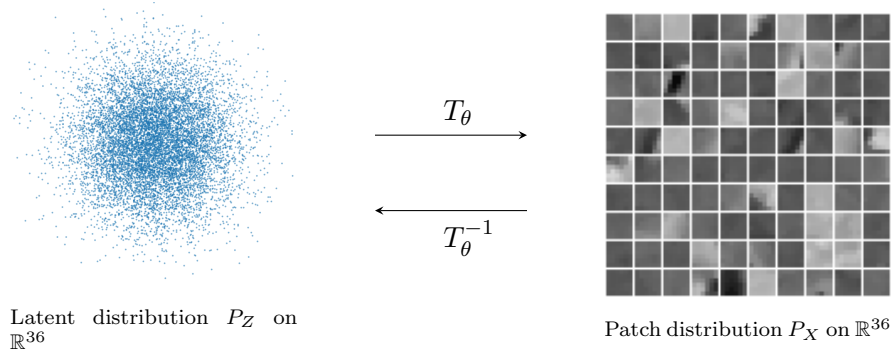
\begin{figure}[b]
\centering
\begin{subfigure}[c]{0.21\textwidth}
\includegraphics[width=\textwidth]{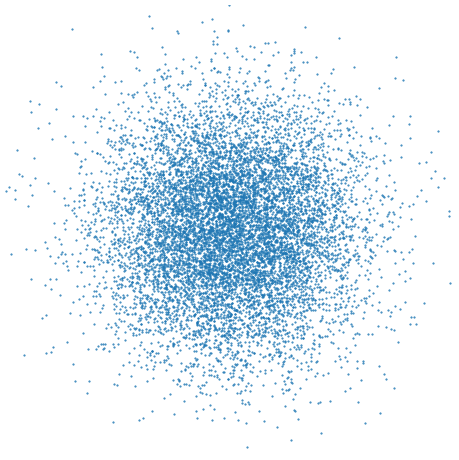}
\caption*{\scriptsize Latent distribution $P_Z$ on $\R^{36}$}
\end{subfigure}
\begin{subfigure}[c]{0.21\textwidth}
\centering
\begin{tikzpicture}
    \draw [-stealth](-0.5,0) --  node [text width=0.2cm,midway,above]{$T_{\theta}$} (1.5,0);
    \draw [stealth-](-0.5,-1) -- node [text width=0.2cm,midway,below]{$T_{\theta}^{-1}$} (1.5,-1);
\end{tikzpicture}
\end{subfigure}
\begin{subfigure}[c]{0.21\textwidth}
\includegraphics[width=\textwidth]{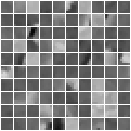}
\caption*{\scriptsize Patch distribution $P_X$ on $\R^{36}$}
\end{subfigure}
\caption{NF between (samples of) standard normal distribution (projection on $\R^2$) and distribution of $6\times 6$ patches.}
\label{NF_patch}
\end{figure}

\begin{figure}[t]
\begin{center}
\includegraphics[width = 0.65\textwidth]{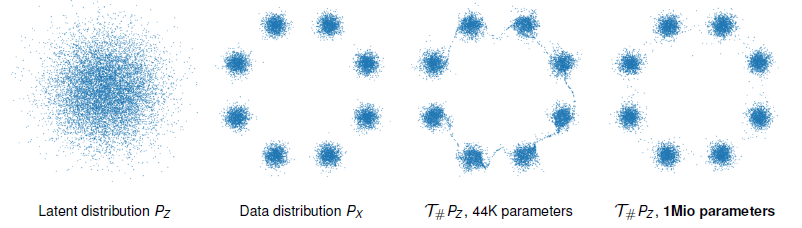}
\caption{NFs between (samples of) 2D standard Gaussian distribution
and multimodal distribution. A good approximation is only possible with the rightmost NF which has
a higher number of parameters and here also a higher Lipschitz constant.}
\label{instabil}
\end{center}
\end{figure}

Another successful approach models the patch distribution using
\emph{normalizing flows} (NFs) \cite{altekruger2023patchnr}. NFs are invertible differentiable mappings.
Currently, there are two main structures that achieve invertibility of a neural network, 
namely invertible residual networks \cite{BGCDJ2019}
and directly invertible networks \cite{ardizzone2018analyzing,dinh2016density}.
For the patchNR, the directly invertible networks are of interest.
The invertibility is ensured by the special network structure 
which in the simplest case consists of a concatenation of $K$ invertible,  differentiable mappings
$T_{\theta_k}: \R^d \to \R^d$
(and some permutation matrices which are skipped for simplicity)
$$
T_\theta=  T_{\theta_K} \circ \dots \circ  T_{\theta_1}.
$$
The invertibility is ensured by a special splitting structure, namely for $d_1 + d_2 = d$, we set
\begin{align} \label{struct}
  T_{\theta_k} (z_1,z_2) 
  &= \begin{pmatrix} x_1\\x_2  \end{pmatrix}
  := \begin{pmatrix} z_1\\ 
          z_1 \odot \exp \big(s_{\theta_{k_1}}(z_1) \big) 
          +  
          t_{\theta_{k_2}}(z_1)  
					\end{pmatrix},	 \quad
     z_i,x_i \in \R^{d_i}, \, i=1,2,    
\end{align}	
where $s_{\theta_{k_1}}, t_{\theta_{k_2}}$ are arbitrary neural networks 
and $\odot$ denotes the component-wise multiplication.
Then its inverse can be simply computed by
\begin{align}
T_{\theta_k}^{-1}(x_1,x_2) 
&=  \begin{pmatrix} z_1\\z_2 \end{pmatrix} 
=
   \begin{pmatrix} x_1\\ 
          \big(x_2 - t_{\theta_{k_2}}(x_1) \big)
          \odot 
          \exp \big(-s_{\theta_{k_1}}(x_1)\big) 
					\end{pmatrix} .    
\end{align}	
This is the simplest, \emph{Real NVP} network architecture \cite{dinh2016density}.
A more sophisticated one is given in \cite{ardizzone2018analyzing}.
Now the idea is to approximate our unknown patch distribution $P_X$
on $\R^d$, $d = p^2$,
using the push-forward by $T_\theta$ of a measure $P_Z$, where it is easy to sample from as, e.g., the $d$-dimensional standard normal distribution $Z \sim \mathcal N(0, I_d)$. Our goal becomes
$$
P_X \approx (T_\theta)_\# P_Z = P_{X_\theta}.
$$
The NF between (samples of) the standard normal distribution in $\R^{36}$
and the distribution of material image patches
is illustrated in Figure \ref{NF_patch}.
Let us take the KL approach to find the parameters of $p_\theta = p_{ (T_\theta)_\# P_Z}$, i.e.,
\begin{align}
\mathrm{KL}\left (P_X, (T_\theta)_\# P_Z \right) 
&= 
\int_{\R^d} \log 
\left(
\frac{ p_X(x) }{ p_{ (T_\theta)_\# P_Z}(x) } 
\right) 
p_X (x) \, \text{d} x
\\
&= 
\int_{\R^d} \underbrace{\log \left( p_X(x) \right) p_X(x) }_{\mathrm{const}}\, \text{d} x
-  
\int_{\R^d} 
\log \left( 
p_{\mathcal (T_\theta)_\# P_Z}(x) 
\right) 
p_X(x) \, \text{d} x.
\end{align}
Using the transformation formula \eqref{trans_formula},
we obtain
(up to a constant)
\begin{align*}
\mathrm{KL}\left(P_X,  (T_\theta)_\# P_Z \right)
&\propto
-  \int_{\R^d}  \log 
\left( 
p_Z
\left( 
(T_\theta)^{-1} (x) 
\right) \, |\mathrm{det} \nabla T_\theta^{-1}(x)| 
\right) 
p_X(x) \, \text{d} x \\
&=
-\mathbb E_{x \sim P_X} 
\big[\log 
p_Z 
\left( 
 T_\theta^{-1}(x)
\right) 
+
  \log 
\left( 
|\mathrm{det} \nabla T_\theta^{-1}(x)|  
\right)  
\big]
\end{align*}
and since $P_Z$ is standard normally distributed further
\begin{align}
\mathrm{KL}\left(P_X,  (T_\theta)_\# P_Z \right)
&\propto
 \mathbb E_{x \sim P_X} 
\Big[ \tfrac12 \| T_\theta^{-1}(x)\|^2 
-
\log \big( 
|\mathrm{det} \nabla  T_\theta^{-1}(x)|  
\big)  
\Big].
\end{align} 
Taking the empirical expectation
provides us with the ML loss function 
$$
\mathcal L (\theta)  
= 
\sum_{j=1}^n \sum_{i=1}^N 
\tfrac12 \| T_\theta^{-1} \left(\operatorname{P}_{i}(x_j) \right) \|^2 
- \log \big( | \det \nabla T_\theta^{-1}\left(\operatorname{P}_{i}(x_j)  \right) | \big).
$$
To minimize this function we use a stochastic gradient descent algorithm, where the special structure \eqref{struct} of the network
can be utilized for the gradient computations.
Once good network parameters $\hat \theta$ are found,
we introduce in \eqref{inv_reg}
the \emph{patchNR}
\begin{equation}
 \mathcal R(x) = \text{patchNR} (x)
 \coloneqq \frac{1}{N}
 \sum_{i=1}^N \tfrac12 \| T_{\hat \theta}^{-1} 
 \left(\operatorname{P}_{i}(x) \right)\|^2 - \log\big( |\det \nabla T_{\hat \theta}^{-1}\left(\operatorname{P}_{i}(x) \right)|  \big).
\end{equation}

\begin{rem}\label{rem:prior_patchNR}
By the relation \eqref{eq:prior_reg} the patchNR defines a prior distribution $p_X (x) = C_\beta \exp(-\beta \text{patchNR}(x))$. The integrability of the function $p_X$ is shown in \cite[Prop. 6]{altekruger2023patchnr}. 
\end{rem}
\begin{rem}
The $\text{KL}$ divergence of measures is neither symmetric nor fulfills a triangular inequality.
Concerning symmetry, the setting $\text{KL}(P_X,P_{X_\theta})$ 
is called \emph{forward KL}.
Changing the order of the measures gives the \emph{backward} (or reverse) \emph{KL} in 
$\text{KL}(P_{X_\theta},P_X)$. These settings have different 
properties as being \emph{mode seeking} or \emph{mode covering}
and the loss functions rely on different data inputs, see
\cite{hagemann2022stochastic}.
There are also mixed variants 
$\alpha \text{KL}(P_X,P_{X_\theta}) + (1-\alpha) \text{KL}(P_{X_\theta},P_X)$, $\alpha \in (0,1)$

as well as the Jensen–Shannon divergence \cite{GPMXWOCB2014}.
\end{rem}

Unfortunately, NFs mapping unimodal to multimodal distributions suffer from exploding Lipschitz constants and are therefore
sensitive to adversarial attacks \cite{behrmann2020understanding,HN2021,JKYB2020,SBDD2022}. 
This is demonstrated in Figure \ref{instabil}.
A remedy is the use of GMMs for latent distribution \cite{HN2021} or of \emph{stochastic NFs} \cite{hagemann2022stochastic,HHS2023,wu2020}.

\subsection{Adversarial Local Regularizers (ALR)} \label{subsec:ALR}
The adversarial local regularizer (ALR) proposed by Prost et al. \cite{prost2021learninglocalar}  makes use of a discriminative model. 
Originally, the ALR was not formulated with a loss of the form \eqref{ML}, 
but by an adversarial approach 
similar to  Wasserstein generative adversarial networks (WGANs) \cite{arjovsky2017wasserstein}.
The basic idea goes back to Lunz et al. \cite{lunz2018adversariallearnedreg}, who suggested learning regularizers through corrupted data. More precisely, the regularizer is a  neural network trained to discriminate between the distribution of ground truth images and the distribution of unregularized reconstructions. 
The ALR is based on the same idea, but it operates on patches instead of whole images. 
Here a \emph{discriminator} $\text{D}_\theta$ between unpaired samples from the original patch distribution $P_X$ and a degraded one, say $P_{\tilde X}$, is trained using the Wasserstein-1 distance.
Conveniently, the Wasserstein-1 distance has the dual formulation 
\begin{equation*}
    \operatorname{W}_1(P_X, P_{\tilde X}) 
    = \sup \limits_{f \in \operatorname{Lip}_1}
        \left\{
        \mathbb{E}_{x \sim P_X} \left[f(x)\right] - 
        \mathbb{E}_{\tilde x \sim P_{\tilde X}}\left[f(\tilde x)\right]\right\},
\end{equation*}
where  $\operatorname{Lip}_1$ denotes the set of all Lipschitz continuous functions on $\R^d$ with Lipschitz constant not larger than 1.
A maximizing function is called optimal \emph{Kantorovich potential} and can be considered as a good separation between the two distributions. 
Unfortunately, obtaining such a potential is computationally intractable. 
Nevertheless, it can be approximated by functions from a parameterized family $\mathcal F$
as, e.g., neural networks 
with a fixed architecture
\begin{equation}
     \argmax_{\text{D}_\theta \in \mathcal F\cap  \text{Lip}_1} 
     \left\{ \mathbb{E}_{x \sim P_X} \big[\text{D}_\theta (x)\big] 
     - \mathbb{E}_{\tilde x \sim P_{\tilde X}}\left[\text{D}_\theta (\tilde x )\right]\right\}.
\end{equation}
One possibility to relax the Lipschitz condition is the addition of a gradient penalty, see \cite{gulrajani2017improved}, to arrive at
\begin{equation}
    \hat \theta 
    = \argmax_{\theta}  
    \Big\{
    \mathbb{E}_{x \sim P_X} \big[ \text{D}_\theta(x)\big]
    - 
   \mathbb{E}_{\tilde x \sim P_{\tilde X}}\big[\text{D}_\theta(\tilde x)\big] 
   - \lambda \mathbb{E}_{x \sim P_{\alpha X + (1-\alpha) \tilde X}}
   \big[\big(\|\nabla \text{D}_\theta(x)\|-1\big)^2 \big] \Big\}, \quad \lambda > 0,
\end{equation}
where $\alpha$ is uniformly distributed in $[0,1]$.
 This can be solved using a stochastic gradient descent algorithm. 
 Finally, to solve our inverse problem, we can use the ALR 
\begin{equation*}
  \mathcal R(x) =      \operatorname{ALR}(x) \coloneqq  \frac{1}{N} \sum \limits_{i=1}^N \text{D}_\theta\left(\operatorname{P}_{i}(x)\right).
\end{equation*}
This parameter estimation is different from the ML estimation of the previous two models, since it employs a discriminative approach.  

\begin{rem}
The original GAN architecture utilizes the Jensen–Shannon divergence \cite{GPMXWOCB2014} and can be replaced with the (forward) KL divergence \eqref{eq:kl_div}. 
This would lead to some form of maximum likelihood estimation in a discriminative setting. Furthermore, GAN architectures within an explicit maximum likelihood framework exist \cite{grover2018flow}.
On closer inspection, however, this still takes a similar form as the EPLL or the patchNR. We assign a value to each patch in an image and sum over the set of resulting values. Moreover, higher values are assigned to patches that are more likely to stem from the true patch distribution. 

\end{rem}
\section{Divergences between Empirical Patch Measures}\label{sec:OT}
In the previous section, we have constructed patch-based regularizers using sums over all patches within an ML approach.
This calls for independently drawn patches from the underlying distribution, an assumption that does not hold true, e.g., for overlapping patches. 
In particular, the same value may be assigned for an image which is a combination of very likely and very unlikely patches.
This makes it desirable to address the patch distribution as a whole by 
assigning empirical measures to the patches as in \eqref{e-measures}.
In the following, we will consider three different ``distances'' between these empirical measures,
namely the Wasserstein-2 distance, the regularized Wasserstein-2 distance, and 
an unbalanced variant.

\subsection{Wasserstein Patch Prior} \label{subsec:WPP}

\begin{figure}[!t] 
\captionsetup[subfigure]{font=normal,justification=centering}
     \centering
     \begin{subfigure}[b]{0.3\textwidth}
         \centering
         \includegraphics[width=\textwidth]{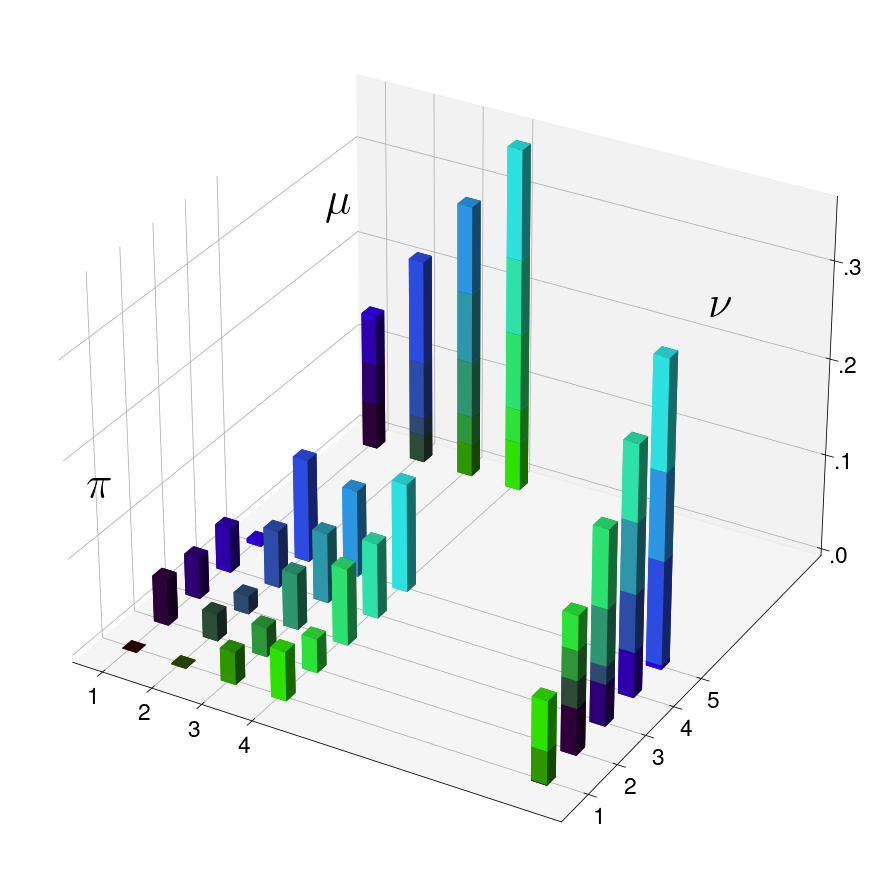}
         \caption*{Cost: $3.335$}
     \end{subfigure}
     \hfill
     \begin{subfigure}[b]{0.3\textwidth}
         \centering
         \includegraphics[width=\textwidth]{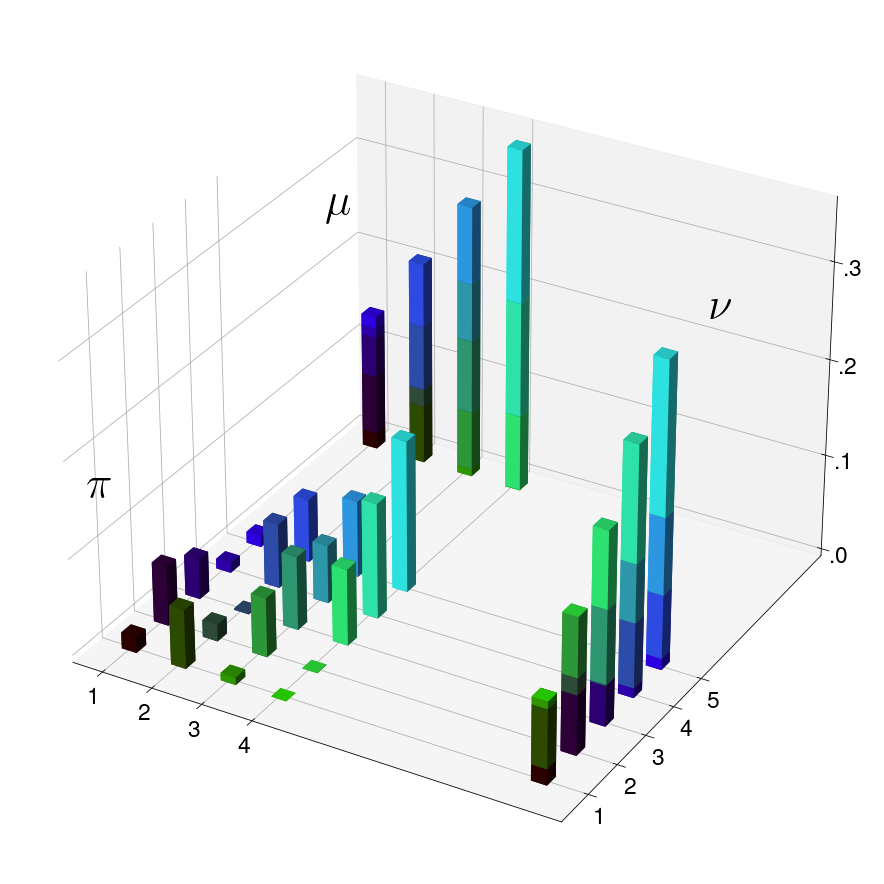}
         \caption*{Cost: $2.145$}
     \end{subfigure}
     \hfill
     \begin{subfigure}[b]{0.3\textwidth}
         \centering
         \includegraphics[width=\textwidth]{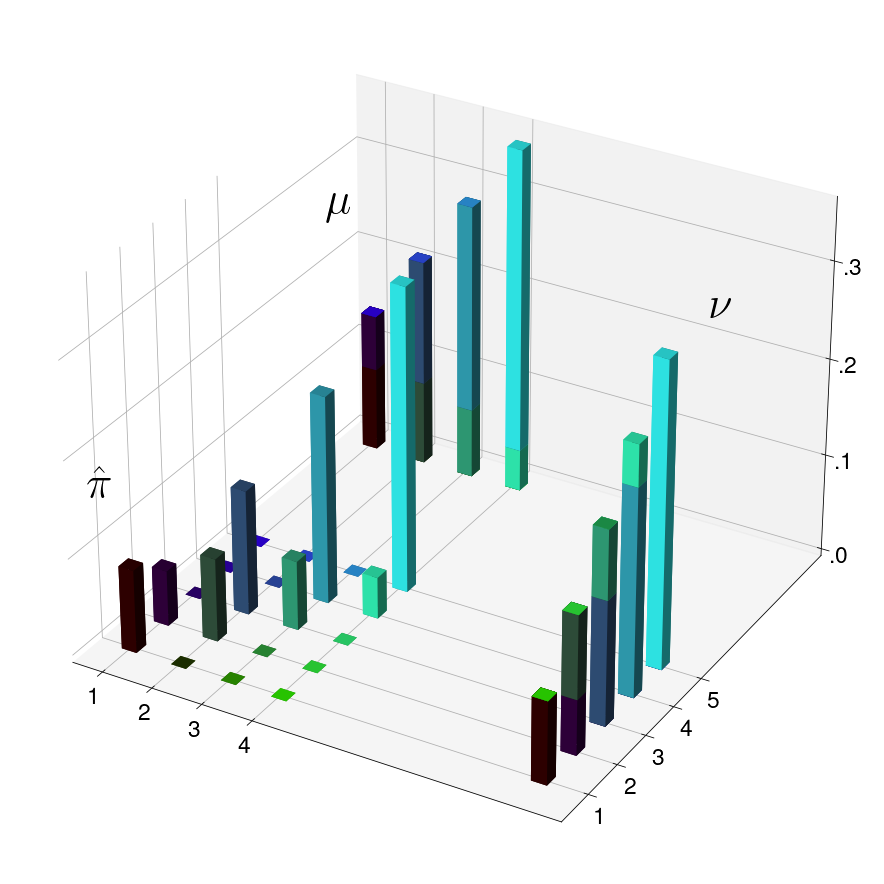}
         \caption*{Cost: $\text{W}_2^2(\mu,\nu)= 0.714$}
     \end{subfigure}
     \caption{Admissible couplings between
     $\mu = \sum_{i=1}^4 \mu_i \delta_i = \frac{2}{14}\delta_{1} + \frac{3}{14}\delta_{2}  + \frac{4}{14} \delta_{3} + \frac{5}{14}\delta_{4}$ 
     and 
     $\nu = \sum_{j=1}^5 \nu_j \delta_j =
     \frac{3}{35} \delta_{1} + \frac{5}{35} \delta_{2}  + \frac{7}{35}\delta_{3} + \frac{9}{35}\delta_{4} + \frac{11}{35}\delta_{5}$ (weights are visualized as  bars). Any admissible coupling takes the form 
     $\pi = \sum_{i,j=1}^{4,5}  \pi_{i, j} \delta_{(i, j)}$, 
     where $\sum_{i=1}^{4}  \pi_{i, j} = \nu_j$ and $\sum_{j=1}^{5}  \pi_{i, j} = \mu_i$.
     The weights of three admissible couplings are visualized.  The Wasserstein-2 distance is characterized by those $\pi$ which minimize the $\text{costs} = \sum_{i,j=1}^{4,5} \pi_{i, j} (i - j)^2$.  
     The squared Wasserstein-2 distance has the sparsest coupling.
     }
     \label{optimal_transport_plan}
\end{figure}

To keep the notation simple, let us rewrite
the empirical measures in  \eqref{e-measures} as 
\begin{equation} 
    \nu  = \frac{1}{nN} \sum_{j=1}^n \sum_{i=1}^N \delta_{\text{P}_i(x_j)}  
    = \frac{1}{M} \sum_{k=1}^M \delta_{y_k}
    \quad \text{and} \quad
    \mu_x  = \frac1N\sum_{i=1}^N \delta_{\text{P}_i(x)} = \frac1N\sum_{i=1}^N  \delta_{x_i}.
    \end{equation}
Then the admissible plans in the Wasserstein-2 distance \eqref{wdist}
have the form
$$
\pi = \sum_{i=1}^N \sum_{k=1}^M p_{i,k} \delta_{i,k}, \qquad 
\sum_{i=1}^N p_{i,k} = \frac{1}{M}, \; \sum_{k=1}^M p_{i,k} = \frac{1}{N}, \;
i=1,\ldots,N, k=1,\ldots,M.
$$
Obviously, they are determined by the weight matrix $\mathbf{\pi} \coloneqq (\pi_{i,k})_{i,k=1}^{N,M}$. Then,  with the cost matrix   
$C\coloneqq (\|x_i-y_k\|^2)_{i,k=1}^{N,M}$, 
the Wasserstein-2 distance becomes
\begin{equation} \label{w2_primal}
    \operatorname{W}^2_2(\mu_x, \nu) 
    = 
    \min_{\mathbf{\pi} \in  \Pi } \, 
     \langle C,  \mathbf{\pi}\rangle, \quad  
     \Pi = \left\{\mathbf{\pi} \in \R_+^{N,M}:  
     \mathbbm{1}_N ^\tT \mathbf{\pi} = \frac1M \mathbbm{1}_M, \mathbf{\pi} \mathbbm{1}_M = \frac1N \mathbbm{1}_N\right\}.
  \end{equation}
  Here $\mathbbm{1}_M \in \R^M$ denotes the vector with all entries one.
  An example of the Wasserstein-2 distance for two discrete measures is given in Figure \ref{optimal_transport_plan}.
  The dual formulation of the linear optimization problem \eqref{w2_primal} reads as
 \begin{align} \label{w2_dual}
    \operatorname{W}^2_2(\mu_x, \nu) 
    &= \max_{ \phi(x_i) + \psi_k \le c_{i,k}}
    \frac{1}{N} 
    \sum \limits_{i=1}^{N} \phi(x_i) + \frac{1}{M} \sum \limits_{k=1}^{M} \psi_k\\
    &=    \max_{\psi \in \R^{M}} \frac{1}{N} 
    \sum \limits_{i=1}^{N} \psi^c\left( x_i\right) + \frac{1}{M} \sum \limits_{k=1}^{M}  \psi_{k} 
\end{align}
with the \emph{$c$-conjugate function}
\begin{equation} \label{mini}
\psi^c\left( x_i\right) \coloneqq \min_{k} \big\{\| x_i - y_k\|^2 - \psi_{k}\big\},
\end{equation}
see \cite{santambrogio2015optimal}.
The maximization problem \eqref{w2_dual} is concave, and for large-scale problems, a gradient ascent algorithm as in \cite{genevay2016stochastic} can be used to find a global maximizer $\hat \psi$.
As in \cite{altekruger2023wppnets,hertrich2022wasserstein}, the optimal vector $\hat \psi$  allows for the computation of the gradient 
of 
\begin{equation}\label{rwpp}
\mathcal R(x) = \text{WPP(x)}
\coloneqq
\operatorname{W}^2_2(\mu_x, \nu)
\end{equation}
in our inverse problem \eqref{inv_reg}.
More precisely, with the minimizer 
$\sigma{(i)} \in \argmin_k \big\{\| x_i - y_k\|^p - \hat \psi_{k}\big\}$
in \eqref{mini} 
we obtain
\begin{align*}
\operatorname{W}^2_2(\mu_x, \nu)
= \frac{1}{N}
\sum \limits_{i=1}^{N}\|x_i - y_{\sigma(i)}\|^2 - \hat \psi_{\sigma(i)} + \frac{1}{M} \sum \limits_{k=1}^{M}  \hat \psi_{k},
\end{align*}
so that if the gradient with regard to the support point $x_i$ of $\mu_x$ exists, it reads as 
\begin{equation}
\label{eq:wasserstein_grad}
\nabla_{x_i} \operatorname{W}^2_2(\mu_x, \nu) = \frac{1}{N}\nabla_{x_i} \|x_i - y_{\sigma{(i)}}\|_2^2 =
\frac{2}{N}(x_i - y_{\sigma{(i)}}).
\end{equation}

\begin{rem} \label{rem:prior_WPP}
By the relation \eqref{eq:prior_reg} the WPP defines a prior distribution $p_X (x) = C_\beta \exp(-\beta \text{WPP}(x))$. The integrability of the function $p_X$ is shown in \cite[Prop. 4.1]{altekruger2023wppnets}. 
\end{rem}
\begin{rem}
Wasserstein patch priors were originally introduced by Gutierrez et al.~\cite{GRGH2017} and Houdard et al.~\cite{houdard2021wasserstein} for texture generation, where a
direct minimization of the regularizer without a data fidelity term was used. Their use was adopted
for regularization in inverse problems by Hertrich et al. \cite{hertrich2022wasserstein}.
Note that this stands in contrast to the previous EPLL-based regularizers, where a direct minimization of these regularizers would result in the synthesis of images with almost equally likely patches. In practice, this would lead to single-color images.
\end{rem}
\subsection{Sinkhorn Patch Prior} \label{subsec:Sinkhorn}

To lower the computational burden in the WPP approach, a combination of the Wasserstein distance with the KL of the coupling and the 
product measure $\mu_x \otimes \nu$ can be used
\begin{align}
\operatorname{W}_{2,\varepsilon}^2(\mu_x, \nu) 
&= \inf_{\pi \in \Pi(\mu_x, \nu) } \int \limits_{\R^d \times \R^d} \|x-y\|^2 \,d \pi(x, y) + \varepsilon \text{KL}(\pi , \mu_x \otimes \nu)\\
&= \inf_{\mathbf{\pi} \in \Pi(\mu_x, \nu)} 
 \langle C,  \mathbf{\pi}\rangle +\varepsilon \sum \limits_{i,k=1}^{N,M} \pi_{i, k} \log \left(MN \pi_{i, k}\right).
\end{align}

\begin{figure}[!t] 
\captionsetup[subfigure]{font=normal,justification=centering}
     \centering
      \begin{subfigure}[b]{0.3\textwidth}
         \centering
         \includegraphics[width=\textwidth]{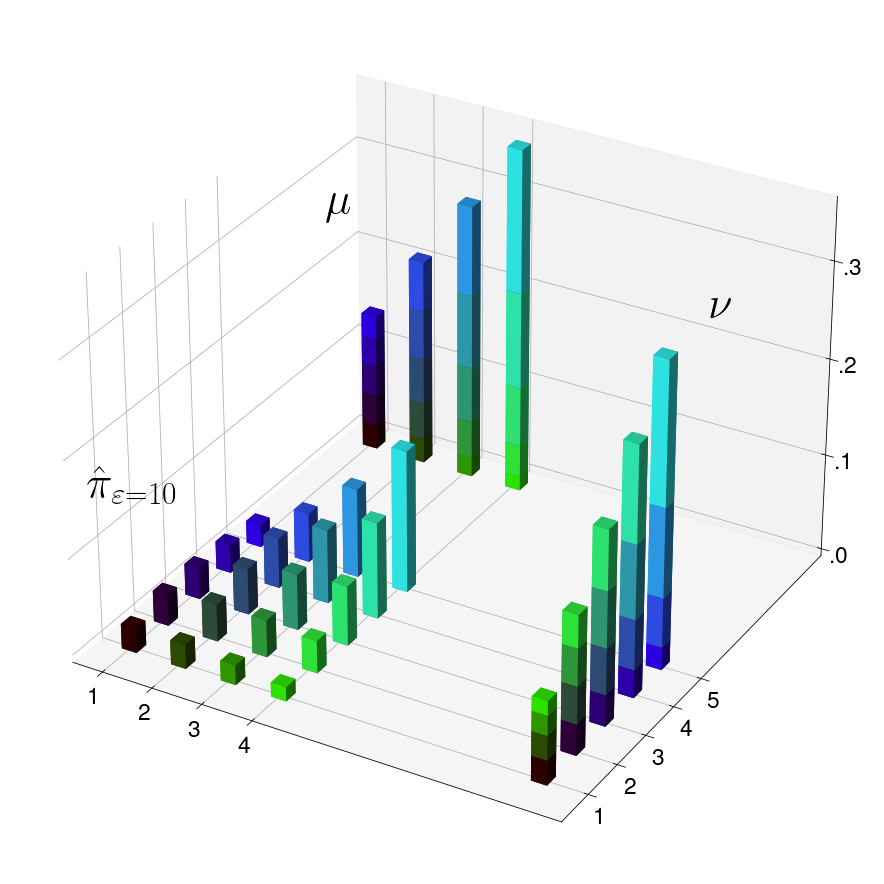}
         \caption*{$\operatorname{W}^2_{2,10}(\mu, \nu) = 2.935$}
     \end{subfigure}
     \hfill
     \begin{subfigure}[b]{0.3\textwidth}
         \centering
         \includegraphics[width=\textwidth]{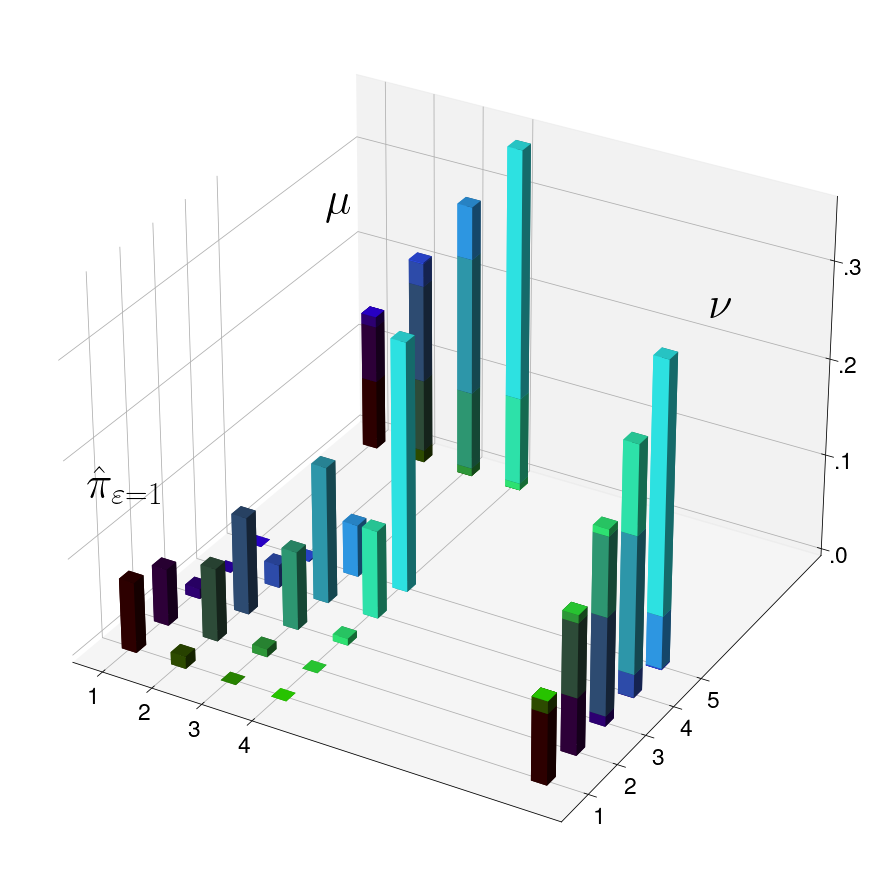}
         \caption*{$\operatorname{W}^2_{2,1}(\mu, \nu) = 1.544$}
     \end{subfigure}
     \hfill
     \begin{subfigure}[b]{0.3\textwidth}
         \centering
         \includegraphics[width=\textwidth]{images/ot_viz/discrete_measure_evot_evot.png}
         \caption*{$\operatorname{W}^2_{2}(\mu, \nu) =0.714$}
     \end{subfigure}
     \caption{Optimal couplings of $\operatorname{W}_{2,\varepsilon}$ for $\varepsilon=10, 1, 0$
     and the measures from Figure \ref{optimal_transport_plan}. With decreasing $\varepsilon$ the coupling matrices become sparser.}
     \label{transport_plans_reg}
\end{figure}

In Figure~\ref{transport_plans_reg} we give an example of $W_{2,\varepsilon}^2$ for different choices of $\varepsilon$ and the same discrete measures as in Figure~\ref{optimal_transport_plan}. 
The dual formulation reads as 
\begin{equation}\label{sink_dual}
    \operatorname{W}^2_{2,\varepsilon}(\mu_x, \nu) = 
    \max_{\phi \in \R^{N}, \psi \in \R^{M}} 
    \frac{1}{N} \sum \limits_{i=1}^{N} \phi_i 
    + \frac{1}{M}  \sum \limits_{k=1}^{M} \psi_{k} 
    -  \frac{\varepsilon}{MN} \sum \limits_{i=1}^{N} \sum \limits_{k=1}^{M} \exp \left(\frac{\phi_i  + \psi_k - \|x_i - y_k\|^2}{\varepsilon}\right) + \varepsilon.
\end{equation}
This problem can be efficiently solved using the \emph{Sinkhorn algorithm} which employs a fixed-point iteration. To this end, fix $\psi^{(r)}$, respectively,
$\phi^{(r)}$ and set the gradient with respect to the other variable in \eqref{sink_dual} to zero. This results in the iterations
\begin{align*}
    \phi^{(r+1)}_i &= 
    -\varepsilon 
    \log \Big(\sum_{k=1}^M
    \exp \Big(  \tfrac{\psi^{(r)}_k- \|x_i - y_k\|^2}{\varepsilon}\Big) \Big) + \varepsilon \log M,\\
    \psi^{(r+1)}_k &= -\varepsilon 
    \log \Big(\sum_{i=1}^N
    \exp \Big(  \tfrac{\phi^{(r)}_i- \|x_i - y_k\|^2}{\varepsilon}\Big) \Big) + \varepsilon \log N, 
\end{align*}
which converge linearly to the fixed points $\hat \phi$ and $\hat \psi$, see, e.g., \cite{feydy2019interpolating}.
Then, noting that by construction of $\hat \phi$ and $\hat \psi$ we have 
\begin{equation}\label{eq:vanishingentropy}
    - \frac{\varepsilon}{M N} \sum \limits_{i=1}^{N} \sum \limits_{k=1}^{M} \exp \left(\frac{
     \hat \phi_i+ \hat \psi_k - \|x_i - y_k\|^2}{\varepsilon}\right) + \varepsilon = 0,
\end{equation}
the regularized Wasserstein distance becomes
\begin{align}\label{eq:regwasserstein_plugin}
\operatorname{W}^2_{2,\varepsilon}(\mu_x, \nu)
&= 
- \frac{\varepsilon}{N} \Big( \sum \limits_{i=1}^{N}      
    \log \Big(\sum_{k=1}^M
    \exp \Big(  \tfrac{\hat \psi_k- \|x_i - y_k\|^2}{\varepsilon}\Big) \Big) - \log M
    \Big)   + \frac{1}{M}  \sum \limits_{k=1}^{M} \hat \psi_{k}, 
\end{align}
which is differentiable with respect to the support points and the gradient is given by 
\begin{equation}\label{eq:sinkhorn_grad}
\nabla_{x_i} \operatorname{W}^2_{2,\varepsilon}(\mu_x, \nu) 
= \frac{2}{N} \Big(\sum_{k=1}^M
    \exp \Big(  \tfrac{\hat \psi_k- \|x_i - y_k\|^2}{\varepsilon}\Big) \Big)^{-1} \sum_{k=1}^M \exp \Big(  \tfrac{\hat \psi_k- \|x_i - y_k\|^2}{\varepsilon}\Big) (x_i-y_k).
\end{equation}
If the Wasserstein gradient from \eqref{eq:wasserstein_grad} exists, it is recovered for $\varepsilon \to 0$. Computation of the gradient can, e.g., be achieved by means of algorithmic differentiation through the Sinkhorn iterations or on the basis of the optimal dual potentials through the Sinkhorn algorithm.
Finally, we can use the Sinkhorn patch prior ($\text{WPP}_\varepsilon$) first used in \cite{mignon2023semi} as a regularizer
in our inverse problem
\begin{equation*}
    \mathcal{R}(x) 
    = \operatorname{WPP}_{\varepsilon}(x) \coloneqq \operatorname{W}^2_{2,\varepsilon} (\mu_x, \nu).
\end{equation*}

\begin{rem} \label{rem:prior_Sinkhorn}
By \eqref{eq:prior_reg} the Sinkhorn regularizer defines a prior distribution $p_X (x) = C_\beta \exp(-\beta \text{WPP}_\varepsilon(x))$. The integrability of the function $p_X$ follows from the integrability of the WPP \cite[Prop. 4.1]{altekruger2023wppnets} and the relation $\text{WPP}_\varepsilon(x) \ge \text{WPP}(x)$. This can be seen immediately since  $\text{KL}(\pi , \mu_x \otimes \nu) \ge 0$.
\end{rem}
\begin{rem}
The regularized Wasserstein-2 distance is no longer a distance. 
It does not fulfill the triangular inequality and is moreover biased,
i.e., $W_{2,\varepsilon}(\mu,\nu)$  does not take its smallest value if and only if $\mu = \nu$.
As a remedy, the debiased regularized Wasserstein distance or \emph{Sinkhorn divergence}
\begin{equation*}    \operatorname{S}^2_{2,\varepsilon}(\mu, \nu) = 
    \operatorname{W}^2_{2,\varepsilon, }(\mu, \nu) 
    - 
    \frac12 \operatorname{W}^2_{2,\varepsilon}(\mu, \mu) 
    - 
    \frac12 \operatorname{W}^2_{\varepsilon}(\nu, \nu)
\end{equation*}
can be used, which is now indeed a statistical distance. Computation with the Sinkhorn divergence is similar to above so it can be used as a regularizer as well. 
For more information see \cite{genevay2018learning,NS2021}.
\end{rem}

\subsection{Semi-Unbalanced Sinkhorn Patch Prior} \label{subsec:UOT}

\begin{figure}[!t] 
\captionsetup[subfigure]{font=normal,justification=centering}
     \centering
          \begin{subfigure}[b]{0.3\textwidth}
         \centering
         \includegraphics[width=\textwidth]{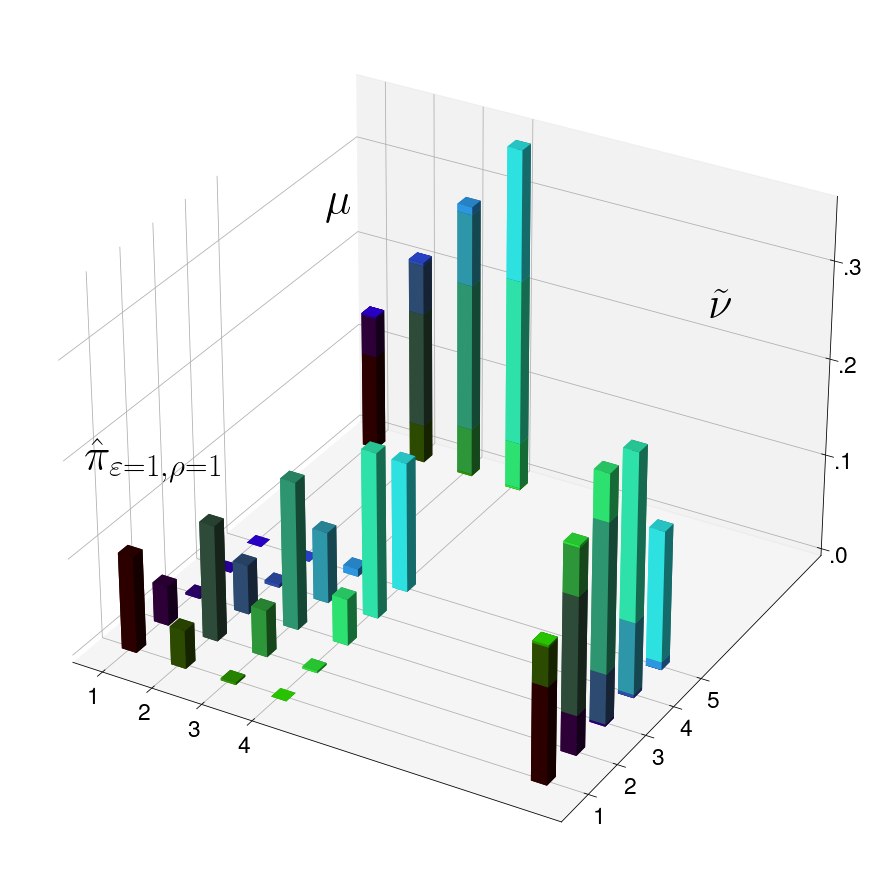}
         \caption*{$\operatorname{W}^2_{2,  1, 1}(\mu, \nu) = 1.272$}
     \end{subfigure}
     \hfill
     \begin{subfigure}[b]{0.3\textwidth}
         \centering
         \includegraphics[width=\textwidth]{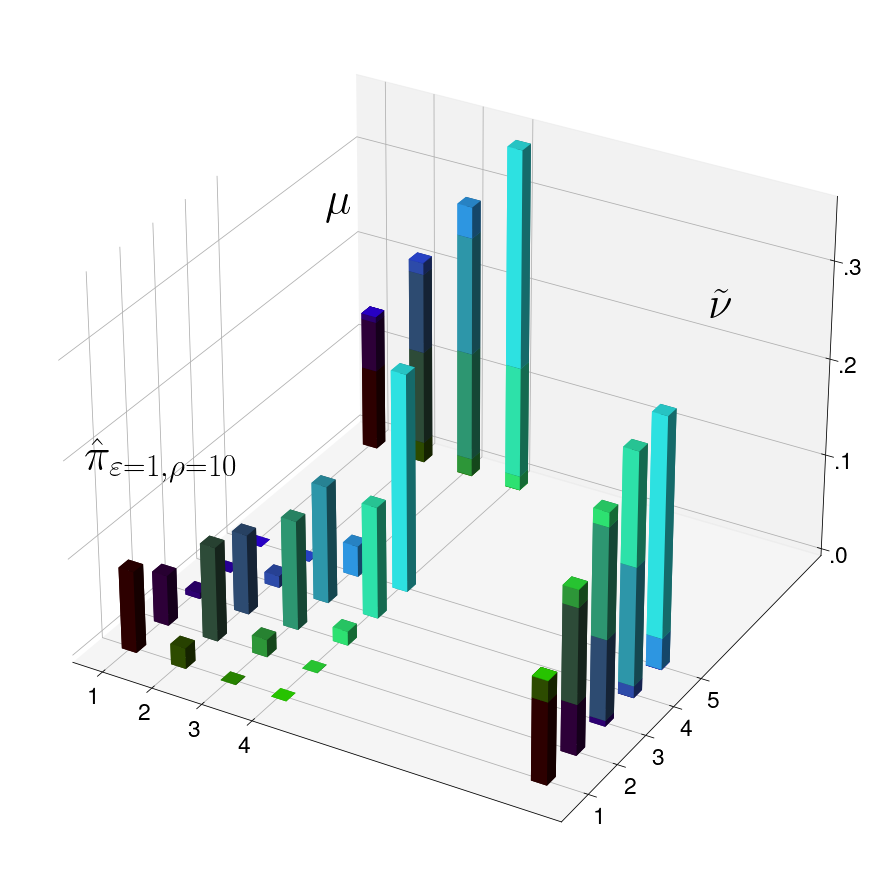}
         \caption*{$\operatorname{W}^2_{2, 1, 10}(\mu, \nu) = 1.453$}
     \end{subfigure}
     \hfill
     \begin{subfigure}[b]{0.3\textwidth}
         \centering
         \includegraphics[width=\textwidth]{images/ot_viz/discrete_measure_evot_eps=1.0_evot_rho=inf.png}
        \caption*{$\operatorname{W}^2_{2, 1}(\mu, \nu) = 1.544$}
     \end{subfigure}
     \caption{Optimal couplings of $\operatorname{W}_{2,\varepsilon,\rho}$ for 
     $\varepsilon =1$ and
     $\rho=1, 10, \infty$ and the measures from Figure~\ref{optimal_transport_plan}. The marginal $\tilde{\nu}$ (left, middle) of the coupling matrix is only an approximation of the original measure $\nu$ (right). Note the gradually decreased probability mass placed on $5$ for $\tilde{\nu}$. By increasing the balancing parameter, we move the approximation $\tilde{\nu}$ towards $\nu$ (left to middle).}
     \label{uot_transport_plans_reg}
\end{figure}

The optimal transport framework for regularizing the patch distribution was extended  by Mignon et al. \cite{mignon2023semi} 
to the \emph{semi-unbalanced case}, where the marginal of the coupling only approximates the target distribution for
\begin{align*}
\operatorname{W}_{2, \varepsilon, \rho}^2(\mu_x, \nu) 
&= \inf_{\substack{\pi \in \mathcal{P}(\R^d \times \R^d)\\
(\text{proj}_1)_\# \pi = \mu_x}} \,\,
\int \limits_{\R^d \times \R^d} \|x-y\|^2 \, \text{d} \pi(x, y) 
+ \varepsilon \text{KL} (\pi , \mu_x \otimes \nu) + \rho \text{KL}((\text{proj}_2)_\# \pi  , \nu)\\
&= 
\inf_{\substack{\mathbf{\pi} \in \R^{N, M}\\
\mathbf{\pi} \mathbbm{1}_M = \frac1N \mathbbm{1}_N}}
 \,\,\langle C,  \mathbf{\pi}\rangle + \varepsilon \sum \limits_{i,k=1}^{N,M} \pi_{i, k} \log \left(MN \pi_{i, k}\right) + \rho \sum \limits_{k=1}^{M} \left(\mathbbm{1}_N ^\tT \mathbf{\pi}\right)_{k} \log \left(M\left(\mathbbm{1}_N ^\tT \mathbf{\pi}\right)_k \right).
\end{align*}
In this setting, probability mass can be added to $\nu$ or removed from $\nu$. This behavior
is controlled by the parameter $\rho$ and leads to a decreased sensitivity with regard to isolated areas in the second distribution.  
An example of $W_{2,\varepsilon,\rho}^2$ for different choices of $\rho$ and the same measures as in Figure~\ref{optimal_transport_plan} is given in Figure~\ref{uot_transport_plans_reg}.
The dual formulation becomes 
\begin{equation*}
    \operatorname{W}^2_{2,\varepsilon, \rho}(\mu_x, \nu) \hspace{-.2em} =  \hspace{-.2em}
    \max_{\substack{\phi \in \R^{M}\\ \psi \in \R^{M}}} \hspace{-.2em}
    \frac{1}{N} \sum \limits_{i=1}^{N} \phi_i
    + \frac{1}{M} \sum \limits_{k=1}^{M} \rho \left(\exp\left(\frac{\psi_k}{\rho}\right)-1\right)      
    -  \frac{\varepsilon}{M N} \sum \limits_{i=1}^{N} \sum \limits_{k=1}^{M} \exp \left(\frac{\phi_i  + \psi_k - \|x_i - y_k\|^2}{\varepsilon}\right) + \varepsilon,
\end{equation*}
see, e.g., \cite{mignon2023semi}. 
This maximization problem can be solved by the following adapted Sinkhorn iterations
\begin{align*}
    \phi^{(r+1)}_i  &= 
    -\varepsilon 
    \log \Big(\sum_{k=1}^M
    \exp \Big(  \tfrac{\psi^{(r)}_k- \|x_i - y_k\|^2}{\varepsilon}\Big) \Big) + \varepsilon \log M, \\
    \psi^{(r+1)}_k &= - \frac{\varepsilon \rho}{\rho + \varepsilon}
    \log \Big(\sum_{i=1}^N
    \exp \Big(  \tfrac{\phi^{(r)}_i- \|x_i - y_k\|^2}{\varepsilon}\Big) \Big) + \varepsilon \log N.
\end{align*}
Note that the fixed point $\hat \phi$ equals the fixed point from Section~\ref{subsec:Sinkhorn} and consequently 

$\hat \phi$ and $\hat \psi$ fulfill \eqref{eq:vanishingentropy}. The semi-unbalanced regularized Wasserstein distance becomes
\begin{align*}
\operatorname{W}^2_{2,\varepsilon, \rho}(\mu_x, \nu)
&= 
- \frac{\varepsilon}{N} \Big( \sum \limits_{i=1}^{N}      
    \log \Big(\sum_{k=1}^M
    \exp \Big(  \tfrac{\hat \psi_k- \|x_i - y_k\|^2}{\varepsilon}\Big) \Big) - \log M
    \Big)   + \frac{1}{M} \sum \limits_{k=1}^{M} \rho \left(\exp\left(\frac{\hat \psi_k}{\rho}\right)-1\right). 
\end{align*}
This expression equals the expression \eqref{eq:regwasserstein_plugin} up to the second term, which does not depend on the support points. As a result, the gradient takes the same form as in \eqref{eq:sinkhorn_grad}, but for a $\hat \psi$ depending on $\rho$. For $\rho \to \infty$ we recover the balanced formulation and hence the gradient from \eqref{eq:sinkhorn_grad}.
Finally, we can use a semi-unbalanced Sinkhorn patch prior ($\text{WPP}_{\varepsilon,\rho}$) defined as
\begin{equation*}
    \mathcal{R}(x) =
    \operatorname{WPP}_{\varepsilon, \rho}(x) \coloneqq \operatorname{W}^2_{2,\varepsilon, \rho}
    \left(\mu_x, \nu\right).
\end{equation*}
This was proposed as an extension of the WPP in \cite{mignon2023semi}.

\begin{rem} \label{rem:prior_uot}
By the relation \eqref{eq:prior_reg} the $\text{WPP}_{\varepsilon,\rho}$ defines a prior distribution $p_X (x) = C_\beta \exp(-\beta \text{WPP}_{\varepsilon,\rho}(x))$. This can be seen as follows:
Using the auxiliary variable $\tilde \nu = (\text{proj}_2)_\# \pi$ we rewrite $\operatorname{W}^2_{2,\varepsilon, \rho}(\mu_x, \nu)$ by
\begin{align*}
\inf_{\substack{\pi \in \Pi(\mu_x,\tilde \nu)\\
\text{supp} (\tilde \nu)  \subseteq \text{supp}(\nu)}} \,
\operatorname{W}^2_2(\mu_x, \tilde \nu) 
+ \varepsilon \text{KL} (\pi , \mu_x \otimes \nu) + \rho \text{KL}(\tilde \nu , \nu) 
\ge \inf_{\substack{\pi \in \Pi(\mu_x,\tilde \nu)\\
\text{supp} (\tilde \nu)  \subseteq \text{supp}(\nu)}} \,
\operatorname{W}^2_2(\mu_x, \tilde \nu).
\end{align*}
The constraint $\text{supp} (\tilde \nu)  \subseteq \text{supp}(\nu)$ is due to the term $\text{KL}(\tilde \nu , \nu)$ which otherwise would be infinite. Exploiting the discrete structure $\nu = \frac{1}{M} \sum_{k=1}^M \delta_{y_k}$, such a measure $\tilde \nu$ needs to be of the form $\tilde \nu = \sum_{k=1}^M a_k \delta_{y_k}$, for $a \in \R_+^M$ with $\sum_{k=1}^M a_k = 1$. The dual formulation \eqref{w2_dual} yields
\begin{align*}
\inf_{\substack{\pi \in \Pi(\mu_x,\tilde \nu)\\
\text{supp} (\tilde \nu)  \subseteq \text{supp}(\nu)}}
\operatorname{W}^2_2(\mu_x, \tilde \nu) = \inf_{\substack{a \in \R_+^M\\
\sum_{k=1}^M a_k = 1}} \,
\Big(\max_{\psi(a) \in \R^M} \frac{1}{N} \sum_{i=1}^N \psi(a)^c(x_i) + \sum_{k=1}^M a_k \psi(a)_k \Big)
\ge \frac{1}{N} \sum_{i=1}^N \psi_0^c(x_i),
\end{align*}
where the last inequality follows from inserting $\psi(a)  = \psi_0 = 0$ for all $a \in \R^M$. Now, the statement follows from the proof of \cite[Prop.~4.1]{altekruger2023wppnets}.
\end{rem}
\begin{rem}
By construction, the semi-unbalanced regularized Wasserstein distance is not symmetric anymore. Moreover, it is again biased.
Similarly, as for the balanced case, the semi-unbalanced regularized Wasserstein distance can be transformed into a (non-symmetric) \emph{semi-unbalanced Sinkhorn divergence}
\begin{equation*}
    \operatorname{S}^2_{2,\varepsilon, \rho}(\mu, \nu) = \operatorname{W}^2_{2,\varepsilon, \rho}(\mu, \nu) - \frac12 \operatorname{W}^2_{2,\varepsilon}(\mu, \mu) - \frac12 \operatorname{W}^2_{2,\varepsilon, \rho, \rho}(\nu, \nu)
\end{equation*}
with the fully unbalanced regularized Wasserstein distance
\begin{equation*}
 \operatorname{W}^2_{2,\varepsilon, \rho, \rho}(\mu, \nu)
    = \inf_{\pi \in \mathcal{M}^+(\R^d \times \R^d)} \hspace{-.3em}
\int \limits_{\R^d \times \R^d} \hspace{-.5em}\|x-y\|^2 \, \text{d} \pi(x, y) 
+ \varepsilon \text{KL} (\pi , \mu_x \otimes \nu) + \rho \text{KL}((\text{proj}_2)_\# \pi  , \nu) +  \rho \text{KL}((\text{proj}_1)_\# \pi  , \mu).
\end{equation*} 
Here, $\mathcal{M}^+(\R^d \times \R^d)$ denotes the set of positive measures on $\R^d \times \R^d$. For more information, see \cite{sejourne2023unbalanced}.
\end{rem}

\section{Uncertainty Quantification via Posterior Sampling} \label{sec:uncertainty}
In contrast to the MAP approaches, which just give point estimates for the most likely solution of the inverse problem, see Paragraph 1 of Section \ref{sec:inv_prob}, we want to approximate the whole posterior measure $P_{X|Y=y}$ now. More precisely, we intend to sample from the approximate posterior to get multiple possible reconstructions of the inverse problem and to quantify the uncertainty in our reconstruction. 
By Bayes' law and relation \eqref{eq:prior_reg}, we know that
$$
p_{X|Y=y}(x) \propto p_{Y|X=x}(y) p_{X}(x), \quad p_X (x) = C_\beta \exp\left(-\beta \mathcal R(x)\right).
$$
While the likelihood $p_{Y|X=x}$ is determined by the noise model and the forward operator, the idea is now to choose  a prior from the previous sections, i.e.,
\begin{equation}\label{approx}
\mathcal R \in \{\text{EPLL, patchNR, ALR, WPP,}\text{ WPP}_\varepsilon,\text{WPP}_{\varepsilon, \rho}\},
\end{equation}
By the Remarks~\ref{rem:prior_EPLL}, \ref{rem:prior_patchNR},\ref{rem:prior_WPP}, \ref{rem:prior_Sinkhorn} and \ref{rem:prior_uot} we have ensured that the corresponding functions
$p_X$ are indeed integrable, except for ALR, where this is probably not the case. Nevertheless, we will use ALR in our computations even without the theoretical foundation.  
Techniques to enforce the integrability of a given regularizer by utilizing a projection onto a compact set, e.g., $[0, 1]^d$, exist in the literature \cite{CTMLSZ2023,LBADDP2022}.

Even if the density of a distribution is known you can in general not sample from this distribution, except for the uniform and the Gaussian distribution.
Established methods for posterior sampling are Markov chain Monte Carlo (MCMC) methods such as Gibbs sampling \cite{RR2004}.
We want to focus on Langevin Monte Carlo methods \cite{N1992,RT1996, WT2011}, which have shown good performance for image applications and come with theoretical guarantees \cite{CTMLSZ2023, LBADDP2022}. In particular, in \cite{friedman2021posterior} the EPLL was used in combination with Gibbs sampling for posterior reconstruction of natural images, and in \cite{CTMLSZ2023} the patchNR was used in combination with Langevin sampling for posterior reconstruction in limited-angle CT. 

Consider the overdamped Langevin stochastic differential equation (SDE)
\begin{align} \label{eq:LSDE}
\mathrm{d} X_t = \nabla \log p_{X|Y=y}(X_t) \mathrm{d}t + \sqrt{2} \mathrm{d} B_t,
\end{align}
where $B_t$ is the $d$-dimensional Brownian motion. If $p_{X|Y=y}$ is proper, smooth and $ x \mapsto \nabla \log p_{X|Y=y}(x)$ is Lipschitz continuous, then Roberts and Tweedie \cite{RT1996} have shown that, for any initial starting point, 
the SDE \eqref{eq:LSDE} has a unique strong solution and  $p_{X|Y=y}$ is the unique stationary density.
For a discrete time approximation, the \emph{Euler-Maruyama discretization} with step size $\delta$ leads to the \emph{unadjusted Langevin algorithm} (ULA)
\begin{align} \label{eq:ULA}
X_{k+1} &= X_k + \delta \nabla \log p_{X|Y=y}(X_k) + \sqrt{2 \delta} Z_{k+1} \\
&=X_k + \delta \nabla \log p_{Y|X=X_k}(y) + \delta \nabla \log p_X(X_k) + \sqrt{2 \delta} Z_{k+1},
\end{align}
where $Z_k \sim \mathcal{N}(0,I)$, $k \in \mathbb N$. The step size $\delta$ provides control between accuracy and convergence speed. The error made due to the discretization step in \eqref{eq:ULA} can be asymptotically removed by a Metropolis-Hastings correction step \cite{RT1996}. The corresponding \emph{Metropolis-adjusted Langevin algorithm} (MALA) comes with additional computational cost and will not be considered here.
Now using an approximation  \eqref{approx} for the prior, we 
get up to an additive constant
\begin{align} \label{eq:ULA_reg}
X_{k+1} =X_k + \delta \nabla \log p_{Y|X=X_k}(y) - \delta \beta \nabla \log \mathcal{R}(X_k) + \sqrt{2 \delta} Z_{k+1}.
\end{align}
In Section~\ref{sec:posterior_inpainting}, we will use this iteration for posterior sampling in image inpainting.

\paragraph{Other methods for sampling from the posterior distribution}

Alternatively to MCMC methods, posterior sampling can be done by conditional neural networks. 
While conditional variational auto-encoders (VAEs) \cite{kingma2013auto,lim2018molecular,SLY2015} approximate the posterior distribution by learning conditional stochastic encoder and decoder networks, conditional generative adversarial networks (GANs) \cite{adler_deep,arjovsky2017wasserstein,GPMXWOCB2014,liu2021wasserstein} learn a conditional generator via adversarial training. Conditional diffusion models \cite{cond_score,song2021maximum,song2021scorebased} map the posterior distribution to an approximate Gaussian distribution and reverse the noising process for sampling from the posterior distribution. For the reverse noising process, the conditional model needs to approximate the \emph{score} $\nabla_x\log p_{X|Y=y}$. 
Conditional normalizing flows \cite{altekruger2023wppnets,graz_inc,ardizzone2019guided, winkler_cond_flows} aim to approximate the posterior distribution using diffeomorphisms. In particular, in \cite{altekruger2023wppnets} the WPP was used as the prior distribution for training the normalizing flow with the backward KL. Recently, gradient flows of the maximum mean discrepancy and the sliced Wasserstein distance were successfully used for posterior sampling \cite{DLPYL2023,HHABCS2023}.

\section{Experiments}\label{sec:exp}

\begin{figure}[!b]\captionsetup[subfigure]{font=normal,justification=centering}
\centering
\begin{subfigure}[b]{0.16\textwidth}
\centering
\includegraphics[width=\textwidth]{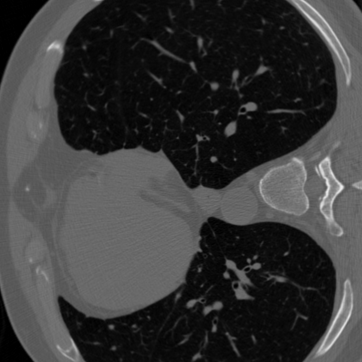}
\end{subfigure}
\hfill
\begin{subfigure}[b]{0.16\textwidth}
\centering
\includegraphics[width=\textwidth]{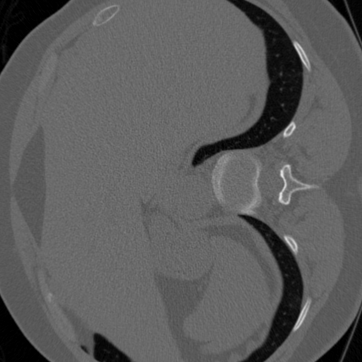}
\end{subfigure}
\hfill
\begin{subfigure}[b]{0.16\textwidth}
\centering
\includegraphics[width=\textwidth]{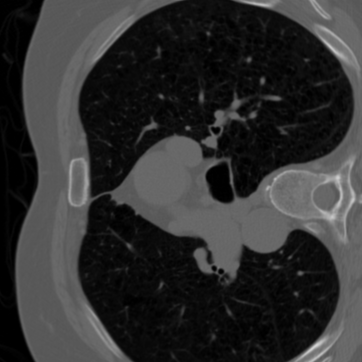}
\end{subfigure}
\hfill
\begin{subfigure}[b]{0.16\textwidth}
\centering
\includegraphics[width=\textwidth]{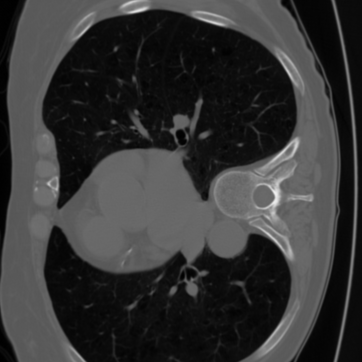}
\end{subfigure}
\hfill
\begin{subfigure}[b]{0.16\textwidth}
\centering
\includegraphics[width=\textwidth]{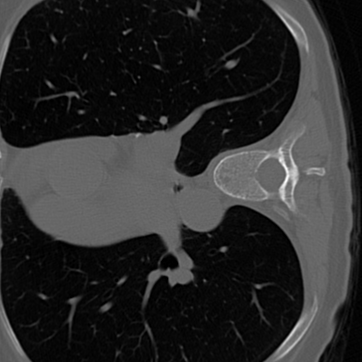}
\end{subfigure}
\hfill
\begin{subfigure}[b]{0.16\textwidth}
\centering
\includegraphics[width=\textwidth]{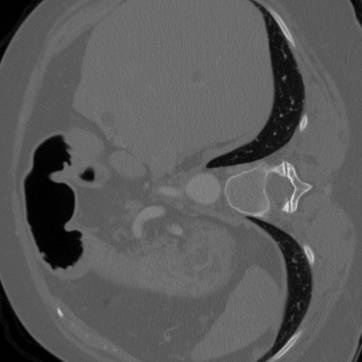}
\end{subfigure}
\caption{CT training images used to train EPLL, patchNR, and ALR as well as for reference patch distribution for WPP, $\text{WPP}_{\varepsilon}$ and $\text{WPP}_{\varepsilon, \rho}$.}
\label{fig:train_imgs_CT}
\end{figure}

 \begin{figure}[!b]\captionsetup[subfigure]{font=tiny}
     \centering
     \begin{subfigure}[b]{0.16\textwidth}
         \centering
         \includegraphics[width=\textwidth]{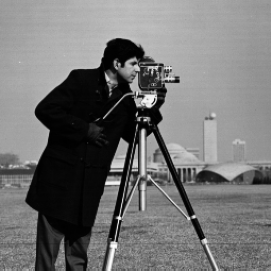}
         \caption{
         Original\\ \phantom{PSNR: 20.3}\\ \phantom{ } \\\phantom{ }
         }
         \label{dude_not_corrputed}
     \end{subfigure}
     \hfill
     \begin{subfigure}[b]{0.16\textwidth}
         \centering
         \includegraphics[width=\textwidth]{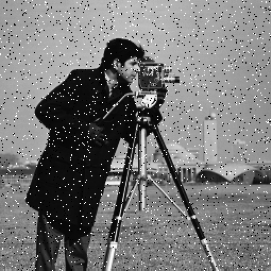}
         \caption{\textbf{PSNR: 20.3\\SSIM: 0.44\\LPIPS: 0.55\\FSIM: 0.79}}
         \label{dude_corrputed}
     \end{subfigure}
     \hfill
     \begin{subfigure}[b]{0.16\textwidth}
         \centering
         \includegraphics[width=\textwidth]{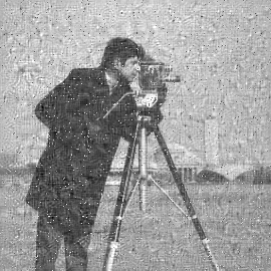}
         \caption{\textbf{PSNR: 20.3}\\SSIM: 0.35\\LPIPS: 2.27\\FSIM: 0.69}
         \label{dude_psnr_constant}
     \end{subfigure}
     \hfill
     \begin{subfigure}[b]{0.16\textwidth}
         \centering
         \includegraphics[width=\textwidth]{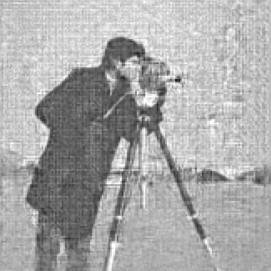}
         \caption{PSNR: 17.6\\\textbf{SSIM: 0.44}\\LPIPS: 1.14\\FSIM: 0.76}
         \label{dude_ssim_constant}
     \end{subfigure}
     \hfill
     \begin{subfigure}[b]{0.16\textwidth}
         \centering
         \includegraphics[width=\textwidth]{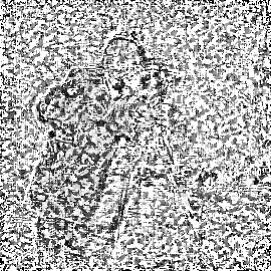}
         \caption{PSNR: 3.21\\SSIM: 0.15\\\textbf{LPIPS: 0.55}\\FSIM: 0.36}
         \label{dude_lpips_constant}
     \end{subfigure}
     \hfill
     \begin{subfigure}[b]{0.16\textwidth}
         \centering
         \includegraphics[width=\textwidth]{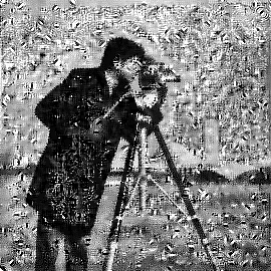}
         \caption{PSNR: 13.49\\SSIM: 0.21\\LPIPS: 2.22\\\textbf{FSIM: 0.79}}
         \label{dude_fsim_constant}
     \end{subfigure}
        \caption{Quality measures for different deteriorated images of the original image (a). For PSNR, SSIM, FSIM the largest value is best, for LPIPS the smallest one. The best values for all measures appear in image (b).
        In the other images, one of the measures is fixed.
        }
        \label{quality_measure_comp}
\end{figure}
In this section, we first use the MAP approach
$$                             
x_{\mathrm{MAP}}(y) 
\in
\argmin_{x \in \R^d} 
\Big\{ \mathcal D(F(x),y)
\, + \, \beta \mathcal R(x)
\Big\}, \quad \beta > 0.
$$
with our different regularizers on $6 \times 6$ image patches
$$
\mathcal R \in \{\text{EPLL, patchNR, ALR, WPP,}\text{ WPP}_\varepsilon,\text{WPP}_{\varepsilon, \rho}\}
$$
for solving various inverse problems. Since the data term $\mathcal D$ depends on the forward operator and the noise model, we have to describe both for each application.
We consider the following problems:
\begin{itemize}
\item computed tomography (CT) in a low-dose and a limited-angle setting,
where we learn the regularizer from just $n=6$ ``clean'' images shown in Figure~\ref{fig:train_imgs_CT}.
The transformed images are corrupted by Poisson noise.
\item super-resolution, where the regularizer is first learned from just $n=1$ ``clean'' image and second from the corrupted image. The later setting is known as zero-shot super-resolution.
Here we have a Gaussian noise model.
\item image inpainting from the corrupted image in a noise-free setting.
\end{itemize}
Second, we provide example from sampling from the posterior in image inpainting and for uncertainty quantification in computed tomography.

The code for all experiments is implemented in PyTorch and is available online\footnote{\url{https://github.com/MoePien/PatchbasedRegularizer}}. 
You can also find all hy\-per\-para\-meters in the GitHub repository. In the experiments, we minimize the variational formulation \eqref{inv_reg} using the Adam optimizer \cite{KB2015}.
The presented experimental set-up for super-resolution and computed tomography is closely related to the set-up of Altekrüger et al. \cite{altekruger2023patchnr}.  

However, before comparing these approaches, we  should give some comments on error measures
in image processing.

\subsection{Error Measures}
There does not exist an ultimate measure for the visual quality of images, since this depends heavily on the human visual perception.
Nevertheless, there are some frequently used quality measures between the original image
$x \in \R^{d_1,d_2}$ and the reconstructed, deteriorated one $\hat x$.
The \emph{peak signal-to-noise ratio} (PSNR) is defined by
\begin{equation*}
    \operatorname{PSNR}(\hat x) = 10 \cdot \log_{10} 
    \left(\frac{d_1d_2\operatorname{max}^2(x)}{\|x - \hat x\|^2} \right),
\end{equation*}
where $\operatorname{max}(x)$ denotes the highest possible pixel value of an image, e.g., $255$ for $8$ bit representations.
Unfortunately, small changes in saturation and brightness of the image have a large impact on the PSNR despite a small impact on the visual quality.
An established alternative meant to alleviate this issue is the \emph{structural similarity index} (SSIM) \cite{psnr_ssim}. It is based on a comparison of pixel means and variances of various local windows of the images. Still, this is a rather simple model for human vision and small pixel shifts heavily influence its value, see \cite{reibman2006quality}. 
Recently, the development of improved similarity metrics has revolved around the importance of low-level features for human visual impressions. 
Hence, alternative metrics focus on the comparison of extracted image features. Prominent examples include the \emph{Feature Similarity Index} (FSIM) \cite{psnr_ssim_fsim,fsim} based on hand-crafted features and the \emph{Learned Perceptual Image Patch Similarity} (LPIPS) \cite{lpips} based on the features learned by a convolutional neural network. 

All these different metrics behave very differently for image distortions as visualized in Figure~\ref{quality_measure_comp}.  The image in Figure~\ref{dude_corrputed} is obtained by corrupting the original image in Figure~\ref{dude_not_corrputed} by 5\% salt-and-pepper noise. The other images were generated with a gradient descent algorithm starting in Figure~\ref{dude_corrputed}
for customized loss functions that penalize one quality metric and deviation from the initial values for the other quality metric.
As a result, the evaluation of reconstructions depends on the chosen metrics, where a single metric may not be suitable for all problems since the requirements differ, e.g., for natural and medical images.

\subsection{Computed Tomography} \label{sec:CT}

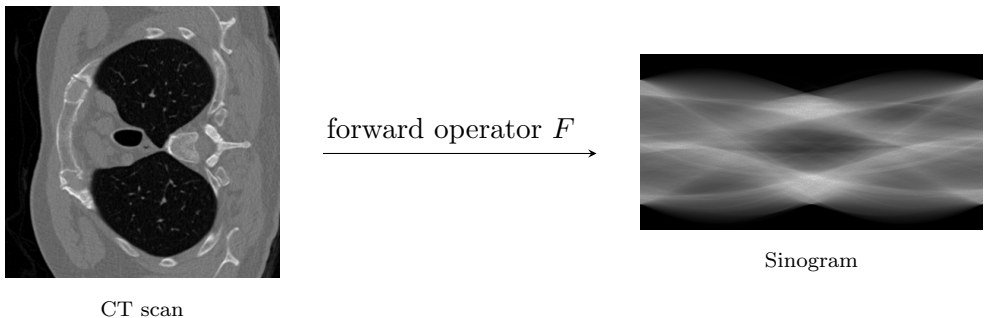
\begin{figure}[!b]
\centering
\begin{subfigure}[c]{0.2\textwidth}
\includegraphics[width=\textwidth]{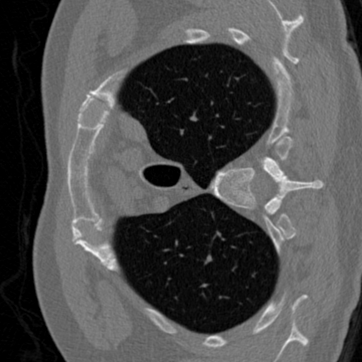}
\caption*{\scriptsize CT scan}
\end{subfigure}
\begin{subfigure}[c]{0.25\textwidth}
\centering
\vspace{-.9cm}
\begin{tikzpicture}
    \draw [-stealth](-1.8,0) --  node [text width=3.5cm,midway,above]{forward operator $F$} (1.8,0);
\end{tikzpicture}
\end{subfigure}
\begin{subfigure}[c]{0.25\textwidth}
\includegraphics[width=\textwidth]{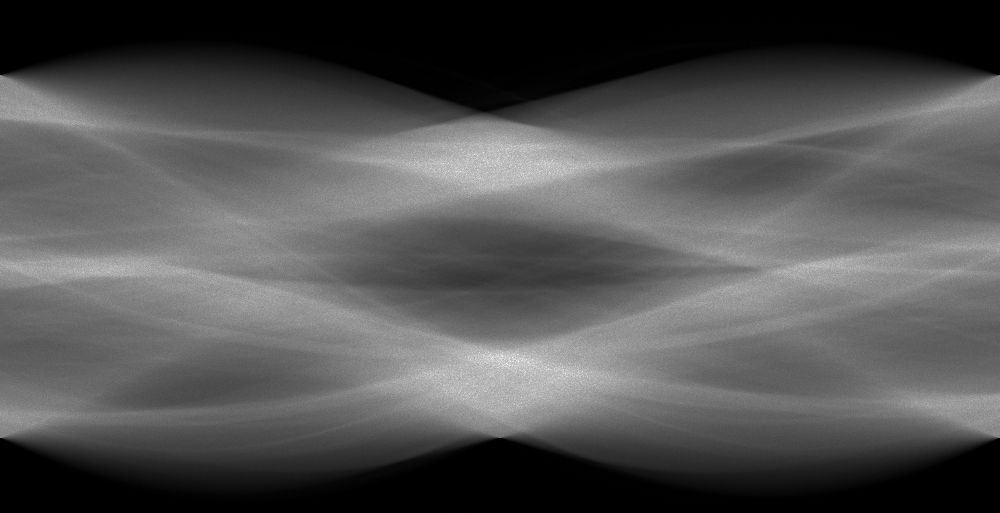}
\caption*{\scriptsize Sinogram}
\end{subfigure}
\caption{Application of the discrete Radon transformation.}
\label{fig:CT_sinogram}
\end{figure}

In CT, we want to reconstruct a CT scan from a given measurement, which is called a sinogram.
We used the LoDoPaB dataset \cite{LoDoPaB21}\footnote{available at \url{https://zenodo.org/record/3384092\#.Ylglz3VBwgM}} 
for low-dose CT imaging is used 
with images of size $362 \times 362$. 
The ground truth images are based on scans of the Lung Image Database Consortium and Image Database Resource Initiative \cite{Armato11} and the measurements are simulated. 
The LoDoPab dataset uses a two-dimensional parallel beam geometry with 513 equidistant detector bins, which results in a linear forward operator $F$ for the discretized Radon transformation. 
A CT scan and its corresponding sinogram is visualized in Figure~\ref{fig:CT_sinogram}.
The noise model follows a Poisson distribution. Recall that  $\text{Pois}(\lambda)$ has probability $p(k|\lambda) = \frac{\lambda^k \exp(-\lambda)}{k!}$ with mean (= variance) $\lambda$.
More specifically, we assume that the pixels $y_i$ are corrupted independently and we have for each pixel
\begin{align*}
Y = - \frac{1}{\mu} \log\Big(\frac{\tilde Y}{N_0} \Big), \quad \tilde{Y} \sim \text{Pois} \left( N_0 \exp \left(-F(x) \mu \right) \right),
\end{align*}
where $N_0 = 4096$ is the mean photon count per detector bin without attenuation and 
$\mu = 81.35858$ is a normalization constant. 
Then we obtain pixelwise
\begin{align*}
 \exp(-Y \mu) N_0 = \tilde{Y} \sim \text{Pois} \left( N_0 \exp \left( \left(-F(x) \mu \right) \right) \right),
\end{align*}
and consequently for the whole data term
\begin{align} \label{eq:ct_nll}
\mathcal D(F(x),y) &=- \log \prod_{i=1}^{\tilde d} p \left(
\exp(-y_i \mu) N_0| \exp \left( -F(x)_i \mu\right) N_0\right)
\\
&= 
\sum_{i=1}^{\tilde d}  \exp\left(-F(x)_i \mu\right) N_0 
+ \exp \left(-y_i \mu \right) N_0 \left( F(x)_i \mu - \log(N_0) \right).
\end{align}
For the initialization, we use the Filtered Backprojection (FBP) described by the adjoint Radon transform  \cite{Radon86}. We used the ODL implementation \cite{Adler2018} with the filter type ``Hann'' and a frequency scaling of $0.641$.

\paragraph{Low-Dose CT}

\begin{figure}\captionsetup[subfigure]{font=normal}
\centering
\begin{subfigure}[t]{.123\textwidth}  
\begin{tikzpicture}[spy using outlines=
{rectangle,white,magnification=5.5,size=2.22cm, connect spies}]
\node[anchor=south west,inner sep=0]  at (0,0) {\includegraphics[width=\linewidth]{images/CT/gt.png}};
\spy on (1.78,1.07) in node [right] at (-0.005,-1.14);
\end{tikzpicture}
\caption*{GT}
\end{subfigure}%
\hfill
\begin{subfigure}[t]{.123\textwidth}  
\begin{tikzpicture}[spy using outlines=
{rectangle,white,magnification=5.5,size=2.22cm, connect spies}]
\node[anchor=south west,inner sep=0]  at (0,0) {\includegraphics[width=\linewidth]{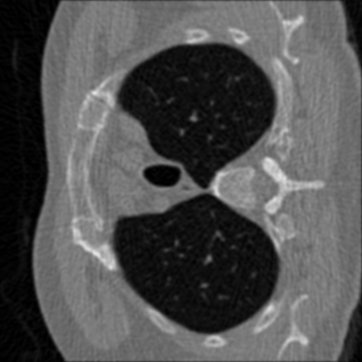}};
\spy on (1.78,1.07) in node [right] at (-0.005,-1.14);
\end{tikzpicture}
  \caption*{FBP}
\end{subfigure}%
\hfill
\begin{subfigure}[t]{.123\textwidth}  
\begin{tikzpicture}[spy using outlines=
{rectangle,white,magnification=5.5,size=2.22cm, connect spies}]
\node[anchor=south west,inner sep=0]  at (0,0) {\includegraphics[width=\linewidth]{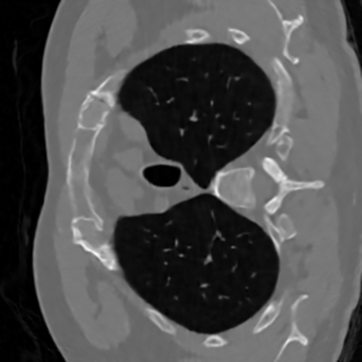}};
\spy on (1.78,1.07) in node [right] at (-0.005,-1.14);
\end{tikzpicture}
  \caption*{EPLL}
\end{subfigure}%
\hfill
\begin{subfigure}[t]{.123\textwidth}  
\begin{tikzpicture}[spy using outlines=
{rectangle,white,magnification=5.5,size=2.22cm, connect spies}]
\node[anchor=south west,inner sep=0]  at (0,0) {\includegraphics[width=\linewidth]{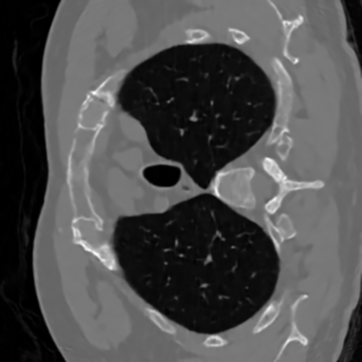}};
\spy on (1.78,1.07) in node [right] at (-0.005,-1.14);
\end{tikzpicture}
  \caption*{patchNR}
\end{subfigure}%
\hfill
\begin{subfigure}[t]{.123\textwidth}  
\begin{tikzpicture}[spy using outlines=
{rectangle,white,magnification=5.5,size=2.22cm, connect spies}]
\node[anchor=south west,inner sep=0]  at (0,0)  {\includegraphics[width=\linewidth]{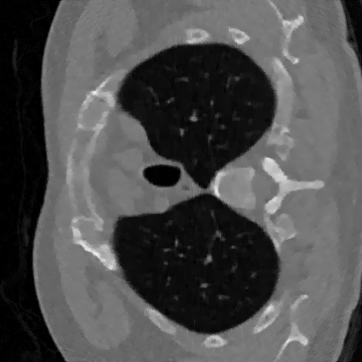}};
\spy on (1.78,1.07) in node [right] at (-0.005,-1.14);
\end{tikzpicture}
  \caption*{ALR}
\end{subfigure}%
\hfill
\begin{subfigure}[t]{.123\textwidth}  
\begin{tikzpicture}[spy using outlines=
{rectangle,white,magnification=5.5,size=2.22cm, connect spies}]
\node[anchor=south west,inner sep=0]  at (0,0) {\includegraphics[width=\linewidth]{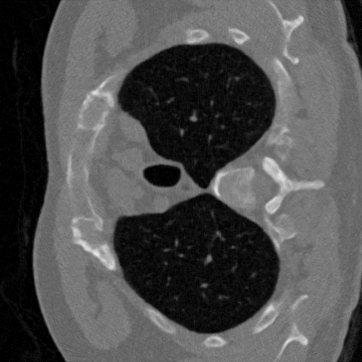}};
\spy on (1.78,1.07) in node [right] at (-0.005,-1.14);
\end{tikzpicture}
  \caption*{WPP}
\end{subfigure}%
\hfill
\begin{subfigure}[t]{.123\textwidth}  
\begin{tikzpicture}[spy using outlines=
{rectangle,white,magnification=5.5,size=2.22cm, connect spies}]
\node[anchor=south west,inner sep=0]  at (0,0) {\includegraphics[width=\linewidth]{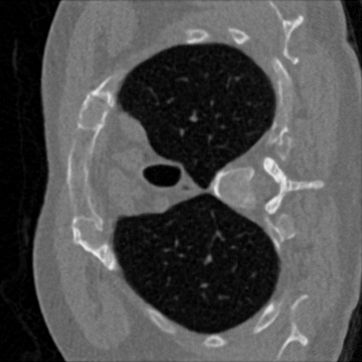}};
\spy on (1.78,1.07) in node [right] at (-0.005,-1.14);
\end{tikzpicture}
  \caption*{$\text{WPP}_{\varepsilon}$} 
\end{subfigure}%
\hfill
\begin{subfigure}[t]{.123\textwidth}  
\begin{tikzpicture}[spy using outlines=
{rectangle,white,magnification=5.5,size=2.22cm, connect spies}]
\node[anchor=south west,inner sep=0]  at (0,0) {\includegraphics[width=\linewidth]{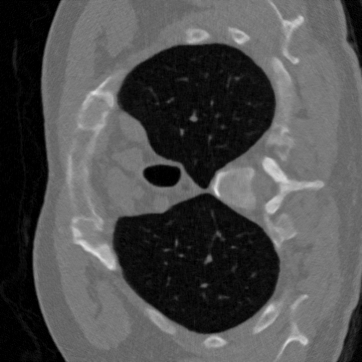}};
\spy on (1.78,1.07) in node [right] at (-0.005,-1.14);
\end{tikzpicture}
  \caption*{$\text{WPP}_{\varepsilon, \rho}$}
\end{subfigure}%
\caption{Comparison of different methods for low-dose CT reconstruction.
The zoomed-in part is marked with a white box.
\textit{Top}: full image. \textit{Bottom}: zoomed-in part. 
} \label{fig:comp_CT}
\end{figure}

First, we consider a low-dose CT example with 1000 angles between $0$ and $\pi$.
In Figure~\ref{fig:comp_CT}, we compare the different regularizers. The ALR tends to oversmooth the reconstructions and the WPP,  the $\text{WPP}_{\varepsilon}$ and the $\text{WPP}_{\varepsilon, \rho}$ are not able to reconstruct sharp edges. Both, the EPLL and the patchNR perform well, while the patchNR gives slightly more accurate and realistic reconstructions. This can be also seen quantitatively in Table~\ref{tab:error_CT}, where we evaluated the methods on the first 100 test images of the dataset. Here the patchNR gives the best results with respect to PSNR and SSIM. The weak performance of the WPP, the $\text{WPP}_{\varepsilon}$ and the $\text{WPP}_{\varepsilon, \rho}$ can be explained by the diversity of the CT dataset, leading to very different patch distributions. Therefore, defining the reference patch distribution as a mixture of patch distributions of the given 6 reference images is not sufficient for a good reconstruction.
Note that for CT data the LPIPS is not meaningful, since the feature-extracting network is trained on natural images, which differ substantially from the CT scans. Therefore, we cannot expect informative results from LPIPS.

\newcommand{\orderedresultctfull}[8]{ #1 & #2 & #4 & #6 & #3 & #5 & #7 & #8}
\begin{table}[!t]
\begin{center}
\scalebox{.85}{
\begin{tabular}[t]{c|cccccccc} 
\orderedresultctfull{}{FBP}{ALR}{EPLL}{WPP}{patchNR}{$\text{WPP}_\varepsilon$}{$\text{WPP}_{\varepsilon, \rho}$}\\  
\hline
\orderedresultctfull{PSNR}{30.37 $\pm$ 2.95}{33.59 $\pm$ 3.73}{34.89 $\pm$ 4.41}{32.61 $\pm$ 3.12}{\textbf{35.19} $\pm$ 4.52}{31.34 $\pm$ 4.22}{32.79 $\pm$ 3.27} \\ 
\orderedresultctfull{SSIM}{0.739 $\pm$ 0.141}{0.808 $\pm$ 0.146}{0.821 $\pm$ 0.154}{0.777 $\pm$ 0.121}{\textbf{0.829} $\pm$ 0.152}{0.757 $\pm$ 0.158}{0.791 $\pm$ 0.131}\\
\orderedresultctfull{FSIM}{0.941 $\pm$ 0.037}{0.945 $\pm$ 0.061}{0.935 $\pm$ 0.073}{0.950 $\pm$ 0.048}{0.935 $\pm$ 0.080}{0.936 $\pm$ 0.064}{\textbf{0.951} $\pm$ 0.049}\\
\end{tabular}}
\caption{Low-dose CT. Averaged quality measures and standard deviations of the high-resolution reconstructions. Evaluated on the first 100 test images of LoDoPab dataset. Best values are marked in bold.} 
\label{tab:error_CT}
\end{center}
\end{table}

\paragraph{Limited-Angle CT}

\begin{figure}\captionsetup[subfigure]{font=normal}
\centering
\begin{subfigure}[t]{.123\textwidth}  
\begin{tikzpicture}[spy using outlines=
{rectangle,white,magnification=4.,size=2.22cm, connect spies}]
\node[anchor=south west,inner sep=0]  at (0,0) {\includegraphics[width=\linewidth]{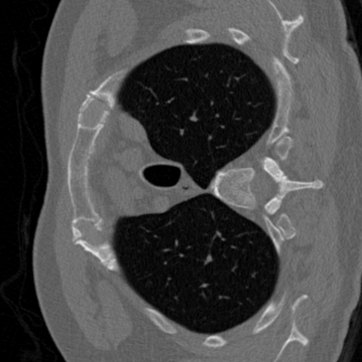}};
\spy on (1.3,1.07) in node [right] at (-0.005,-1.14);
\end{tikzpicture}
\caption*{GT}
\end{subfigure}%
\hfill
\begin{subfigure}[t]{.123\textwidth}  
\begin{tikzpicture}[spy using outlines=
{rectangle,white,magnification=4,size=2.22cm, connect spies}]
\node[anchor=south west,inner sep=0]  at (0,0) {\includegraphics[width=\linewidth]{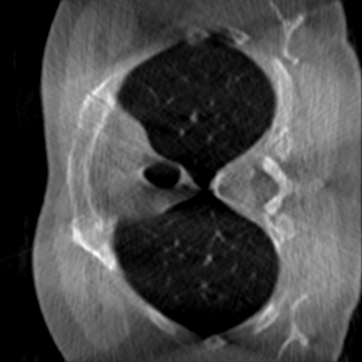}};
\spy on (1.3,1.07) in node [right] at (-0.005,-1.14);
\end{tikzpicture}
  \caption*{FBP}
\end{subfigure}%
\hfill
\begin{subfigure}[t]{.123\textwidth}  
\begin{tikzpicture}[spy using outlines=
{rectangle,white,magnification=4,size=2.22cm, connect spies}]
\node[anchor=south west,inner sep=0]  at (0,0) {\includegraphics[width=\linewidth]{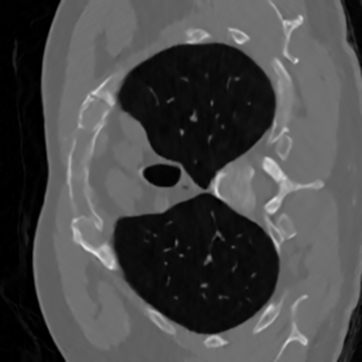}};
\spy on (1.3,1.07) in node [right] at (-0.005,-1.14);
\end{tikzpicture}
  \caption*{EPLL}
\end{subfigure}%
\hfill
\begin{subfigure}[t]{.123\textwidth}  
\begin{tikzpicture}[spy using outlines=
{rectangle,white,magnification=4,size=2.22cm, connect spies}]
\node[anchor=south west,inner sep=0]  at (0,0) {\includegraphics[width=\linewidth]{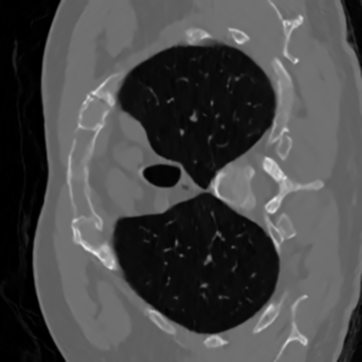}};
\spy on (1.3,1.07) in node [right] at (-0.005,-1.14);
\end{tikzpicture}
  \caption*{patchNR}
\end{subfigure}%
\hfill
\begin{subfigure}[t]{.123\textwidth}  
\begin{tikzpicture}[spy using outlines=
{rectangle,white,magnification=4,size=2.22cm, connect spies}]
\node[anchor=south west,inner sep=0]  at (0,0)  {\includegraphics[width=\linewidth]{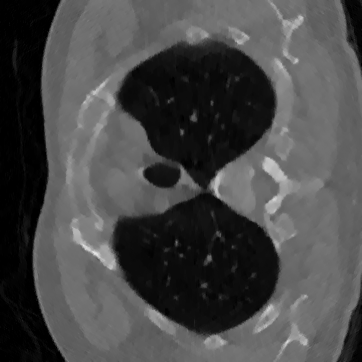}};
\spy on (1.3,1.07) in node [right] at (-0.005,-1.14);
\end{tikzpicture}
  \caption*{ALR}
\end{subfigure}%
\hfill
\begin{subfigure}[t]{.123\textwidth}  
\begin{tikzpicture}[spy using outlines=
{rectangle,white,magnification=4,size=2.22cm, connect spies}]
\node[anchor=south west,inner sep=0]  at (0,0) {\includegraphics[width=\linewidth]{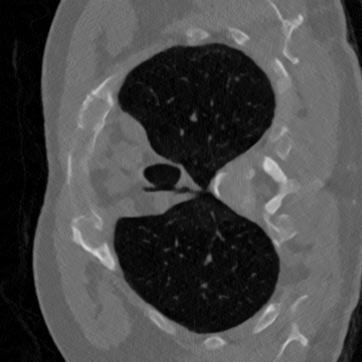}};
\spy on (1.3,1.07) in node [right] at (-0.005,-1.14);
\end{tikzpicture}
  \caption*{WPP}
\end{subfigure}%
\hfill
\begin{subfigure}[t]{.123\textwidth}  
\begin{tikzpicture}[spy using outlines=
{rectangle,white,magnification=4,size=2.22cm, connect spies}]
\node[anchor=south west,inner sep=0]  at (0,0) {\includegraphics[width=\linewidth]{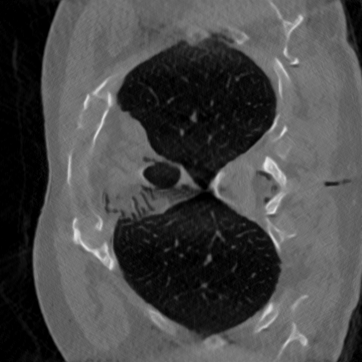}};
\spy on (1.3,1.07) in node [right] at (-0.005,-1.14);
\end{tikzpicture}
  \caption*{$\text{WPP}_{\varepsilon}$} 
\end{subfigure}%
\hfill
\begin{subfigure}[t]{.123\textwidth}  
\begin{tikzpicture}[spy using outlines=
{rectangle,white,magnification=4,size=2.22cm, connect spies}]
\node[anchor=south west,inner sep=0]  at (0,0) {\includegraphics[width=\linewidth]{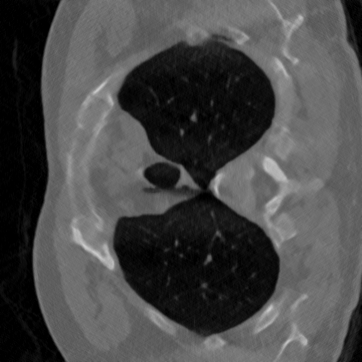}};
\spy on (1.3,1.07) in node [right] at (-0.005,-1.14);
\end{tikzpicture}
  \caption*{$\text{WPP}_{\varepsilon, \rho}$}
\end{subfigure}%
\caption{Comparison of different methods for limited-angle CT reconstruction.
The zoomed-in part is marked with a white box.
\textit{Top}: full image. \textit{Bottom}: zoomed-in part.
As visualized in the zoomed-in part, only EPLL and patchNR are able to reconstruct the missing part in the middle. The patchNR gives the sharpest results on the right of the zoomed-in part.
} \label{fig:comp_CT_lim}
\end{figure}

Next, we consider a limited-angle CT setting, i.e., instead of using 1000 equidistant angles between $0$ and $\pi$, we cut off the first and last 100 angles such that we consider $144^{\circ}$ instead of $180^{\circ}$. This leads to a much worse FBP due to the missing part in the measurement.
In Figure~\ref{fig:comp_CT_lim}, we compare the different regularizers. Again, the ALR smooths out the reconstruction, and the WPP,  the $\text{WPP}_{\varepsilon}$ and the $\text{WPP}_{\varepsilon, \rho}$ are not able to reconstruct the missing parts well.
In contrast, the EPLL and the patchNR give good reconstructions, although the patchNR gives sharper edges as can be seen in the right part of the zoomed-in part. In Table~\ref{tab:error_CT_lim}, a quantitative comparison is given. Again, the patchNR gives the best results with respect to PSNR and SSIM.

\newcommand{\orderedresultctlimited}[8]{ #1 & #2 & #4 & #6 & #3 & #5 & #7 & #8}
\begin{table}[!t]
\begin{center}
\scalebox{0.85}{
\begin{tabular}[t]{c|ccccccc} 
\orderedresultctlimited{}{FBP}{ALR}{EPLL}{WPP}{patchNR}{$\text{WPP}_\varepsilon$}{$\text{WPP}_{\varepsilon, \rho}$}\\  
\hline
\orderedresultctlimited{PSNR}{21.96 $\pm$ 2.25}{31.27 $\pm$ 2.94}{33.14 $\pm$ 3.58}{29.92 $\pm$ 2.36}{\textbf{33.26} $\pm$ 3.58}{25.10 $\pm$ 3.98}{30.00 $\pm$ 2.55}\\ 
\orderedresultctlimited{SSIM}{0.531 $\pm$ 0.097}{0.783 $\pm$ 0.143}{0.804 $\pm$ 0.154}{0.737 $\pm$ 0.114}{\textbf{0.811} $\pm$ 0.151}{0.642 $\pm$ 0.144}{0.753 $\pm$ 0.125} \\
\orderedresultctlimited{FSIM}{0.913 $\pm$ 0.032}{\textbf{0.929} $\pm$ 0.053}{0.920 $\pm$ 0.071}{0.920 $\pm$ 0.048}{0.921 $\pm$ 0.077}{0.846 $\pm$ 0.111}{0.922 $\pm$ 0.048} \\
\end{tabular}}
\caption{Limited-angle CT. Averaged quality measures and standard deviations of the high-resolution reconstructions. Evaluated on the first 100 test images of LoDoPab dataset. Best values are marked in bold.} 
\label{tab:error_CT_lim}
\end{center}
\end{table}

\subsection{Super-Resolution} \label{sec:superres}

For image super-resolution, we want to recover a high-resolution image from a given low-resolution image. The forward operator $F$ consists of a convolution with a $16 \times 16$ Gaussian blur kernel of a certain standard deviation specified below and a subsampling process. 
For the noise model, we consider additive Gaussian noise with standard deviation $\Xi \sim \mathcal{N}(0,\sigma^2 I)$ with standard deviation $\sigma = 0.01$. Consequently, we want to minimize the variational problem \eqref{Gaussian_noise} with  $\alpha = \beta\sigma^2$.

We consider two different types of super-resolution: First, we deal with the super-resolution of material data. Here we assume that we are given one high-resolution reference image of the material which we can use as prior knowledge.
Second, we consider zero-shot super-resolution of natural images, where no reference data is known.

\paragraph{Material Data}

The dataset consists of 2D slices of size $600 \times 600$ from a 3D material image of size $2560 \times 2560 \times 2120$. This has been acquired by synchrotron micro-computed tomography at the SLS beamline TOMCAT. More specifically, we consider a composite (``SiC Diamond'') obtained by microwave sintering of silicon and diamonds, see \cite{vaucher2007line}. We assume that we are given one high-resolution reference image of size $600 \times 600$. 
The blur kernel of the forward operator $F$ has standard deviation $2$ and we consider a subsampling factor of $4$ (in each direction). 

\begin{figure}[t]\captionsetup[subfigure]{font=normal}
\centering
\begin{subfigure}[t]{.123\textwidth}  
\begin{tikzpicture}[spy using outlines=
{rectangle,black,magnification=10,size=2.17cm, connect spies}]
\node[anchor=south west,inner sep=0]  at (0,0) {\includegraphics[width=\linewidth]{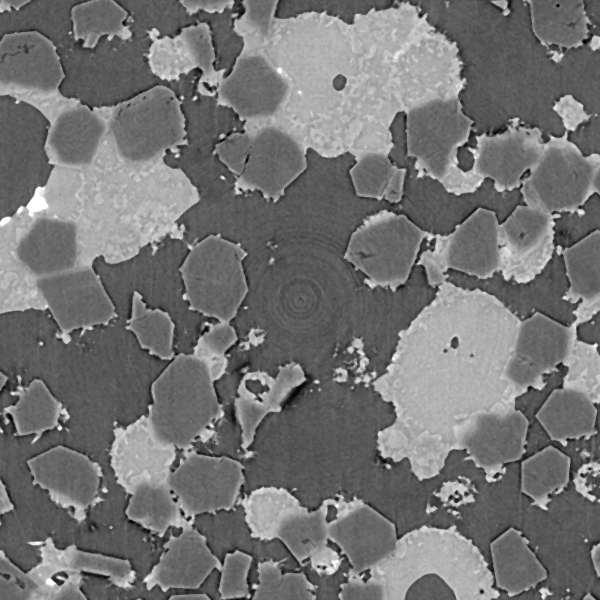}};
\spy on (1.3,.875) in node [right] at (0.02,-1.17);
\end{tikzpicture}
\caption*{HR}
\end{subfigure}%
\hfill
\begin{subfigure}[t]{.123\textwidth}
\begin{tikzpicture}[spy using outlines=
{rectangle,black,magnification=10,size=2.17cm, connect spies}]
\node[anchor=south west,inner sep=0]  at (0,0) {\includegraphics[width=\linewidth]{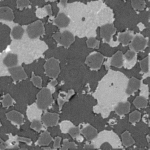}};
\spy on (1.282,.893) in node [right] at (0.02,-1.17);
\end{tikzpicture}
  \caption*{LR}
\end{subfigure}%
\hfill
\begin{subfigure}[t]{.123\textwidth}
\begin{tikzpicture}[spy using outlines=
{rectangle,black,magnification=10,size=2.17cm, connect spies}]
\node[anchor=south west,inner sep=0]  at (0,0) {\includegraphics[width=\linewidth]{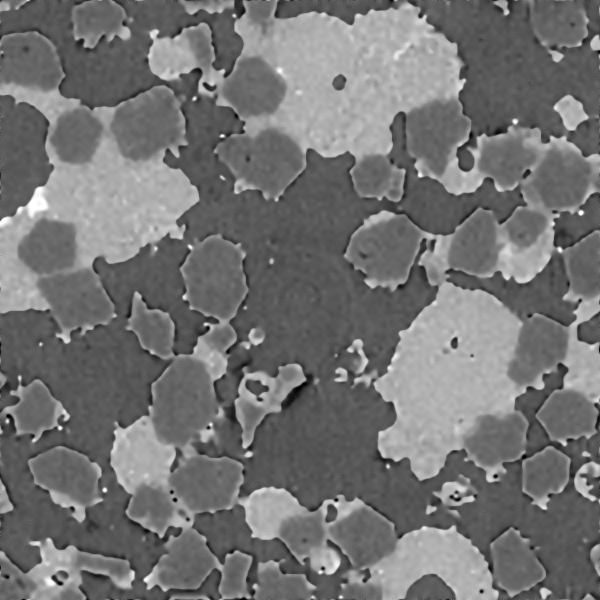}};
\spy on (1.3,.875) in node [right] at (0.02,-1.17);
\end{tikzpicture}
  \caption*{EPLL}
\end{subfigure}%
\hfill
\begin{subfigure}[t]{.123\textwidth}
\begin{tikzpicture}[spy using outlines=
{rectangle,black,magnification=10,size=2.17cm, connect spies}]
\node[anchor=south west,inner sep=0]  at (0,0) {\includegraphics[width=\linewidth]{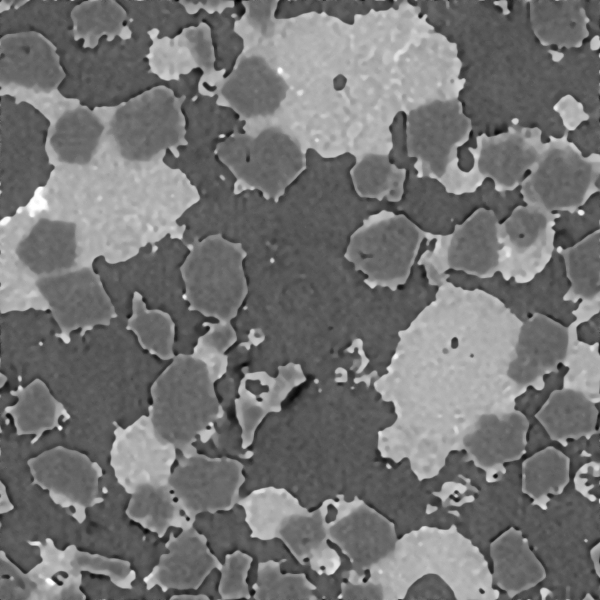}};
\spy on (1.3,.875) in node [right] at (0.02,-1.17);
\end{tikzpicture}
  \caption*{patchNR}
\end{subfigure}%
\hfill
\begin{subfigure}[t]{.123\textwidth}
\begin{tikzpicture}[spy using outlines=
{rectangle,black,magnification=10,size=2.17cm, connect spies}]
\node[anchor=south west,inner sep=0]  at (0,0) {\includegraphics[width=\linewidth]{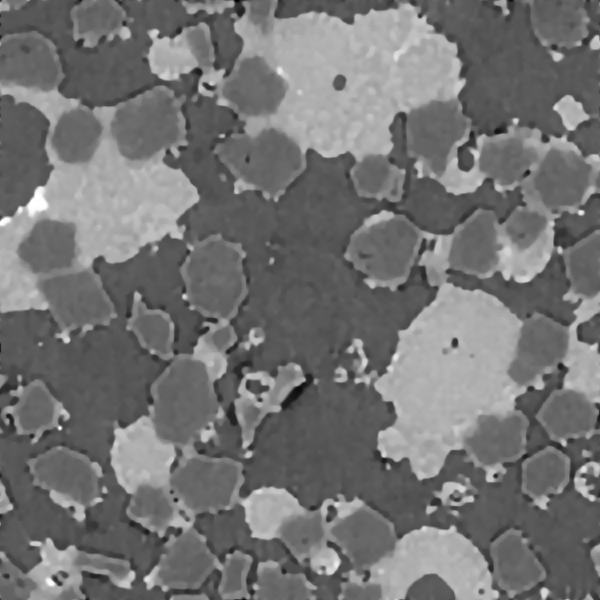}};
\spy on (1.3,.875) in node [right] at (0.02,-1.17);
\end{tikzpicture}
  \caption*{ALR}
\end{subfigure}%
\hfill
\begin{subfigure}[t]{.123\textwidth}
\begin{tikzpicture}[spy using outlines=
{rectangle,black,magnification=10,size=2.17cm, connect spies}]
\node[anchor=south west,inner sep=0]  at (0,0) {\includegraphics[width=\linewidth]{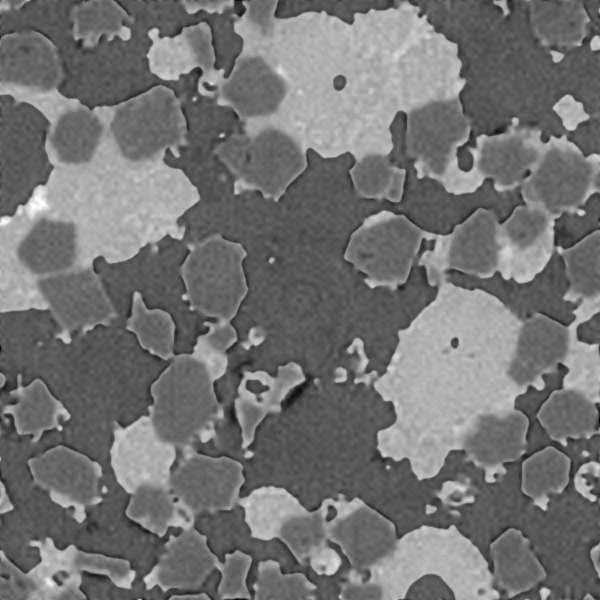}};
\spy on (1.3,.875) in node [right] at (0.02,-1.17);
\end{tikzpicture}
  \caption*{WPP}
\end{subfigure}%
\hfill
\begin{subfigure}[t]{.123\textwidth}
\begin{tikzpicture}[spy using outlines=
{rectangle,black,magnification=10,size=2.17cm, connect spies}]
\node[anchor=south west,inner sep=0]  at (0,0) {\includegraphics[width=\linewidth]{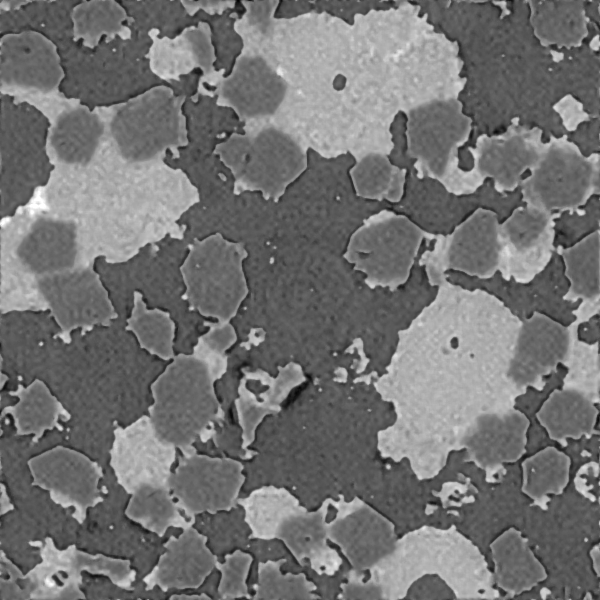}};
\spy on (1.3,.875) in node [right] at (0.02,-1.17);
\end{tikzpicture}
  \caption*{WP$\text{P}_{\varepsilon}$} 
\end{subfigure}%
\hfill
\begin{subfigure}[t]{.123\textwidth}
\begin{tikzpicture}[spy using outlines=
{rectangle,black,magnification=10,size=2.17cm, connect spies}]
\node[anchor=south west,inner sep=0]  at (0,0) {\includegraphics[width=\linewidth]{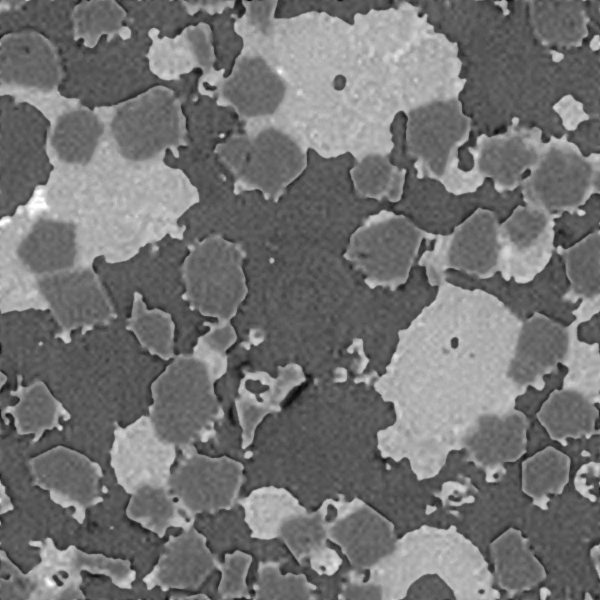}};
\spy on (1.3,.875) in node [right] at (0.02,-1.17);
\end{tikzpicture}
  \caption*{WP$\text{P}_{\varepsilon, \rho}$} 
\end{subfigure}%
\caption{Comparison of different methods for super-resolution. 
The zoomed-in part is marked with a black box.
\textit{Top}: full image. \textit{Bottom}: zoomed-in part.
} \label{fig:comp_SR_material}
\end{figure}

In Figure~\ref{fig:comp_SR_material}, we compare the different regularizers, where we choose the bicubic interpolation as initialization. In the reconstruction of the ALR and the WPP, we can observe a significant blur, in particular in the regions between the edges. 
In contrast, the EPLL and the patchNR reconstructions are sharper and more realistic. 
The WPP, the WP$\text{P}_{\varepsilon}$ and the WP$\text{P}_{\varepsilon, \rho}$ reconstructions have quite similar lower quality. 
A quantitative comparison is given in Table~\ref{tab:error_material}.

\newcommand{\orderedresultssr}[8]{ #1 & #2 & #4 & #6 & #3 & #5 & #7 & #8}
\begin{table}[!t]
\begin{center}
\scalebox{.85}{
\begin{tabular}[t]{c|ccccccc} 
\orderedresultssr{}{bicubic}{ALR}{EPLL}{WPP}{patchNR}{WP$\text{P}_\varepsilon$}{WP$\text{P}_{\varepsilon, \rho}$}\\  
\hline
\orderedresultssr{PSNR}{25.63 $\pm$ 0.56  }{ 27.76 $\pm$ 0.52}{28.34 $\pm$ 0.50}{27.55 $\pm$ 0.46}{\textbf{28.53} $\pm$ 0.49}{26.60 $\pm$ 0.30}{27.46 $\pm$ 0.48}\\
\orderedresultssr{SSIM}{0.699 $\pm$ 0.012 }{0.758 $\pm$ 0.005}{0.770 $\pm$ 0.008}{0.737 $\pm$ 0.007}{\textbf{0.780} $\pm$ 0.008}{0.698 $\pm$ 0.020}{0.727 $\pm$ 0.006}\\
\orderedresultssr{LPIPS}{0.414 $\pm$ 0.011}{0.187 $\pm$ 0.005}{0.175 $\pm$ 0.009}{0.188 $\pm$ 0.008}{\textbf{0.161} $\pm$ 0.007}{0.177 $\pm$ 0.027}{0.186 $\pm$ 0.007}\\
\orderedresultssr{FSIM}{0.878 $\pm$ 0.005}{0.932 $\pm$ 0.003}{0.933 $\pm$ 0.005}{0.937 $\pm$ 0.003}{\textbf{0.940} $\pm$ 0.004}{0.919 $\pm$ 0.007}{0.938 $\pm$ 0.003} 
\end{tabular}}
\caption{Super-resolution. Averaged quality measures and standard deviations of the high-resolution reconstructions. Evaluated on the material data test set. Best values are marked in bold.} 
\label{tab:error_material}
\end{center}
\end{table}

\paragraph{Zero-Shot Super-Resolution} 

\begin{figure}[!t]
\captionsetup[subfigure]{font=normal,justification=centering}
\centering
\begin{subfigure}{0.245\textwidth}
\includegraphics[width=0.65\textwidth]{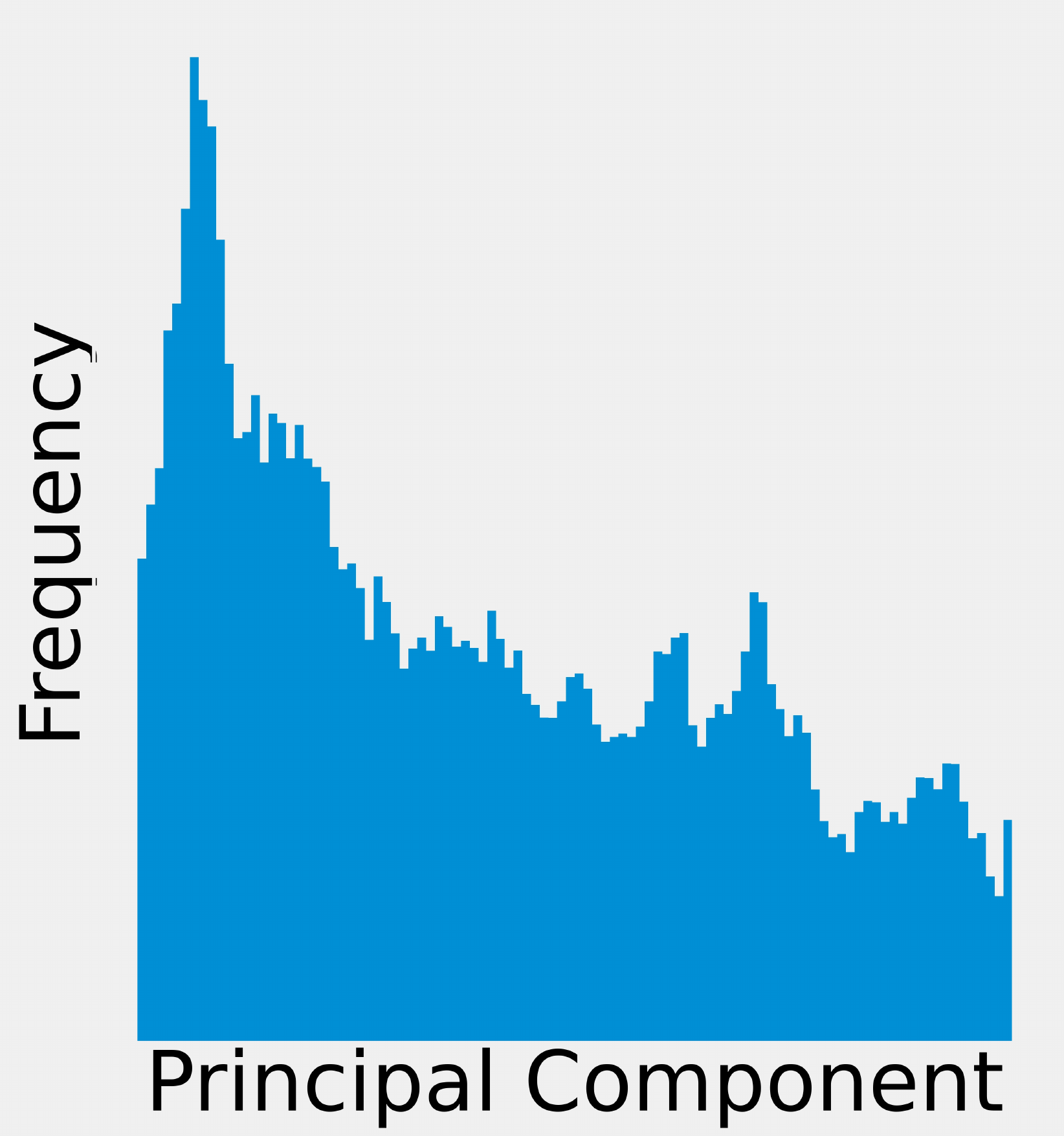} 
\hfill
\begin{tikzpicture}[spy using outlines={rectangle,black,magnification=13,size=1.07cm, connect spies}]
\node {\includegraphics[trim=0 1.14cm 0 0, width=1.1cm]{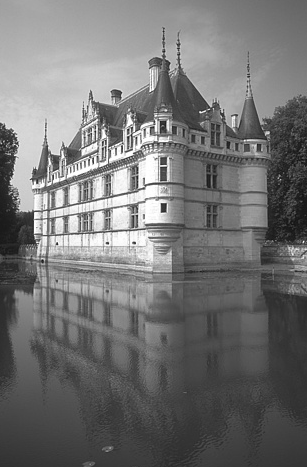}};

\spy on (0.04,0.18) in node [right] at (-.535cm, 1.57);
\end{tikzpicture} 
\caption*{Original}
\end{subfigure}
\hfill
\begin{subfigure}{0.245\textwidth}
\includegraphics[width=0.65\textwidth]{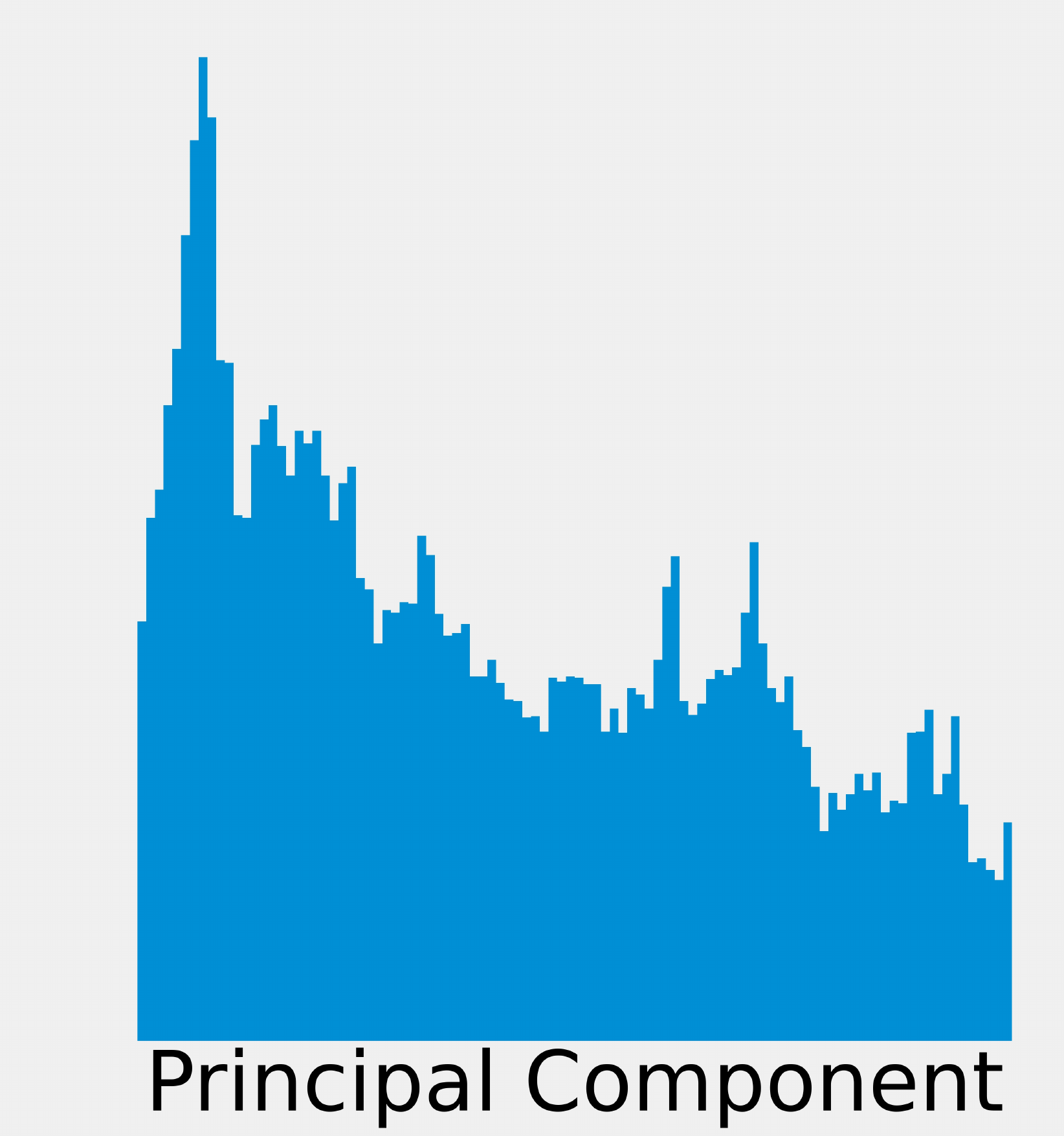} 
\hfill
\begin{tikzpicture}[spy using outlines={rectangle,black,magnification=13,size=1.07cm, connect spies}]
\node {\includegraphics[trim=0 .57cm 0 0, width=1.1cm]{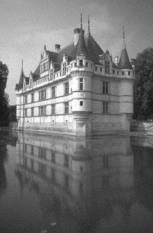}};

\spy on (0.04,0.18) in node [right] at (-.535cm, 1.57);
\end{tikzpicture} 
\caption*{Downsampling $\times$ 2}
\end{subfigure}%
\hfill
\begin{subfigure}{0.245\textwidth}
\includegraphics[width=0.65\textwidth]{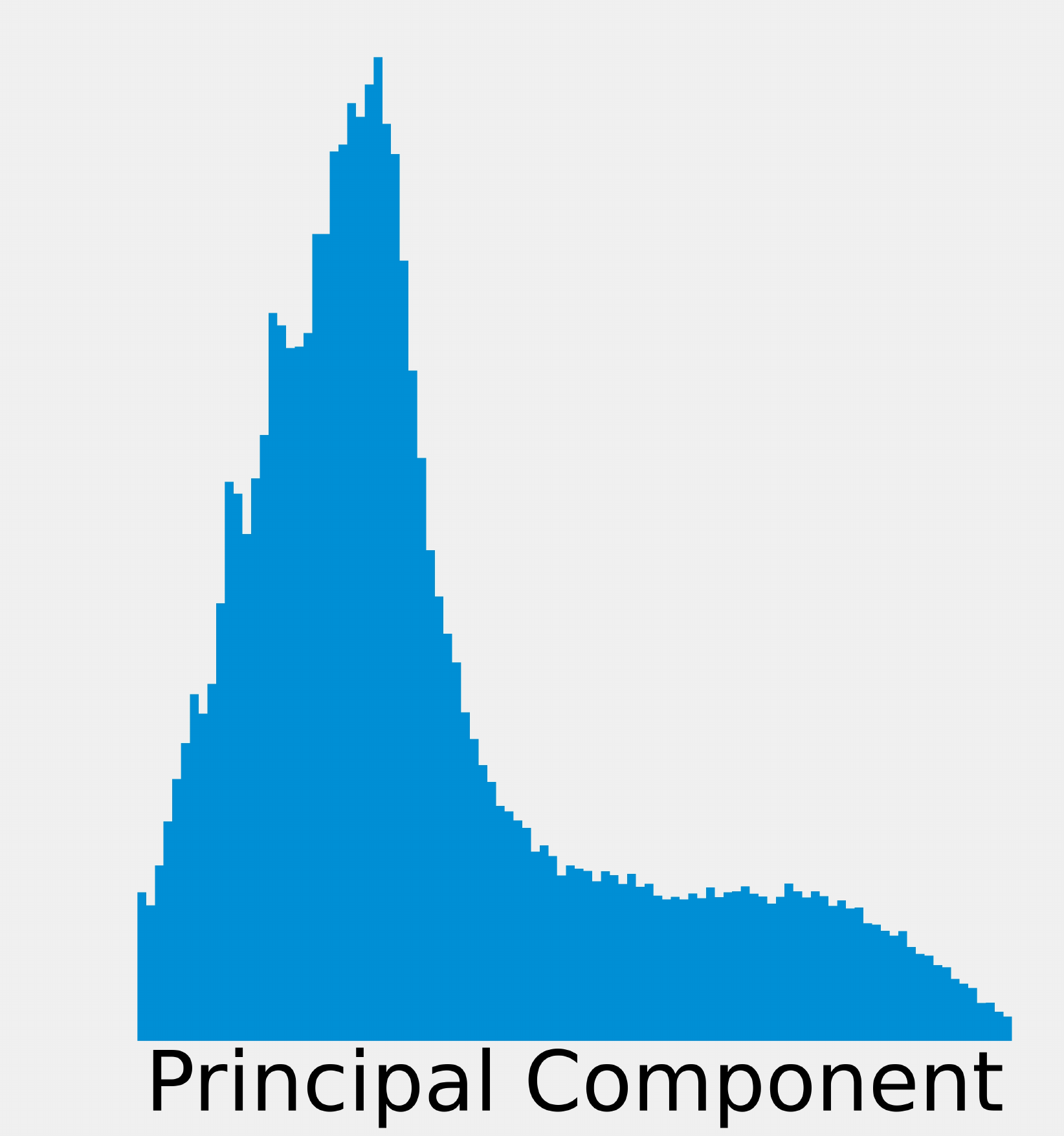} 
\hfill
\begin{tikzpicture}[spy using outlines={rectangle,black,magnification=13,size=1.07cm, connect spies}]
\node {\includegraphics[trim=0 1.14cm 0 0, width=1.1cm]{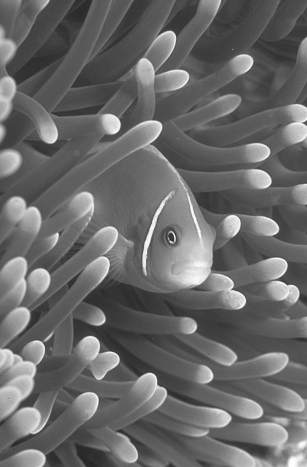}};
\spy on (0.06,-0.07) in node [right] at (-.535cm, 1.57);
\end{tikzpicture} 
\caption*{Original}
\end{subfigure}
\hfill
\begin{subfigure}{0.245\textwidth}
\includegraphics[width=0.65\textwidth]{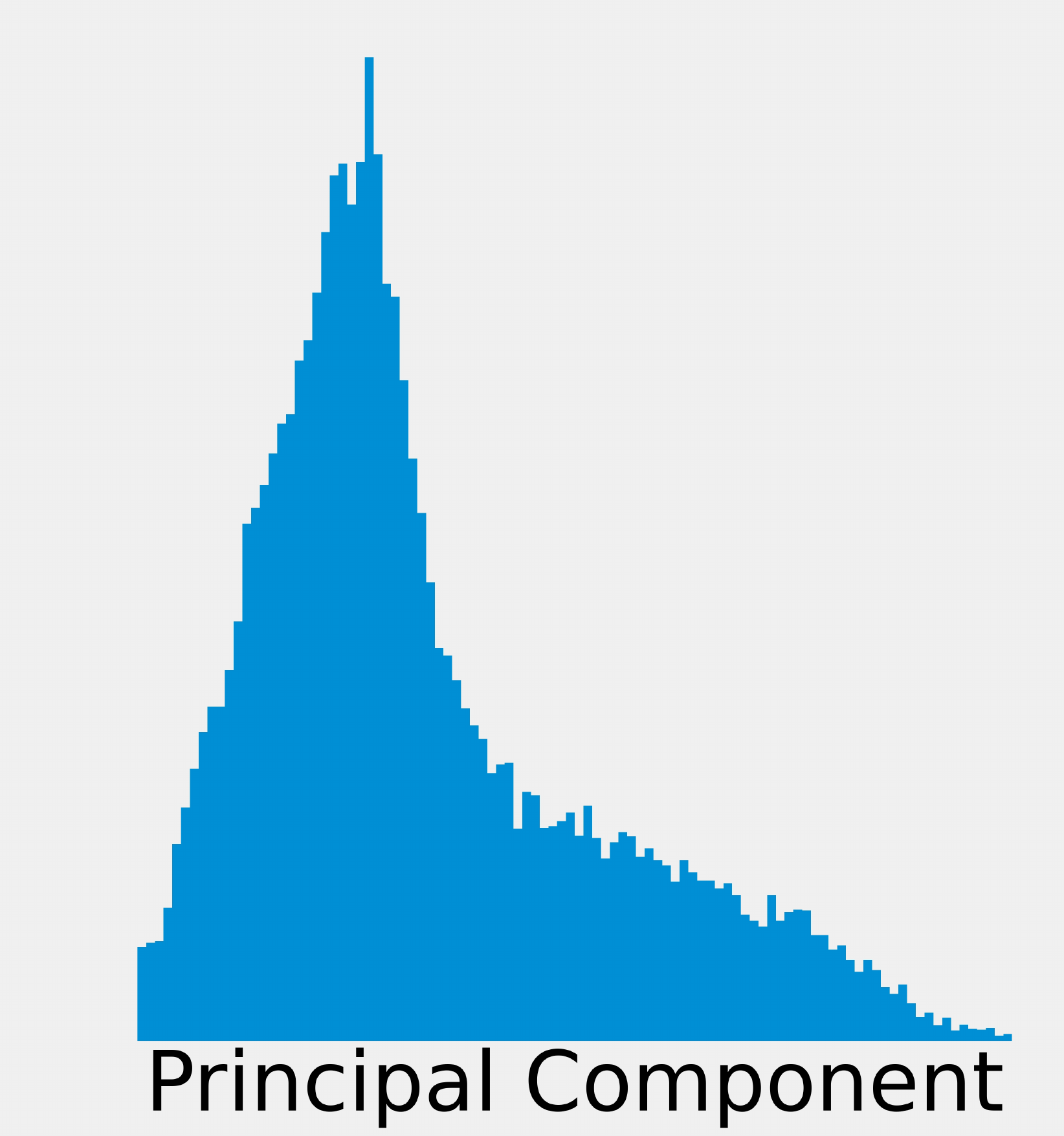} 
\hfill
\begin{tikzpicture}[spy using outlines={rectangle,black,magnification=13,size=1.07cm, connect spies}]
\node {\includegraphics[trim=0 .57cm 0 0, width=1.1cm]{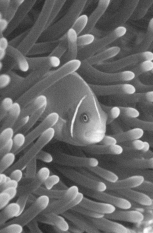}};
\spy on (0.06,-0.07) in node [right] at (-.535cm, 1.57);

\end{tikzpicture} 
\caption*{Downsampling $\times$ 2}
\end{subfigure}%
\caption{We utilize the same method as in Figure~\ref{patchpca_fig1} to visualize the patch distribution for two natural images from the BSD68 dataset \cite{bsd_data}. Both images are downsampled by a factor of 2 using a Gaussian blur operator in combination with additive Gaussian noise. We project all patches onto the direction of the first principal component and illustrate the resulting distribution with a histogram. For every image, the first principal component explains more than 70\% of the total variance. Visually, the projected patch distributions of the original images and their downsampled versions are highly similar.}
\label{patchpca_scales_fig}
\end{figure}

\newcommand{\orderedresultssrzero}[8]{ #1 & #2 & #4 & #6 & #3 & #5 & #7 & #8}
\begin{table}[b!]
\begin{center}
\scalebox{.85}{
\begin{tabular}[t]{c|ccccccc} 
\orderedresultssrzero{}{bicubic}{ALR}{EPLL}{WPP}{patchNR}{WP$\text{P}_\varepsilon$}{WP$\text{P}_{\varepsilon, \rho}$}\\  
\hline
\orderedresultssrzero{PSNR}{27.06 $\pm$ 3.39}{28.61 $\pm$ 3.51}{28.90 $\pm$ 3.53}{28.20 $\pm$ 3.03}{\textbf{29.08} $\pm$ 3.58}{27.92 $\pm$ 2.81}{28.36 $\pm$ 3.24}\\ 
\orderedresultssrzero{SSIM}{0.782 $\pm$ 0.075}{0.829 $\pm$ 0.066}{0.838 $\pm$ 0.065}{0.790 $\pm$ 0.055}{\textbf{0.846} $\pm$ 0.061}{0.771 $\pm$ 0.056}{0.806 $\pm$ 0.055} \\
\orderedresultssrzero{LPIPS}{0.327 $\pm$ 0.075}{\textbf{0.196} $\pm$ 0.072}{0.204 $\pm$ 0.079}{0.251 $\pm$ 0.062}{0.203 $\pm$ 0.075}{0.259 $\pm$ 0.066}{0.245 $\pm$ 0.065} \\
\orderedresultssrzero{FSIM}{0.952 $\pm$ 0.023}{0.974 $\pm$ 0.013}{0.977 $\pm$ 0.010}{0.969 $\pm$ 0.023}{\textbf{0.980} $\pm$ 0.008}{0.964 $\pm$ 0.029}{0.973 $\pm$ 0.016} \\
\end{tabular}}
\caption{Zero-shot super-resolution. Averaged quality measures and standard deviations of the high-resolution reconstructions. Evaluated on BSD68. Best values are marked in bold.} 
\label{tab:error_zershot}
\end{center}
\end{table}

We consider the grayscale BSD68 dataset \cite{bsd_data}. The blur kernel of the forward operator $F$ has standard deviation 1 and we consider a subsampling factor of 2. 

\begin{figure}[t]\captionsetup[subfigure]{font=normal}
\centering
\begin{subfigure}[t]{.123\textwidth}  
\begin{tikzpicture}[spy using outlines=
{rectangle,black,magnification=10.5,width=2.17cm, height=1.42cm, connect spies}]
\node[anchor=south west,inner sep=0]  at (0,0) {\includegraphics[width=\linewidth]{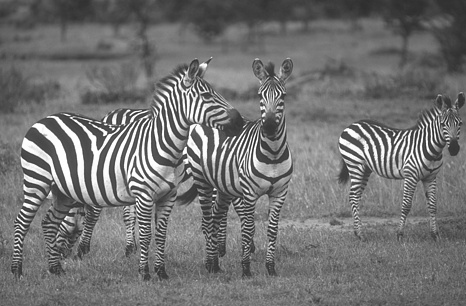}};
\spy on (1.95,.76) in node [right] at (0.02,-.76);
\end{tikzpicture}
\end{subfigure}%
\hfill
\begin{subfigure}[t]{.123\textwidth}
\begin{tikzpicture}[spy using outlines=
{rectangle,black,magnification=10.5,width=2.17cm, height=1.42cm, connect spies}]
\node[anchor=south west,inner sep=0]  at (0,0) {\includegraphics[width=\linewidth]{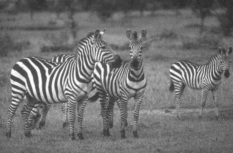}};
\spy on (1.95,.76) in node [right] at (0.02,-.76);
\end{tikzpicture}
\end{subfigure}%
\hfill
\begin{subfigure}[t]{.123\textwidth}
\begin{tikzpicture}[spy using outlines=
{rectangle,black,magnification=10.5,width=2.17cm, height=1.42cm, connect spies}]
\node[anchor=south west,inner sep=0]  at (0,0) {\includegraphics[width=\linewidth]{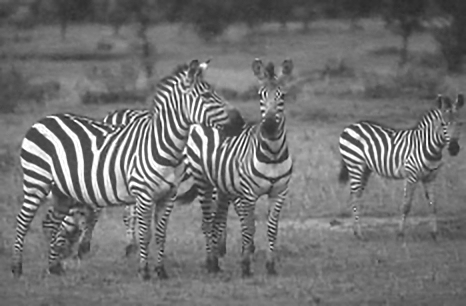}};
\spy on (1.95,.76) in node [right] at (0.02,-.76);
\end{tikzpicture}
\end{subfigure}%
\hfill
\begin{subfigure}[t]{.123\textwidth}
\begin{tikzpicture}[spy using outlines=
{rectangle,black,magnification=10.5,width=2.17cm, height=1.42cm, connect spies}]
\node[anchor=south west,inner sep=0]  at (0,0) {\includegraphics[width=\linewidth]{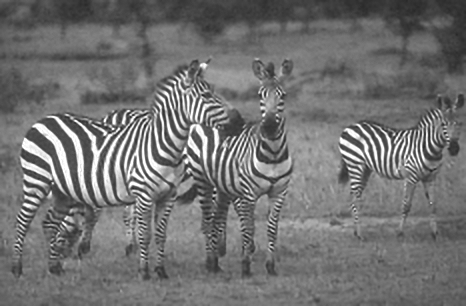}};
\spy on (1.95,.76) in node [right] at (0.02,-.76);
\end{tikzpicture}
\end{subfigure}%
\hfill
\begin{subfigure}[t]{.123\textwidth}
\begin{tikzpicture}[spy using outlines=
{rectangle,black,magnification=10.5,width=2.17cm, height=1.42cm, connect spies}]
\node[anchor=south west,inner sep=0]  at (0,0) {\includegraphics[width=\linewidth]{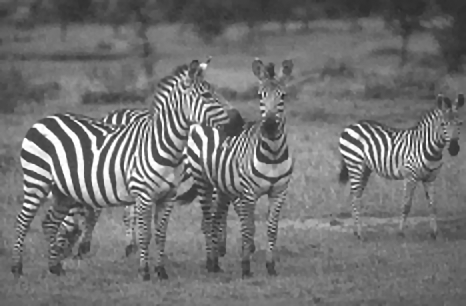}};
\spy on (1.95,.76) in node [right] at (0.02,-.76);
\end{tikzpicture}
\end{subfigure}%
\hfill
\begin{subfigure}[t]{.123\textwidth}
\begin{tikzpicture}[spy using outlines=
{rectangle,black,magnification=10.5,width=2.17cm, height=1.42cm, connect spies}]
\node[anchor=south west,inner sep=0]  at (0,0) {\includegraphics[width=\linewidth]{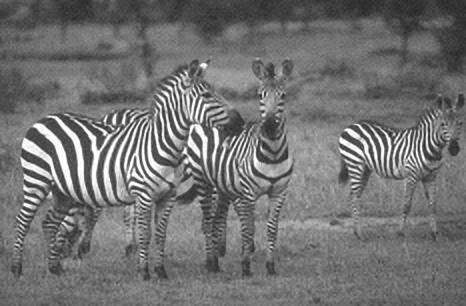}};
\spy on (1.95,.76) in node [right] at (0.02,-.76);
\end{tikzpicture}
\end{subfigure}%
\hfill
\begin{subfigure}[t]{.123\textwidth}
\begin{tikzpicture}[spy using outlines=
{rectangle,black,magnification=10.5,width=2.17cm, height=1.42cm, connect spies}]
\node[anchor=south west,inner sep=0]  at (0,0) {\includegraphics[width=\linewidth]{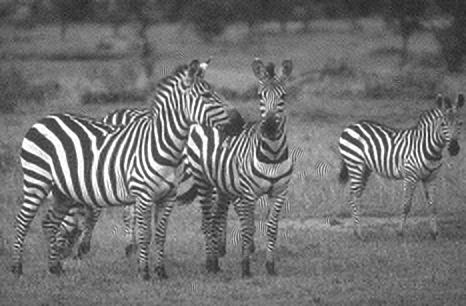}};
\spy on (1.95,.76) in node [right] at (0.02,-.76);
\end{tikzpicture}

\end{subfigure}%
\hfill
\begin{subfigure}[t]{.123\textwidth}
\begin{tikzpicture}[spy using outlines=
{rectangle,black,magnification=10.5,width=2.17cm, height=1.42cm, connect spies}]
\node[anchor=south west,inner sep=0]  at (0,0) {\includegraphics[width=\linewidth]{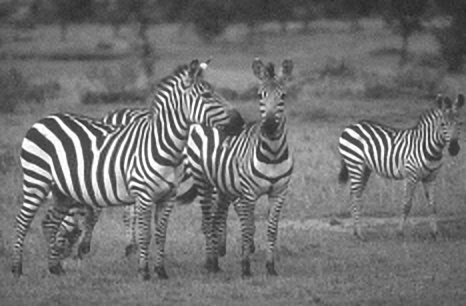}};
\spy on (1.95,.76) in node [right] at (0.02,-.76);
\end{tikzpicture}
\end{subfigure}%

\vspace{0.2cm}

\begin{subfigure}[t]{.123\textwidth}  
\begin{tikzpicture}[spy using outlines=
{rectangle,black,magnification=10.5,width=2.17cm, height=1.42cm, connect spies}]
\node[anchor=south west,inner sep=0]  at (0,0) {\includegraphics[width=\linewidth]{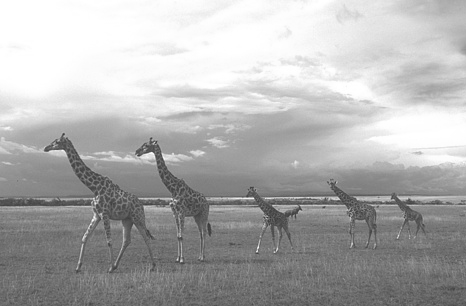}};
\spy on (.9,.49) in node [right] at (0.02,-.76);
\end{tikzpicture}
\caption*{HR}
\end{subfigure}%
\hfill
\begin{subfigure}[t]{.123\textwidth}
\begin{tikzpicture}[spy using outlines=
{rectangle,black,magnification=10.5,width=2.17cm, height=1.42cm, connect spies}]
\node[anchor=south west,inner sep=0]  at (0,0) {\includegraphics[width=\linewidth]{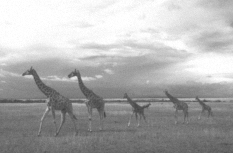}};
\spy on (.9,.49) in node [right] at (0.02,-.76);
\end{tikzpicture}
\caption*{LR}
\end{subfigure}%
\hfill
\begin{subfigure}[t]{.123\textwidth}
\begin{tikzpicture}[spy using outlines=
{rectangle,black,magnification=10.5,width=2.17cm, height=1.42cm, connect spies}]
\node[anchor=south west,inner sep=0]  at (0,0) {\includegraphics[width=\linewidth]{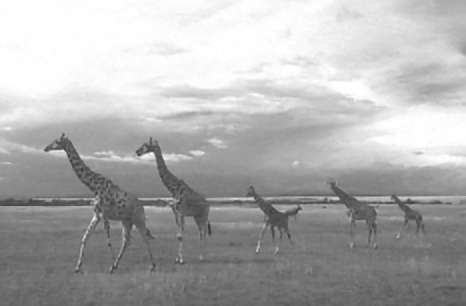}};
\spy on (.9,.49) in node [right] at (0.02,-.76);
\end{tikzpicture}
\caption*{EPLL}
\end{subfigure}%
\hfill
\begin{subfigure}[t]{.123\textwidth}
\begin{tikzpicture}[spy using outlines=
{rectangle,black,magnification=10.5,width=2.17cm, height=1.42cm, connect spies}]
\node[anchor=south west,inner sep=0]  at (0,0) {\includegraphics[width=\linewidth]{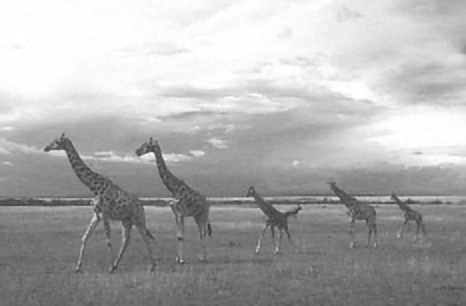}};
\spy on (.9,.49) in node [right] at (0.02,-.76);
\end{tikzpicture}
\caption*{patchNR}
\end{subfigure}%
\hfill
\begin{subfigure}[t]{.123\textwidth}
\begin{tikzpicture}[spy using outlines=
{rectangle,black,magnification=10.5,width=2.17cm, height=1.42cm, connect spies}]
\node[anchor=south west,inner sep=0]  at (0,0) {\includegraphics[width=\linewidth]{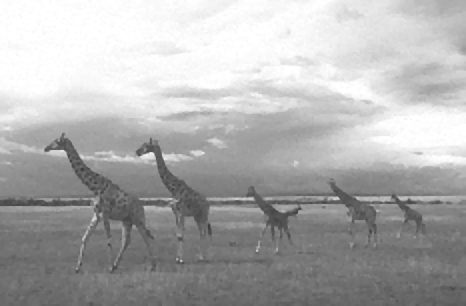}};
\spy on (.9,.49) in node [right] at (0.02,-.76);
\end{tikzpicture}
\caption*{ALR}
\end{subfigure}%
\hfill
\begin{subfigure}[t]{.123\textwidth}
\begin{tikzpicture}[spy using outlines=
{rectangle,black,magnification=10.5,width=2.17cm, height=1.42cm, connect spies}]
\node[anchor=south west,inner sep=0]  at (0,0) {\includegraphics[width=\linewidth]{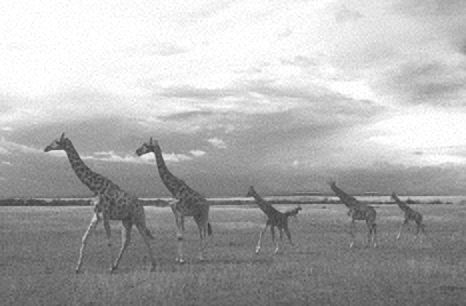}};
\spy on (.9,.49) in node [right] at (0.02,-.76);
\end{tikzpicture}
\caption*{WPP}
\end{subfigure}%
\hfill
\begin{subfigure}[t]{.123\textwidth}
\begin{tikzpicture}[spy using outlines=
{rectangle,black,magnification=10.5,width=2.17cm, height=1.42cm, connect spies}]
\node[anchor=south west,inner sep=0]  at (0,0) {\includegraphics[width=\linewidth]{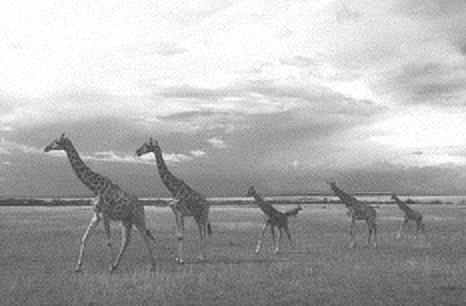}};
\spy on (.9,.49) in node [right] at (0.02,-.76);
\end{tikzpicture}
\caption*{WP$\text{P}_{\varepsilon}$} 
\end{subfigure}%
\hfill
\begin{subfigure}[t]{.123\textwidth}
\begin{tikzpicture}[spy using outlines=
{rectangle,black,magnification=10.5,width=2.17cm, height=1.42cm, connect spies}]
\node[anchor=south west,inner sep=0]  at (0,0) {\includegraphics[width=\linewidth]{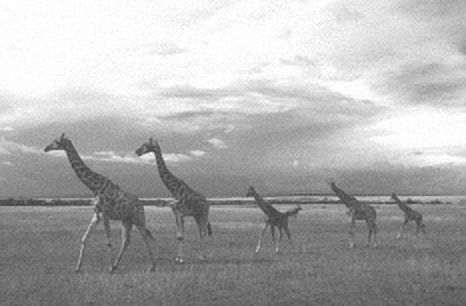}};
\spy on (.9,.49) in node [right] at (0.02,-.76);
\end{tikzpicture}
\caption*{WP$\text{P}_{\varepsilon, \rho}$} 
\end{subfigure}%
\caption{Comparison of different methods for zero-shot super-resolution. 
The zoomed-in part is marked with a black box.
\textit{Top}: full image. \textit{Bottom}: zoomed-in part.
The ALR and the EPLL smooth out parts of the reconstruction when these parts are blurry in the low-resolution part, see, e.g., the stripes of the zebra or the fur pattern of the giraffe.
} \label{fig:comp_SR_zeroshot}
\end{figure}

We assume that no reference data is given so that we need to extract our prior information from the given low-resolution observation. 
Here we exploit the concepts of zero-shot super-resolution by internal learning. The main observation is that the patch distribution of natural images is self-similar across the scales \cite{GBI2009,SCI2018,zontak2011internalpatchstatistics}. Thus the patch distributions of the same image are similar at different resolutions. An illustrative example with two images from the BSD68 dataset is given in Figure~\ref{patchpca_scales_fig}. 
The reconstruction of the unknown high-resolution image using the different regularizers is visualized in Figure~\ref{fig:comp_SR_zeroshot}. Here ALR and  EPLL smooth out parts of the reconstruction, in particular when these parts are blurry in the low-resolution part, see, e.g., the stripes of the zebra or the fur pattern of the giraffe. In contrast,  the WPP,  WP$\text{P}_{\varepsilon, \rho}$ and the patchNR are able to reconstruct well and without blurred parts. The WP$\text{P}_{\varepsilon}$ reconstructions admit structured noise, which can be seen in the upper right corner of the zoomed-in part of the giraffe.
A quantitative comparison is given in Table~\ref{tab:error_zershot}. Again, the patchNR performs best in terms of quality measures.

\subsection{Inpainting} \label{sec:img_inpainting}

The task of image inpainting is to reconstruct missing data in the observation. For a given inpainting mask $m \in \{0,1 \}^n$, the forward operator $F$ is given by $F(x) = x \odot m$. In this subsection, we focus on region inpainting, where large regions of data are missing in the observation. 
We assume that there is no additional noise 
in the observation, leading to the negative log-likelihood
\begin{align} \label{eq:inpainting_likelihood}
- \log (p_{Y|X=x}(y)) =
\begin{cases}
0, & ~ \text{if} ~ F(x) = y, \\
+ \infty, & ~ \text{else}.
\end{cases}
\end{align}
Consequently, we are searching for 
\begin{align*}
\argmin{x \in \mathbb R^d} \left\{ \mathcal{R}(x): \,  F(x) = y \right\}.
\end{align*}
We consider the Set5 dataset \cite{BRGA2012} and assume that no reference data is given, such that we extract the prior information from  a predefined area around the missing part of the observation.

\begin{figure}[!bt]\captionsetup[subfigure]{font=normal}
\centering
\begin{subfigure}[bt]{.123\textwidth}  
\begin{tikzpicture}[spy using outlines=
{rectangle,black,magnification=5,size=2.17cm, connect spies}]
\node[anchor=south west,inner sep=0]  at (0,0) {\includegraphics[width=\linewidth]{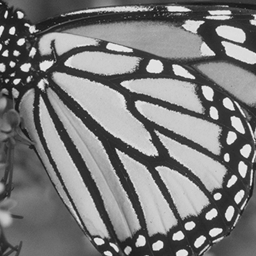}};
\spy on (1.55,.66) in node [right] at (0.02,-1.14);
\end{tikzpicture}
\caption*{GT}
\end{subfigure}%
\hfill
\begin{subfigure}[bt]{.123\textwidth}
\begin{tikzpicture}[spy using outlines=
{rectangle,black,magnification=5,size=2.17cm, connect spies}]
\node[anchor=south west,inner sep=0]  at (0,0) {\includegraphics[width=\linewidth]{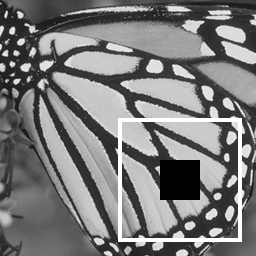}};
\spy on (1.55,.66) in node [right] at (0.02,-1.14);
\end{tikzpicture}
  \caption*{Observation}
\end{subfigure}%
\hfill
\begin{subfigure}[bt]{.123\textwidth}
\begin{tikzpicture}[spy using outlines=
{rectangle,black,magnification=5,size=2.17cm, connect spies}]
\node[anchor=south west,inner sep=0]  at (0,0) {\includegraphics[width=\linewidth]{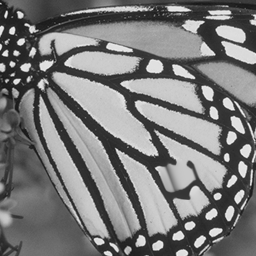}};
\spy on (1.55,.66) in node [right] at (0.02,-1.14);
\end{tikzpicture}
  \caption*{EPLL}
\end{subfigure}%
\hfill
\begin{subfigure}[bt]{.123\textwidth}
\begin{tikzpicture}[spy using outlines=
{rectangle,black,magnification=5,size=2.17cm, connect spies}]
\node[anchor=south west,inner sep=0]  at (0,0) {\includegraphics[width=\linewidth]{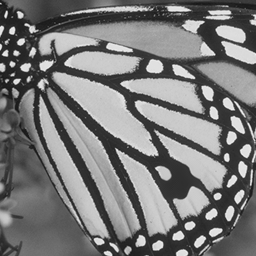}};
\spy on (1.55,.66) in node [right] at (0.02,-1.14);
\end{tikzpicture}
  \caption*{patchNR}
\end{subfigure}%
\hfill
\begin{subfigure}[bt]{.123\textwidth}
\begin{tikzpicture}[spy using outlines=
{rectangle,black,magnification=5,size=2.17cm, connect spies}]
\node[anchor=south west,inner sep=0]  at (0,0) {\includegraphics[width=\linewidth]{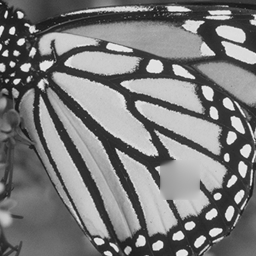}};
\spy on (1.55,.66) in node [right] at (0.02,-1.14);
\end{tikzpicture}
  \caption*{ALR}
\end{subfigure}%
\hfill
\begin{subfigure}[bt]{.123\textwidth}
\begin{tikzpicture}[spy using outlines=
{rectangle,black,magnification=5,size=2.17cm, connect spies}]
\node[anchor=south west,inner sep=0]  at (0,0) {\includegraphics[width=\linewidth]{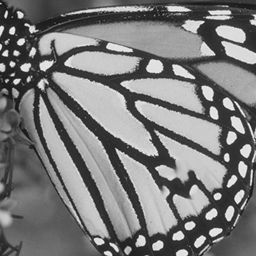}};
\spy on (1.55,.66) in node [right] at (0.02,-1.14);
\end{tikzpicture}
  \caption*{WPP}
\end{subfigure}%
\hfill
\begin{subfigure}[bt]{.123\textwidth}
\begin{tikzpicture}[spy using outlines=
{rectangle,black,magnification=5,size=2.17cm, connect spies}]
\node[anchor=south west,inner sep=0]  at (0,0) {\includegraphics[width=\linewidth]{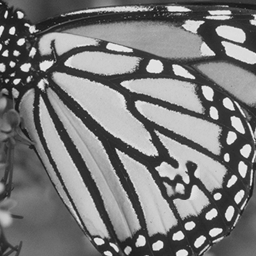}};
\spy on (1.55,.66) in node [right] at (0.02,-1.14);
\end{tikzpicture}
  \caption*{WP$\text{P}_{\varepsilon}$}
\end{subfigure}%
\hfill
\begin{subfigure}[bt]{.123\textwidth}
\begin{tikzpicture}[spy using outlines=
{rectangle,black,magnification=5,size=2.17cm, connect spies}]
\node[anchor=south west,inner sep=0]  at (0,0) {\includegraphics[width=\linewidth]{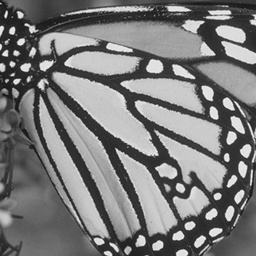}};
\spy on (1.55,.66) in node [right] at (0.02,-1.14);
\end{tikzpicture}
  \caption*{WP$\text{P}_{\varepsilon, \rho}$}
\end{subfigure}%

\caption{Comparison of different methods for  inpainting. The black boxes mark missing parts. The reference patches for the regularizers are obtained from the region within the white border.
The zoomed-in part is marked with a black box.
\textit{Top}: full image. \textit{Bottom}: zoomed-in part.
} \label{fig:comp_inpainting}
\end{figure}

In Figure~\ref{fig:comp_inpainting}, we compare the results of the different regularizers for the inpainting task. The missing part is the black rectangle in the observation and the reference patches are extracted around the missing part, which is visualized with the larger white box. We  observe that ALR fails completely. In contrast,  EPLL and patchNR are able to connect the lower missing black line. 
Here, the patchNR gives visually better results, in particular, the black lines are much sharper. Further, WPP,  WP$\text{P}_{\varepsilon}$ and  WP$\text{P}_{\varepsilon, \rho}$ fill out the missing part in a different way, as they aim to match the patch distribution between the reference part and the missing part. Obviously, the filled area is influenced by the patch distribution of the lower right corner in the reference part.

\subsection{Posterior Sampling} \label{sec:posterior_inpainting}

In this section, we apply ULA \eqref{eq:ULA}. 
First, we use it for posterior sampling in image inpainting, where we can expect a high variety in the reconstructions due to the highly ill-posed problem. 
Then we quantify the uncertainty in
limited-angle CT reconstructions.

\begin{figure}[!t]\captionsetup[subfigure]{font=normal}
\centering
\begin{subfigure}[bt]{.15\textwidth}  
\caption*{\phantom{\textbf{EPLL}}}
\includegraphics[width=\linewidth]{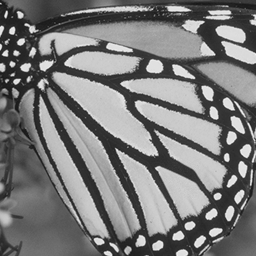}
\end{subfigure}%
\hfill
\begin{subfigure}[bt]{.15\textwidth}  
\caption*{\textbf{EPLL}}
\includegraphics[width=\linewidth]{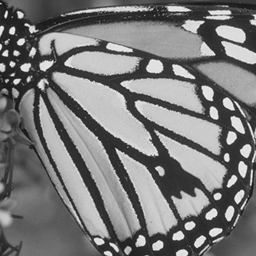}
\end{subfigure}%
\hfill
\begin{subfigure}[bt]{.15\textwidth}  
\caption*{\phantom{\textbf{EPLL}}}
\includegraphics[width=\linewidth]{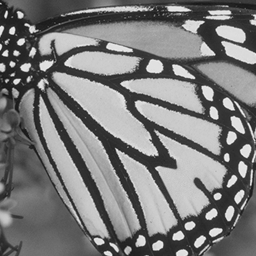}
\end{subfigure}%
\hspace{1.3cm}
\begin{subfigure}[bt]{.15\textwidth}
\caption*{\phantom{\textbf{patchNR}}}
\includegraphics[width=\linewidth]{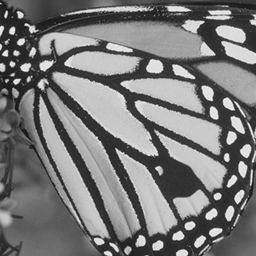}
\end{subfigure}%
\hfill
\begin{subfigure}[bt]{.15\textwidth}
\caption*{\textbf{patchNR}}
\includegraphics[width=\linewidth]{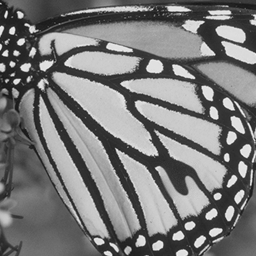}
\end{subfigure}%
\hfill
\begin{subfigure}[bt]{.15\textwidth}
\caption*{\phantom{\textbf{patchNR}}}
\includegraphics[width=\linewidth]{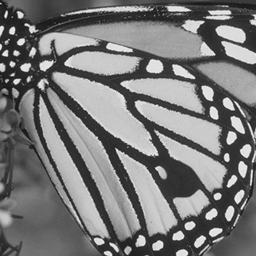}
\end{subfigure}%

\begin{subfigure}[bt]{.15\textwidth}  
\caption*{\phantom{\textbf{ALR}}}
\includegraphics[width=\linewidth]{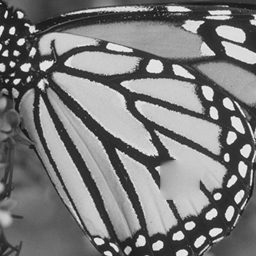}
\end{subfigure}%
\hfill
\begin{subfigure}[bt]{.15\textwidth}  
\caption*{\textbf{ALR}}
\includegraphics[width=\linewidth]{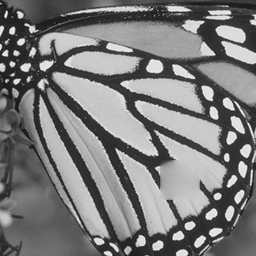}
\end{subfigure}%
\hfill
\begin{subfigure}[bt]{.15\textwidth}  
\caption*{\phantom{\textbf{ALR}}}
\includegraphics[width=\linewidth]{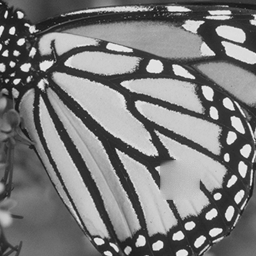}
\end{subfigure}%
\hspace{1.3cm}
\begin{subfigure}[bt]{.15\textwidth}
\caption*{\phantom{\textbf{WPP}}}
\includegraphics[width=\linewidth]{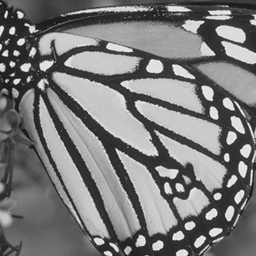}
\end{subfigure}%
\hfill
\begin{subfigure}[bt]{.15\textwidth}
\caption*{\textbf{WPP}}
\includegraphics[width=\linewidth]{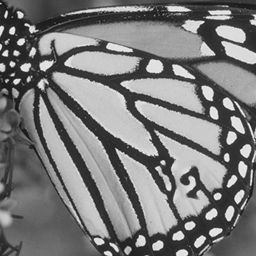}
\end{subfigure}%
\hfill
\begin{subfigure}[bt]{.15\textwidth}
\caption*{\phantom{\textbf{WPP}}}
\includegraphics[width=\linewidth]{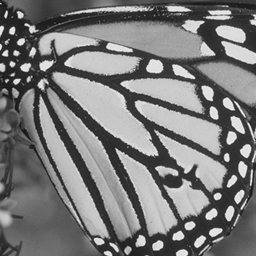}
\end{subfigure}%

\begin{subfigure}[bt]{.15\textwidth}  
\caption*{\phantom{\textbf{$\text{WPP}_{\varepsilon}$}}}
\vspace{0.05cm}
\includegraphics[width=\linewidth]{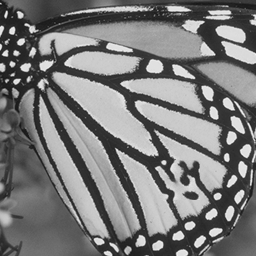}
\end{subfigure}%
\hfill
\begin{subfigure}[bt]{.15\textwidth}  
\caption*{\textbf{$\text{WPP}_{\varepsilon}$}} \vspace{0.05cm}
\includegraphics[width=\linewidth]{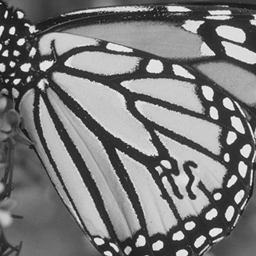}
\end{subfigure}%
\hfill
\begin{subfigure}[bt]{.15\textwidth}  
\caption*{\phantom{\textbf{$\text{WPP}_{\varepsilon}$}}}
\vspace{0.05cm}
\includegraphics[width=\linewidth]{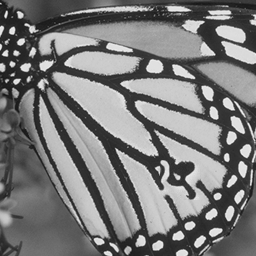}
\end{subfigure}%
\hspace{1.3cm}
\begin{subfigure}[bt]{.15\textwidth}
\caption*{\phantom{\textbf{$\text{WPP}_{\varepsilon,\rho}$}}}
\vspace{0.05cm}
\includegraphics[width=\linewidth]{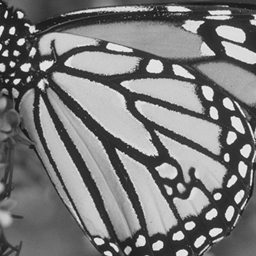}
\end{subfigure}%
\hfill
\begin{subfigure}[bt]{.15\textwidth}
\caption*{\textbf{$\text{WPP}_{\varepsilon,\rho}$}}
\vspace{0.05cm}
\includegraphics[width=\linewidth]{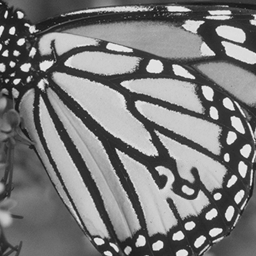}
\end{subfigure}%
\hfill
\begin{subfigure}[bt]{.15\textwidth}
\caption*{\phantom{\textbf{$\text{WPP}_{\varepsilon,\rho}$}}}
\vspace{0.05cm}
\includegraphics[width=\linewidth]{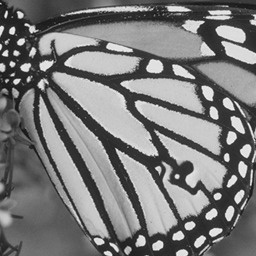}
\end{subfigure}%
\caption{Comparison of regularizers for posterior sampling for inpainting. The ground truth and the observation images are the same as in Figure~\ref{fig:comp_inpainting}. 
} \label{fig:comp_in_langevin}
\end{figure}

\paragraph{Posterior Sampling for Image Inpainting}

We apply ULA \eqref{eq:ULA} with the different regularizers for the same task of image inpainting as in Section~\ref{sec:img_inpainting}. Again, the data-fidelity term vanishes, so that \eqref{eq:ULA_reg} simplifies to
\begin{align*}
X_{k+1} = X_k - \delta \lambda \nabla \mathcal R(X_k) + \sqrt{2 \delta} Z_{k+1}.
\end{align*}
Since the inverse problem is highly ill-posed due to the missing part, we can expect a high variety in the reconstructions.
In Figure~\ref{fig:comp_in_langevin}, we compare the different methods. The ground truth and the observation are the same as in Figure~\ref{fig:comp_inpainting}. We illustrated three different reconstruction samples.
Again, we observe differences between the regularizers of Sections~\ref{sec:EPLL} and \ref{sec:OT}. First, we note that the ALR is, similar to MAP inpainting, not able to give meaningful reconstructions. On the other hand, the EPLL and the patchNR can reconstruct well, although the EPLL reconstructions look more realistic and are more diverse.
The regularizers from Section~\ref{sec:OT} give the most diverse reconstructions. Here the reconstruction quality is similar for the WPP,  the WP$\text{P}_{\varepsilon}$ and the WP$\text{P}_{\varepsilon, \rho}$.

\paragraph{Uncertainty Quantification for Limited-Angle CT}

\begin{figure}[!t]\captionsetup[subfigure]{font=normal}
\centering
\begin{subfigure}[bt]{.15\textwidth}  
\caption*{\phantom{\textbf{EPLL}}}
\includegraphics[width=\linewidth]{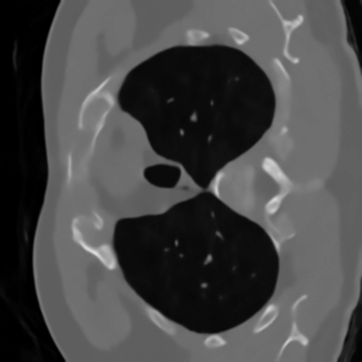}

\end{subfigure}%
\hfill
\begin{subfigure}[bt]{.15\textwidth}  
\caption*{\hspace{-2.7cm}\textbf{EPLL}}
\includegraphics[width=\linewidth]{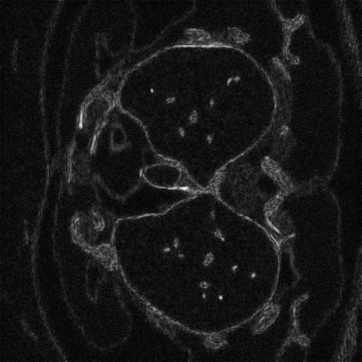}

\end{subfigure}%
\hspace{.7cm}
\begin{subfigure}[bt]{.15\textwidth}
\caption*{\phantom{\textbf{patchNR}}}
\includegraphics[width=\linewidth]{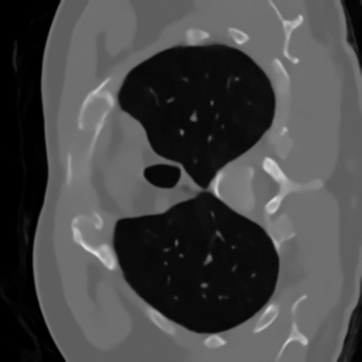}
\end{subfigure}%
\hfill
\begin{subfigure}[bt]{.15\textwidth}
\caption*{\hspace{-2.7cm}\textbf{patchNR}}
\includegraphics[width=\linewidth]{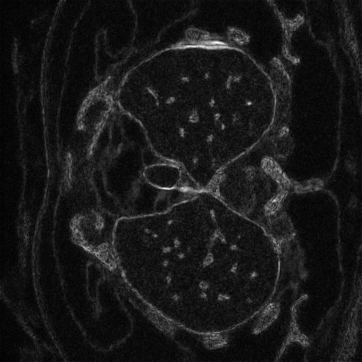}
\end{subfigure}%
\hspace{.7cm}
\begin{subfigure}[bt]{.15\textwidth}  
\caption*{\phantom{\textbf{ALR}}}
\includegraphics[width=\linewidth]{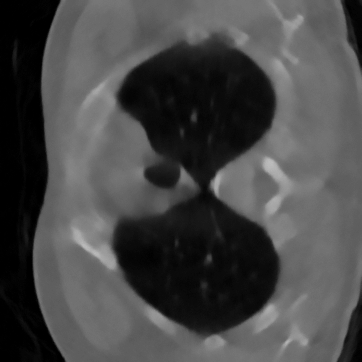}

\end{subfigure}%
\hfill
\begin{subfigure}[bt]{.15\textwidth}  
\caption*{\hspace{-2.7cm}\textbf{ALR}}
\includegraphics[width=\linewidth]{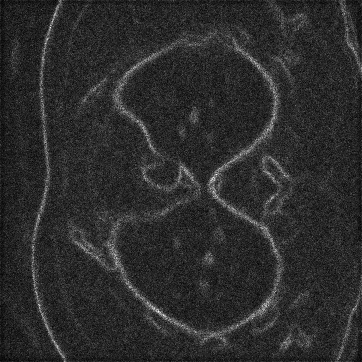}
\end{subfigure}%

\begin{subfigure}[bt]{.15\textwidth}
\caption*{\phantom{\textbf{WPP}}}
\includegraphics[width=\linewidth]{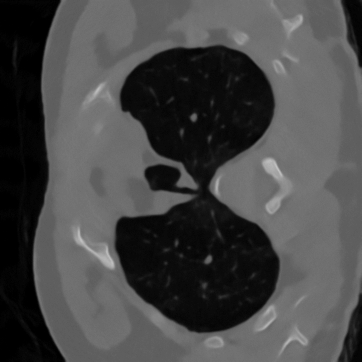}
\end{subfigure}%
\hfill
\begin{subfigure}[bt]{.15\textwidth}
\caption*{\hspace{-2.7cm}\textbf{WPP}}
\includegraphics[width=\linewidth]{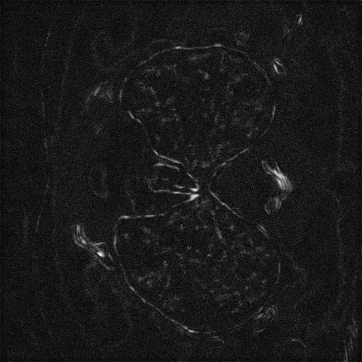}
\end{subfigure}%
\hspace{.7cm}
\begin{subfigure}[bt]{.15\textwidth}  
\caption*{\phantom{\textbf{$\text{WPP}_{\varepsilon}$}}}
\includegraphics[width=\linewidth]{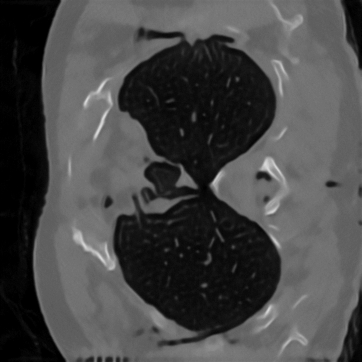}

\end{subfigure}%
\hfill
\begin{subfigure}[bt]{.15\textwidth}  
\caption*{\hspace{-2.7cm}\textbf{$\text{WPP}_{\varepsilon}$}}
\includegraphics[width=\linewidth]{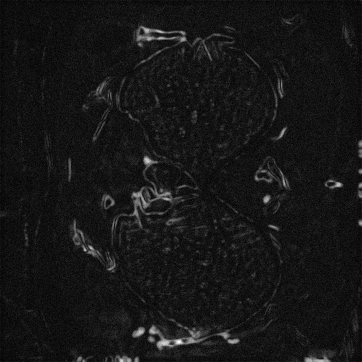}

\end{subfigure}%
\hspace{.7cm}
\begin{subfigure}[bt]{.15\textwidth}
\caption*{\phantom{\textbf{$\text{WPP}_{\varepsilon,\rho}$}}}
\includegraphics[width=\linewidth]{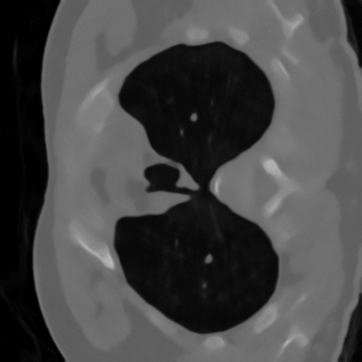}
\end{subfigure}%
\hfill
\begin{subfigure}[bt]{.15\textwidth}
\caption*{\hspace{-2.7cm}\textbf{$\text{WPP}_{\varepsilon,\rho}$}}
\vspace{0.04cm}
\includegraphics[width=\linewidth]{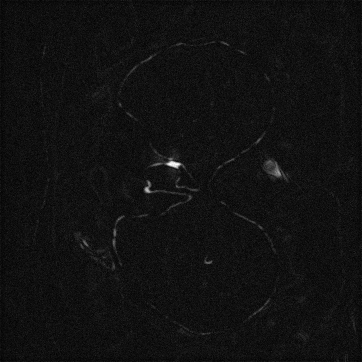}
\end{subfigure}%
\caption{Comparison of regularizers for uncertainty quantification for limited-angle CT. Mean image (left) and pixel-wise standard deviation (right) of the reconstructions. The ground truth and the FBP are the same as in Figure~\ref{fig:comp_CT_lim}.
} \label{fig:langevin_CT}
\end{figure}

Finally, we consider the limited-angle CT reconstruction as in Section~\ref{sec:CT}. The negative log-likelihood is given by \eqref{eq:ct_nll} so that \eqref{eq:ULA_reg} reads as
\begin{align*}
X_{k+1} = X_k  + \delta \nabla \sum_{i=1}^d  e^{-F(X_k)_i \mu} N_0 + e^{-y_i \mu} N_0 \big(F(X_k)_i \mu - \log(N_0) \big)  - \delta \alpha \nabla \mathcal R(X_k) + \sqrt{2 \delta} Z_{k+1}.
\end{align*}
In Figure~\ref{fig:langevin_CT}, we compare the reconstructions of the different regularizers. We illustrate the mean image (left) and the pixel-wise standard deviation (right) of the corresponding regularizers for 10 reconstructions. The standard deviation can be seen as the uncertainty in the reconstruction and the brighter a pixel is, the less secure is the model in its reconstruction. 
As in the MAP reconstruction, EPLL and patchNR are able to reconstruct best. Moreover, the standard deviation of EPLL and patchNR are most meaningful and the highest uncertainty is in regions, where the FBP has missing parts. In contrast, the ALR smooths out the reconstruction.
While the reconstructions of WPP and $\text{WPP}_{\varepsilon,\rho}$ appear almost similar at first glance, the WPP admits more uncertainty in its reconstructions. Nevertheless, both regularizers are not able to reconstruct the corrupted parts in the FBP. The $\text{WPP}_{\varepsilon}$ has a lot of artifacts in its reconstructions. Moreover, most of the uncertainty is observable in the artificially reconstructed artifacts.

\section*{Acknowledgements}
M.P. and G.S. acknowledge funding from the German Research Foundation (DFG) within the project BIOQIC (GRK2260). 
F.A., A.W. and G.S. acknowledge support from the DFG through Germany's Excellence Strategy – The Berlin Mathematics Research Center MATH+ under project AA5-6. P.H. acknowledges support from the DFG within the SPP 2298 "Theoretical Foundations of Deep Learning" (STE 571/17-1). 
J.H. acknowledges support from the DFG within the project STE 571/16-1.

The material data presented in Section \ref{sec:superres} was obtained as part of the EU Horizon 2020 Marie Sklodowska-Curie Actions Innovative Training Network MUMMERING (MUltiscale, Multimodal, and Multidimensional imaging for EngineeRING, Grant Number 765604) at the TOMCAT beamline of the Swiss Light Source (SLS), performed by A. Saadaldin, D. Bernard, and F. Marone Welford. We express our gratitude to the Paul Scherrer Institut, Villigen, Switzerland, for providing synchrotron radiation beamtime at the TOMCAT beamline X02DA of the SLS.
\bibliographystyle{abbrv}
\bibliography{references.bib}

\begin{thebibliography}{100}

\bibitem{Adler2018}
J.~Adler, H.~Kohr, A.~Ringh, J.~Moosmann, S.~Banert, M.~J. Ehrhardt, G.~R. Lee,
  K.~Niinimaki, B.~Gris, O.~Verdier, J.~Karlsson, W.~J. Palenstijn, O.~Öktem,
  C.~Chen, H.~A. Loarca, and M.~Lohmann.
\newblock Operator discretization library {(ODL)}, 2018.

\bibitem{adler_deep}
J.~Adler and O.~Öktem.
\newblock Deep {Bayesian} inversion.
\newblock {\em arXiv preprint arXiv:1811.05910}, 2018.

\bibitem{altekruger2023patchnr}
F.~Altekr{\"u}ger, A.~Denker, P.~Hagemann, J.~Hertrich, P.~Maass, and
  G.~Steidl.
\newblock {PatchNR}: {L}earning from very few images by patch normalizing flow
  regularization.
\newblock {\em Inverse Problems}, 39(6):064006, 2023.

\bibitem{AHS2023a}
F.~Altekr\"uger, P.~Hagemann, and G.~Steidl.
\newblock Conditional generative models are provably robust: pointwise
  guarantees for {B}ayesian inverse problems.
\newblock {\em Transactions on Machine Learning Research}, 2023.

\bibitem{altekruger2023wppnets}
F.~Altekr{\"u}ger and J.~Hertrich.
\newblock {WPPNets and WPPFlows:} {T}he power of {W}asserstein patch priors for
  superresolution.
\newblock {\em SIAM Journal on Imaging Sciences}, 16(3):1033--1067, 2023.

\bibitem{graz_inc}
A.~Andrle, N.~Farchmin, P.~Hagemann, S.~Heidenreich, V.~Soltwisch, and
  G.~Steidl.
\newblock Invertible neural networks versus {MCMC} for posterior reconstruction
  in grazing incidence x-ray fluorescence.
\newblock {\em International Conference on Scale Space and Variational Methods
  in Computer Vision}, page 528–539, 2021.

\bibitem{ardizzone2018analyzing}
L.~Ardizzone, J.~Kruse, C.~Rother, and U.~K{\"o}the.
\newblock Analyzing inverse problems with invertible neural networks.
\newblock {\em International Conference on Learning Representations}, 2018.

\bibitem{ardizzone2019guided}
L.~Ardizzone, C.~L{\"u}th, J.~Kruse, C.~Rother, and U.~K{\"o}the.
\newblock Guided image generation with conditional invertible neural networks.
\newblock {\em arXiv preprint arXiv:1907.02392}, 2019.

\bibitem{arjovsky2017wasserstein}
M.~Arjovsky, S.~Chintala, and L.~Bottou.
\newblock Wasserstein generative adversarial networks.
\newblock {\em International Conference on Machine Learning}, pages 214--223,
  2017.

\bibitem{Armato11}
S.~G. Armato~et al.
\newblock The {Lung Image Database Consortium (LIDC)} and {Image Database
  Resource Initiative (IDRI)}: A completed reference database of lung nodules
  on {CT} scans.
\newblock {\em Medical Physics}, 38(2):915--931, 2011.

\bibitem{AMOS2019}
S.~Arridge, P.~Maass, \"Oktem, and C.~Schß"onlieb.
\newblock Solving inverse problems using data-driven models.
\newblock {\em Acta Numerica}, 28:1--174, 2019.

\bibitem{cond_score}
G.~Batzolis, J.~Stanczuk, C.-B. Schönlieb, and C.~Etmann.
\newblock Conditional image generation with score-based diffusion models.
\newblock {\em arXiv preprint arXiv:2111.13606}, 2021.

\bibitem{BGCDJ2019}
J.~Behrmann, W.~Grathwohl, R.~Chen, D.~Duvenaud, and J.-H. Jacobsen.
\newblock Invertible residual networks.
\newblock {\em International Conference on Machine Learning}, pages 573--582,
  2019.

\bibitem{behrmann2020understanding}
J.~Behrmann, P.~Vicol, K.-C. Wang, R.~Grosse, and J.-H. Jacobsen.
\newblock Understanding and mitigating exploding inverses in invertible neural
  networks.
\newblock {\em ArXiv 2006.09347}, 2020.

\bibitem{benning2018modern}
M.~Benning and M.~Burger.
\newblock Modern regularization methods for inverse problems.
\newblock {\em Acta Numerica}, 27:1--111, 2018.

\bibitem{bertero2021introduction}
M.~Bertero, P.~Boccacci, and C.~De~Mol.
\newblock {\em Introduction to Inverse Problems in Imaging}.
\newblock CRC Press, 2021.

\bibitem{BRGA2012}
M.~Bevilacqua, A.~Roumy, C.~M. Guillemot, and M.-L. Alberi-Morel.
\newblock Low-complexity single-image super-resolution based on nonnegative
  neighbor embedding.
\newblock {\em British Machine Vision Conference}, 2012.

\bibitem{CTMLSZ2023}
Z.~Cai, J.~Tang, S.~Mukherjee, J.~Li, C.~B. Schönlieb, and X.~Zhang.
\newblock {NF-ULA:} {Langevin Monte Carlo} with normalizing flow prior for
  imaging inverse problems.
\newblock {\em arXiv preprint arXiv:2304.08342}, 2023.

\bibitem{chen2016patchstyle}
T.~Q. Chen and M.~Schmidt.
\newblock Fast patch-based style transfer of arbitrary style.
\newblock {\em Advances in Neural Information Processing Systems}, 2016.

\bibitem{dabov2007}
K.~Dabov, A.~Foi, V.~Katkovnik, and K.~Egiazarian.
\newblock Image denoising by sparse 3-d transform-domain collaborative
  filtering.
\newblock {\em IEEE Transactions on Image Processing}, 16(8):2080--2095, 2007.

\bibitem{deledalle2018image}
C.-A. Deledalle, S.~Parameswaran, and T.~Q. Nguyen.
\newblock Image denoising with generalized {G}aussian mixture model patch
  priors.
\newblock {\em SIAM Journal on Imaging Sciences}, 11(4):2568--2609, 2018.

\bibitem{dempster1977maximum}
A.~P. Dempster, N.~M. Laird, and D.~B. Rubin.
\newblock Maximum likelihood from incomplete data via the {EM} algorithm.
\newblock {\em Journal of the Royal Statistical Society: Series B},
  39(1):1--22, 1977.

\bibitem{dinh2016density}
L.~Dinh, J.~Sohl-Dickstein, and S.~Bengio.
\newblock Density estimation using {Real NVP}.
\newblock {\em International Conference on Learning Representations}, 2016.

\bibitem{dong2011sparsity}
W.~Dong, X.~Li, L.~Zhang, and G.~Shi.
\newblock Sparsity-based image denoising via dictionary learning and structural
  clustering.
\newblock {\em IEEE Conference on Computer Vision and Pattern Recognition},
  pages 457--464, 2011.

\bibitem{DLPYL2023}
C.~Du, T.~Li, T.~Pang, S.~Yan, and M.~Lin.
\newblock Nonparametric generative modeling with conditional
  sliced-{Wasserstein} flows.
\newblock {\em International Conference on Machine Learning}, pages 8565--8584,
  2023.

\bibitem{elnekave2022generating}
A.~Elnekave and Y.~Weiss.
\newblock Generating natural images with direct patch distributions matching.
\newblock {\em European Conference on Computer Vision}, pages 544--560, 2022.

\bibitem{engl1996regularization}
H.~W. Engl, M.~Hanke, and A.~Neubauer.
\newblock {\em Regularization of inverse problems}, volume 375.
\newblock Springer Science \& Business Media, 1996.

\bibitem{feydy2019interpolating}
J.~Feydy, T.~S{\'e}journ{\'e}, F.-X. Vialard, S.-i. Amari, A.~Trouv{\'e}, and
  G.~Peyr{\'e}.
\newblock Interpolating between optimal transport and {MMD} using {S}inkhorn
  divergences.
\newblock {\em International Conference on Artificial Intelligence and
  Statistics}, pages 2681--2690, 2019.

\bibitem{friedman2021posterior}
R.~Friedman and Y.~Weiss.
\newblock Posterior sampling for image restoration using explicit patch priors.
\newblock {\em arXiv preprint arXiv:2104.09895}, 2021.

\bibitem{gatys2015texture}
L.~Gatys, A.~S. Ecker, and M.~Bethge.
\newblock Texture synthesis using convolutional neural networks.
\newblock {\em Advances in Neural Information Processing Systems}, 28, 2015.

\bibitem{gatys2016image}
L.~Gatys, A.~S. Ecker, and M.~Bethge.
\newblock Image style transfer using convolutional neural networks.
\newblock {\em IEEE Conference on Computer Vision and Pattern Recognition},
  pages 2414--2423, 2016.

\bibitem{genevay2016stochastic}
A.~Genevay, M.~Cuturi, G.~Peyr{\'e}, and F.~Bach.
\newblock Stochastic optimization for large-scale optimal transport.
\newblock {\em Advances in Neural Information Processing Systems}, 29, 2016.

\bibitem{genevay2018learning}
A.~Genevay, G.~Peyr{\'e}, and M.~Cuturi.
\newblock Learning generative models with {S}inkhorn divergences.
\newblock {\em International Conference on Artificial Intelligence and
  Statistics}, pages 1608--1617, 2018.

\bibitem{gilton2019learned}
D.~Gilton, G.~Ongie, and R.~Willett.
\newblock Learned patch-based regularization for inverse problems in imaging.
\newblock {\em IEEE International Workshop on Computational Advances in
  Multi-Sensor Adaptive Processing}, pages 211--215, 2019.

\bibitem{GBI2009}
D.~Glasner, S.~Bagon, and M.~Irani.
\newblock Super-resolution from a single image.
\newblock {\em IEEE International Conference on Computer Vision}, pages
  349--356, 2009.

\bibitem{GPMXWOCB2014}
I.~Goodfellow, J.~Pouget-Abadie, M.~Mirza, B.~Xu, D.~Warde-Farley, S.~Ozair,
  A.~Courville, and Y.~Bengio.
\newblock Generative adversarial nets.
\newblock {\em Advances in Neural Information Processing Systems}, 27, 2014.

\bibitem{GFO2017}
D.~Grana, T.~Fjeldstad, and H.~Omre.
\newblock Bayesian {G}aussian mixture linear inversion for geophysical inverse
  problems.
\newblock {\em Mathematical Geosciences}, 49(4):493--515, 2017.

\bibitem{granot2022drop}
N.~Granot, B.~Feinstein, A.~Shocher, S.~Bagon, and M.~Irani.
\newblock Drop the {GAN}: {I}n defense of patches nearest neighbors as single
  image generative models.
\newblock {\em IEEE/CVF Conference on Computer Vision and Pattern Recognition},
  pages 13460--13469, 2022.

\bibitem{GW2008}
A.~Griewank and A.~Walther.
\newblock {\em Evaluating Derivatives: Principles and Techniques of Algorithmic
  Differentiation}.
\newblock SIAM, 2008.

\bibitem{grover2018flow}
A.~Grover, M.~Dhar, and S.~Ermon.
\newblock Flow-{GAN}: {C}ombining maximum likelihood and adversarial learning
  in generative models.
\newblock {\em Proceedings of the AAAI Conference on Artificial Intelligence},
  32(1), 2018.

\bibitem{gulrajani2017improved}
I.~Gulrajani, F.~Ahmed, M.~Arjovsky, V.~Dumoulin, and A.~C. Courville.
\newblock Improved training of {W}asserstein {GANs}.
\newblock {\em Advances in Neural Information Processing Systems}, 30, 2017.

\bibitem{GRGH2017}
J.~Gutierrez, J.~Rabin, B.~Galerne, and T.~Hurtut.
\newblock Optimal patch assignment for statistically constrained texture
  synthesis.
\newblock {\em International Conference on Scale Space and Variational Methods
  in Computer Vision}, pages 172--183, 2017.

\bibitem{HHABCS2023}
P.~Hagemann, J.~Hertrich, F.~Altekr\"uger, R.~Beinert, J.~Chemseddine, and
  G.~Steidl.
\newblock Posterior sampling based on gradient flows of the {MMD} with negative
  distance kernel.
\newblock {\em arXiv preprint arXiv:2310.03054}, 2023.

\bibitem{hagemann2022stochastic}
P.~Hagemann, J.~Hertrich, and G.~Steidl.
\newblock Stochastic normalizing flows for inverse problems: {A} {M}arkov
  chains viewpoint.
\newblock {\em SIAM/ASA Journal on Uncertainty Quantification},
  10(3):1162--1190, 2022.

\bibitem{HHS2023}
P.~Hagemann, J.~Hertrich, and G.~Steidl.
\newblock {\em Generalized Normalizing Flows via Markov Chains}.
\newblock Elements in Non-local Data Interactions: Foundations and
  Applications. Cambridge University Press, 2023.

\bibitem{HN2021}
P.~Hagemann and S.~Neumayer.
\newblock Stabilizing invertible neural networks using mixture models.
\newblock {\em Inverse Problems}, 37(8), 2021.

\bibitem{HHLS2021}
M.~Hasannasab, J.~Hertrich, F.~Laus, and G.~.Steidl.
\newblock Alternatives to the {EM} algorithm for {ML} estimation of location,
  scatter matrix, and degree of freedom of the {S}tudent-t distribution.
\newblock {\em Numerical Algorithms}, 87(1):77--118, 2021.

\bibitem{he2018image}
W.~He, R.~Yu, Y.~Zheng, and T.~Jiang.
\newblock Image denoising using asymmetric {G}aussian mixture models.
\newblock {\em IEEE International Symposium in Sensing and Instrumentation in
  {IoT} Era}, pages 1--4, 2018.

\bibitem{hertrich2022wasserstein}
J.~Hertrich, A.~Houdard, and C.~Redenbach.
\newblock Wasserstein patch prior for image superresolution.
\newblock {\em IEEE Transactions on Computational Imaging}, 8:693--704, 2022.

\bibitem{HL2022}
J.~Hertrich, D.~P.~L. Nguyen, J.-F. Aujol, D.~Bernard, Y.~Berthoumieu,
  A.~Saadaldin, and G.~Steidl.
\newblock {{PCA}} reduced {{Gauss}}ian mixture models with application in
  superresolution.
\newblock {\em Inverse Problems and Imaging}, 16(2):341--366, 2022.

\bibitem{psnr_ssim}
A.~Hore and D.~Ziou.
\newblock Image quality metrics: {PSNR vs. SSIM}.
\newblock {\em International Conference on Pattern Recognition}, pages
  2366--2369, 2010.

\bibitem{houdard2018high}
A.~Houdard, C.~Bouveyron, and J.~Delon.
\newblock High-dimensional mixture models for unsupervised image denoising
  ({HDMI}).
\newblock {\em SIAM Journal on Imaging Sciences}, 11(4):2815--2846, 2018.

\bibitem{houdard2021wasserstein}
A.~Houdard, A.~Leclaire, N.~Papadakis, and J.~Rabin.
\newblock Wasserstein generative models for patch-based texture synthesis.
\newblock {\em International Conference on Scale Space and Variational Methods
  in Computer Vision}, pages 269--280, 2021.

\bibitem{houdard2023generative}
A.~Houdard, A.~Leclaire, N.~Papadakis, and J.~Rabin.
\newblock A generative model for texture synthesis based on optimal transport
  between feature distributions.
\newblock {\em Journal of Mathematical Imaging and Vision}, 65(1):4--28, 2023.

\bibitem{JKYB2020}
P.~Jaini, I.~Kobyzev, Y.~Yu, and M.~Brubaker.
\newblock Tails of {L}ipschitz triangular flows.
\newblock {\em International Conference on Machine Learning}, 119:4673--4681,
  2020.

\bibitem{KB2015}
D.~P. Kingma and J.~Ba.
\newblock Adam: a method for stochastic optimization.
\newblock {\em International Conference on Learning Representations}, 2015.

\bibitem{kingma2013auto}
D.~P. Kingma and M.~Welling.
\newblock Auto-encoding variational {B}ayes.
\newblock {\em arXiv preprint arXiv:1312.6114}, 2013.

\bibitem{Latz2020}
J.~Latz.
\newblock On the well-posedness of {Bayesian} inverse problems.
\newblock {\em SIAM/ASA Journal on Uncertainty Quantification}, 8(1):451–482,
  2020.

\bibitem{LBADDP2022}
R.~Laumont, V.~D. Bortoli, A.~Almansa, J.~Delon, A.~Durmus, and M.~Pereyra.
\newblock Bayesian imaging using {P}lug \& {P}lay priors: When {L}angevin meets
  {T}weedie.
\newblock {\em SIAM Journal on Imaging Sciences}, 15(2):701--737, 2022.

\bibitem{LNPS17}
F.~Laus, M.~Nikolova, J.~Persch, and G.~Steidl.
\newblock A nonlocal denoising algorithm for manifold-valued images using
  second order statistics.
\newblock {\em SIAM Journal on Imaging Sciences}, 10(1):416–448, 2017.

\bibitem{LBM2013}
M.~Lebrun, A.~Buades, and J.-M. Morel.
\newblock A nonlocal {B}ayesian image denoising algorithm.
\newblock {\em SIAM Journal on Imaging Sciences}, 6(3):1665--1688, 2013.

\bibitem{LCBM2012}
M.~Lebrun, M.~Colom, A.~Buades, and J.~Morel.
\newblock Secrets of image denoising cuisine.
\newblock {\em Acta Numerica}, 21:475--576, 2012.

\bibitem{LoDoPaB21}
J.~Leuschner, M.~Schmidt, D.~O. Baguer, and P.~Maass.
\newblock {LoDoPaB-CT}, a benchmark dataset for low-dose computed tomography
  reconstruction.
\newblock {\em Scientific Data}, 8(109), 2021.

\bibitem{demyst_neural_style}
Y.~Li, N.~Wang, J.~Liu, and X.~Hou.
\newblock Demystifying neural style transfer.
\newblock {\em International Joint Conference on Artificial Intelligence}, page
  2230–2236, 2017.

\bibitem{liang2001realtexture}
L.~Liang, C.~Liu, Y.-Q. Xu, B.~Guo, and H.-Y. Shum.
\newblock Real-time texture synthesis by patch-based sampling.
\newblock {\em ACM Transactions on Graphics}, 20(3):127--150, 2001.

\bibitem{lim2018molecular}
J.~Lim, S.~Ryu, J.~W. Kim, and W.~Y. Kim.
\newblock Molecular generative model based on conditional variational
  autoencoder for de novo molecular design.
\newblock {\em Journal of cheminformatics}, 10(1):1--9, 2018.

\bibitem{liu2021wasserstein}
S.~Liu, X.~Zhou, Y.~Jiao, and J.~Huang.
\newblock Wasserstein generative learning of conditional distribution.
\newblock {\em arXiv preprint arXiv:2112.10039}, 2021.

\bibitem{lunz2018adversariallearnedreg}
S.~Lunz, O.~{\"O}ktem, and C.-B. Sch{\"o}nlieb.
\newblock Adversarial regularizers in inverse problems.
\newblock {\em Advances in Neural Information Processing Systems}, 2018.

\bibitem{bsd_data}
D.~Martin, C.~Fowlkes, D.~Tal, and J.~Malik.
\newblock A database of human segmented natural images and its application to
  evaluating segmentation algorithms and measuring ecological statistics.
\newblock {\em IEEE International Conference on Computer Vision}, pages
  416--423, 2001.

\bibitem{mignon2023semi}
S.~Mignon, B.~Galerne, M.~Hidane, C.~Louchet, and J.~Mille.
\newblock Semi-unbalanced regularized optimal transport for image restoration.
\newblock {\em European Signal Processing Conference}, pages 466--470, 2023.

\bibitem{N1992}
R.~Neal.
\newblock Bayesian learning via stochastic dynamics.
\newblock {\em Advances in Neural Information Processing Systems}, 5, 1992.

\bibitem{NS2021}
S.~Neumayer and G.~Steidl.
\newblock From optimal transport to discrepancy.
\newblock {\em Handbook of Mathematical Models and Algorithms in Computer
  Vision and Imaging}, pages 1--36, 2021.

\bibitem{NHAB2023}
D.-P.-L. Nguyen, J.~Hertrich, J.-F. Aujol, and Y.~Berthoumieu.
\newblock Image super-resolution with {PCA} reduced generalized {G}aussian
  mixture models in materials science.
\newblock {\em Inverse Problems and Imaging}, 17(6):1165--1192, 2023.

\bibitem{osher2017low}
S.~Osher, Z.~Shi, and W.~Zhu.
\newblock Low dimensional manifold model for image processing.
\newblock {\em SIAM Journal on Imaging Sciences}, 10(4):1669--1690, 2017.

\bibitem{papyan2015multi}
V.~Papyan and M.~Elad.
\newblock Multi-scale patch-based image restoration.
\newblock {\em IEEE Transactions on Image Processing}, 25(1):249--261, 2015.

\bibitem{parameswaran2018accelerating}
S.~Parameswaran, C.-A. Deledalle, L.~Denis, and T.~Q. Nguyen.
\newblock Accelerating {GMM}-based patch priors for image restoration: {T}hree
  ingredients for a 100x speed-up.
\newblock {\em IEEE Transactions on Image Processing}, 28(2):687--698, 2018.

\bibitem{PyTorch2019}
A.~Paszke, S.~Gross, F.~Massa, A.~Lerer, J.~Bradbury, G.~Chanan, T.~Killeen,
  Z.~Lin, N.~Gimelshein, L.~Antiga, A.~Desmaison, A.~Kopf, E.~Yang, Z.~DeVito,
  M.~Raison, A.~Tejani, S.~Chilamkurthy, B.~Steiner, L.~Fang, J.~Bai, and
  S.~Chintala.
\newblock {PyTorch: A}n imperative style, high-performance deep learning
  library.
\newblock {\em Advances in Neural Information Processing Systems}, 32, 2019.

\bibitem{peyre2008non}
G.~Peyr{\'e}, S.~Bougleux, and L.~Cohen.
\newblock Non-local regularization of inverse problems.
\newblock {\em European Conference on Computer Vision}, pages 57--68, 2008.

\bibitem{prost2021learninglocalar}
J.~Prost, A.~Houdard, A.~Almansa, and N.~Papadakis.
\newblock Learning local regularization for variational image restoration.
\newblock {\em International Conference on Scale Space and Variational Methods
  in Computer Vision}, pages 358--370, 2021.

\bibitem{Radon86}
J.~Radon.
\newblock On the determination of functions from their integral values along
  certain manifolds.
\newblock {\em IEEE Transactions on Medical Imaging}, 5(4):170--176, 1986.

\bibitem{reibman2006quality}
A.~R. Reibman, R.~M. Bell, and S.~Gray.
\newblock Quality assessment for super-resolution image enhancement.
\newblock {\em IEEE International Conference on Image Processing}, pages
  2017--2020, 2006.

\bibitem{RR2004}
G.~O. Roberts and J.~S. Rosenthal.
\newblock {General state space Markov chains and MCMC algorithms}.
\newblock {\em Probability Surveys}, 1:20 -- 71, 2004.

\bibitem{RT1996}
G.~O. Roberts and R.~L. Tweedie.
\newblock {Exponential convergence of Langevin distributions and their discrete
  approximations}.
\newblock {\em Bernoulli}, 2(4):341 -- 363, 1996.

\bibitem{roth2005fields}
S.~Roth and M.~J. Black.
\newblock Fields of experts: {A} framework for learning image priors.
\newblock {\em IEEE Computer Society Conference on Computer Vision and Pattern
  Recognition}, 2:860--867, 2005.

\bibitem{rudin1992nonlinear}
L.~I. Rudin, S.~Osher, and E.~Fatemi.
\newblock Nonlinear total variation based noise removal algorithms.
\newblock {\em Physica D: Nonlinear Phenomena}, 60(1-4):259--268, 1992.

\bibitem{SBDD2022}
A.~Salmona, V.~D. Bortoli, J.~Delon, and A.~Desolneux.
\newblock Can push-forward generative models fit multimodal distributions?
\newblock {\em Advances in Neural Information Processing Systems}, 2022.

\bibitem{santambrogio2015optimal}
F.~Santambrogio.
\newblock {\em {Optimal Transport for Applied Mathematicians Calculus of
  Variations, PDEs, and Modeling}}, volume~55.
\newblock Springer, 2015.

\bibitem{psnr_ssim_fsim}
U.~Sara, M.~Akter, and M.~S. Uddin.
\newblock Image quality assessment through {FSIM, SSIM, MSE and PSNR—A
  comparative study}.
\newblock {\em Journal of Computer and Communications}, 7(3):8--18, 2019.

\bibitem{sejourne2023unbalanced}
T.~S{\'e}journ{\'e}, G.~Peyr{\'e}, and F.-X. Vialard.
\newblock Unbalanced optimal transport, from theory to numerics.
\newblock {\em Handbook of Numerical Analysis}, 24:407--471, 2023.

\bibitem{shaham2019singan}
T.~R. Shaham, T.~Dekel, and T.~Michaeli.
\newblock Singan: {L}earning a generative model from a single natural image.
\newblock {\em IEEE/CVF International Conference on Computer Vision}, pages
  4570--4580, 2019.

\bibitem{shocher2019ingan}
A.~Shocher, S.~Bagon, P.~Isola, and M.~Irani.
\newblock {InGAN}: {C}apturing and retargeting the {"DNA"} of a natural image.
\newblock {\em IEEE/CVF International Conference on Computer Vision}, pages
  4492--4501, 2019.

\bibitem{SCI2018}
A.~Shocher, N.~Cohen, and M.~Irani.
\newblock “{Z}ero-shot” super-resolution using deep internal learning.
\newblock {\em IEEE Conference on Computer Vision and Pattern Recognition},
  pages 3118--3126, 2018.

\bibitem{simoncelli2001natural}
E.~P. Simoncelli and B.~A. Olshausen.
\newblock Natural image statistics and neural representation.
\newblock {\em Annual Review of Neuroscience}, 24(1):1193--1216, 2001.

\bibitem{vgg19}
K.~Simonyan and A.~Zisserman.
\newblock Very deep convolutional networks for large-scale image recognition.
\newblock {\em International Conference on Learning Representations}, 2015.

\bibitem{SLY2015}
K.~Sohn, H.~Lee, and X.~Yan.
\newblock Learning structured output representation using deep conditional
  generative models.
\newblock {\em Advances in Neural Information Processing Systems},
  28:3483--3491, 2015.

\bibitem{song2021maximum}
Y.~Song, C.~Durkan, I.~Murray, and S.~Ermon.
\newblock Maximum likelihood training of score-based diffusion models.
\newblock {\em Advances in Neural Information Processing Systems},
  34:1415--1428, 2021.

\bibitem{song2021scorebased}
Y.~Song, J.~Sohl-Dickstein, D.~P. Kingma, A.~Kumar, S.~Ermon, and B.~Poole.
\newblock Score-based generative modeling through stochastic differential
  equations.
\newblock {\em International Conference on Learning Representations}, 2021.

\bibitem{Sprungk2020}
B.~Sprungk.
\newblock On the local {Lipschitz} stability of {{B}ayesian} inverse problems.
\newblock {\em Inverse Problems}, 36(5), 2020.

\bibitem{Stuart2010}
A.~M. Stuart.
\newblock Inverse problems: A {Bayesian} perspective.
\newblock {\em Acta Numerica}, 19:451–559, 2010.

\bibitem{tikhonov1963solution}
A.~N. Tikhonov.
\newblock On the solution of ill-posed problems and the method of
  regularization.
\newblock {\em Doklady Akademii Nauk}, 151(3):501--504, 1963.

\bibitem{torralba2003statistics}
A.~Torralba and A.~Oliva.
\newblock Statistics of natural image categories.
\newblock {\em Network: Computation in Neural Systems}, 14(3):391, 2003.

\bibitem{vaucher2007line}
S.~Vaucher, P.~Unifantowicz, C.~Ricard, L.~Dubois, M.~Kuball, J.-M.
  Catala-Civera, D.~Bernard, M.~Stampanoni, and R.~Nicula.
\newblock On-line tools for microscopic and macroscopic monitoring of microwave
  processing.
\newblock {\em Physica B: Condensed Matter}, 398(2):191--195, 2007.

\bibitem{wang2013sure}
Y.-Q. Wang and J.-M. Morel.
\newblock {SURE} guided {G}aussian mixture image denoising.
\newblock {\em SIAM Journal on Imaging Sciences}, 6(2):999--1034, 2013.

\bibitem{WT2011}
M.~Welling and Y.~W. Teh.
\newblock Bayesian learning via stochastic gradient {L}angevin dynamics.
\newblock {\em International Conference on Machine Learning}, page 681–688,
  2011.

\bibitem{winkler_cond_flows}
C.~Winkler, D.~Worrall, E.~Hoogeboom, and M.~Welling.
\newblock Learning likelihoods with conditional normalizing flows.
\newblock {\em arXiv preprint arXiv:1912.00042}, 2019.

\bibitem{wu2020}
H.~Wu, J.~K\"{o}hler, and F.~Noé.
\newblock Stochastic normalizing flows.
\newblock {\em Advances in Neural Information Processing Systems}, 33, 2020.

\bibitem{yu2023epll}
Q.~Yu, G.~Cao, H.~Shi, Y.~Zhang, and P.~Fu.
\newblock {EPLL} image denoising with multi-feature dictionaries.
\newblock {\em Digital Signal Processing}, 137, 2023.

\bibitem{zach2023explicit}
M.~Zach, T.~Pock, E.~Kobler, and A.~Chambolle.
\newblock Explicit diffusion of {G}aussian mixture model based image priors.
\newblock {\em International Conference on Scale Space and Variational Methods
  in Computer Vision}, pages 3--15, 2023.

\bibitem{fsim}
L.~Zhang, L.~Zhang, X.~Mou, and D.~Zhang.
\newblock {FSIM:} {A} feature similarity index for image quality assessment.
\newblock {\em IEEE Transactions on Image Processing}, 20(8):2378--2386, 2011.

\bibitem{lpips}
R.~Zhang, P.~Isola, A.~A. Efros, E.~Shechtman, and O.~Wang.
\newblock The unreasonable effectiveness of deep features as a perceptual
  metric.
\newblock {\em IEEE Conference on Computer Vision and Pattern Recognition},
  pages 586--595, 2018.

\bibitem{zontak2011internalpatchstatistics}
M.~Zontak and M.~Irani.
\newblock Internal statistics of a single natural image.
\newblock {\em IEEE Conference on Computer Vision and Pattern Recognition},
  pages 977--984, 2011.

\bibitem{epll}
D.~Zoran and Y.~Weiss.
\newblock From learning models of natural image patches to whole image
  restoration.
\newblock {\em International Conference on Computer Vision}, pages 479--486,
  2011.

\end{thebibliography}
\end{document}